\pdfoutput=1

\documentclass[11pt, table]{article}

\usepackage[preprint]{acl}

\usepackage{times}
\usepackage{latexsym}

\usepackage[T5,T1]{fontenc}
\DeclareTextSymbolDefault{\ohorn}{T5}
\DeclareTextSymbolDefault{\uhorn}{T5}

\usepackage[utf8]{inputenc}

\usepackage{microtype}

\usepackage{inconsolata}

\usepackage{graphicx}

\usepackage{booktabs}
\usepackage{multirow}
\usepackage{adjustbox}
\usepackage[inline]{enumitem}

\usepackage{placeins}

\newcommand{\dataset}[2]{#1\_\allowbreak#2}

\title{Analyzing the Effect of Linguistic Similarity on Cross-Lingual Transfer:\\Tasks and Experimental Setups Matter}

\newcommand\blfootnote[1]{%
  \begingroup
  \renewcommand\thefootnote{}\footnote{#1}%
  \addtocounter{footnote}{-1}%
  \endgroup
}

\newcommand{\measure}[1]{\texttt{#1}}
\newcommand{\grambank}{\measure{gb}}
\newcommand{\syntax}{\measure{syn}}
\newcommand{\phono}{\measure{pho}}
\newcommand{\inventory}{\measure{inv}}
\newcommand{\asjp}{\measure{lex}}
\newcommand{\genetic}{\measure{gen}}
\newcommand{\geo}{\measure{geo}}
\newcommand{\overlapchar}{\measure{chr}}
\newcommand{\overlapword}{\measure{wor}}
\newcommand{\overlapsubword}{\measure{swt}}
\newcommand{\overlaptri}{\measure{tri}}
\newcommand{\trainingsize}{\measure{size}}

\newcommand{\ravg}[1]{\mbox{$r$\textit{\textsubscript{avg\_#1}}}}
\newcommand{\stdev}[1]{\textsubscript{#1}}
\newcommand{\smallasterisk}{\rlap{\textsuperscript{*}}}

\newcommand{\topicsbase}{\texttt{topics-\allowbreak{}base}}
\newcommand{\topicstranslit}{\texttt{topics-\allowbreak{}translit}}
\newcommand{\topicsmbert}{\texttt{topics-\allowbreak{}mbert}}
\newcommand{\Topicsmbert}{\texttt{Topics-\allowbreak{}mbert}}

\newcommand{\https}[1]{\href{https://#1}{\texttt{#1}}}

\author{
 \textbf{Verena Blaschke\textsuperscript{1,2,*}} \qquad
 \textbf{Masha Fedzechkina\textsuperscript{3}} \qquad
 \textbf{Maartje ter Hoeve\textsuperscript{3}}
\\
 \textsuperscript{1}Center for Information and Language Processing (CIS), LMU Munich\\
 \textsuperscript{2}Munich Center for Machine Learning\\
 \textsuperscript{3}Apple
\\
 \texttt{blaschke@cis.lmu.de}, \texttt{\{mfedzechkina, m\_terhoeve\}@apple.com}
}

\begin{document}
\maketitle
\begin{abstract}
Cross-lingual transfer is a popular approach to increase the amount of training data for NLP tasks in a low-resource context. However, the best strategy to decide which cross-lingual data to include is unclear. Prior research often focuses on a small set of languages from a few language families and/or a single task. It is still an open question how these findings extend to a wider variety of languages and tasks. In this work, we analyze cross-lingual transfer for 263 languages from a wide variety of language families. Moreover, we include three popular NLP tasks: POS tagging, dependency parsing, and topic classification. 
Our findings indicate that the effect of linguistic similarity on transfer performance depends on a range of factors: the NLP task, the (mono- or multilingual) input representations, and the definition of linguistic similarity.
\end{abstract}

\blfootnote{*Work done while interning at Apple.}

\section{Introduction}
\label{sec:introduction}

For many of the world's languages, the available data to train natural language processing (NLP) models is scarce. If data is available, it is often only enough for an evaluation set, raising the question of how to select the training set. Two approaches are intuitive: \begin{enumerate*}[label=(\roman*)]
    \item based on linguistic similarity measures: find the training data in a language that is linguistically as close as possible to the target language, where ``linguistically close'' is defined by one or multiple linguistic similarity measures, or
    \item based on dataset dependent measures: find a dataset in another language that is similar to the target dataset, e.g., in terms of high word or $n$-gram overlap.
\end{enumerate*} Naturally, these two types of approaches are not mutually exclusive.

\begin{figure}[t]
    \centering
    \includegraphics[width=\linewidth]{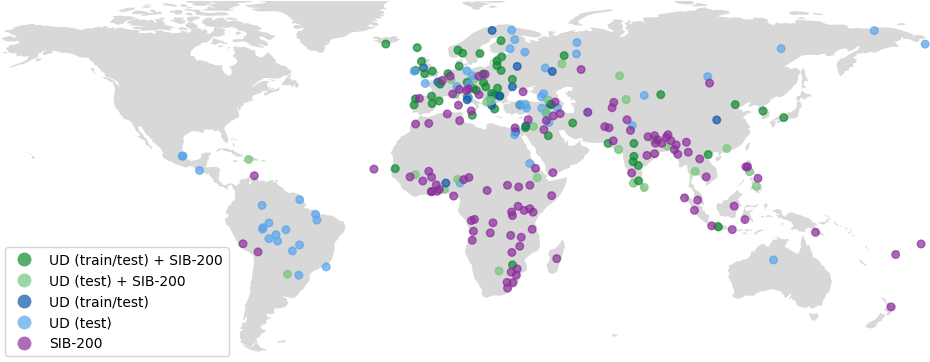}
    \caption{\textbf{Languages included in our experiments.} Green indicates languages included in all tasks, blue languages only used for POS tagging and parsing, and purple languages only used for topic classification.
    Base map via naturalearthdata.com (CC0).
    }
    \label{fig:map-languages-datasets}
\vspace{-1em}
\end{figure}

However, it is unclear which of these measures are most important to select the source language for cross-lingual transfer, despite previous work focusing on this question 
(see \S\ref{sec:related-work}): 
earlier studies often contradict each other, and therefore leave a number of important avenues for improvement. For example, prior work often lacks a large representation of languages and language families,
as well as NLP tasks,
and sometimes relies on synthetically constructed datasets. 
Furthermore, popular out-of-the-box measures for linguistic similarity are sometimes intransparent or faulty (\citealp{toossi-etal-2024-reproducibility}; \citealp{khan-etal-2025-uriel}; \S\ref{sec:methodology_similarity_metrics}), and correlations between different similarity measures are often not taken into account. Summarizing, how findings from prior work generalize across a larger variety of languages and tasks, and how we should interpret different similarity scores, are open questions.

In this work, we contribute to these questions by analyzing transfer between 263 different languages from 33 language families (Figure~\ref{fig:map-languages-datasets}) in three different tasks: \begin{enumerate*}[label=(\roman*)]
    \item part-of-speech (POS) tagging,
    \item dependency parsing, and
    \item topic classification.
\end{enumerate*}
In choosing our tasks, we are motivated by the following considerations: \begin{enumerate*}[label=(\roman*)]
\item availability of datasets: we select tasks for which we can find data for a multitude of languages from different language families, and
\item types of tasks: we include two word-level grammatical tasks (i.e., POS tagging and dependency parsing) and one sentence-level topic identification task. 
\end{enumerate*} 
To allow for a clean analysis, we opt for a zero-shot approach, in line with prior work~\cite{de-vries-etal-2022-make}
-- we train our models on the task in one particular source language, and evaluate on the target language without additional training or fine-tuning in the target language.

Our \textbf{key findings} can be summarized as:
\begin{enumerate}[leftmargin=*,nosep]
    \item For different tasks and input representations, different similarity measures are most predictive for cross-lingual transfer performance.
    For instance, syntactic similarity is most predictive for POS tagging and parsing, whereas trigram overlap is the most important predictor for $n$-gram-based topic classification~(\S\ref{sec:results-similarity-measures});
    \item \textbf{Practical implication \textit{within} studied tasks:}
    Choosing a source language based on a pertinent similarity measure leads to adequate transfer results~(\S\ref{sec:results-choosing-most-similar});
    \item \textbf{Practical implication \textit{across} studied tasks:}
    When no information about pertinent similarity measures is available for a given task, choosing a source language based on findings for 
    a conceptually
    \textit{similar} experiment is a relatively safe choice~(\S\ref{sec:results-choosing-based-on-experiment}).
\end{enumerate}

\section{Related Work}
\label{sec:related-work}

Table~\ref{tab:related-work} provides an overview of related work, highlighting an apparent trade-off between including many languages or including multiple tasks.
An important difference that distinguishes our work from prior work is the focus on both. 
We argue that including a large number of both source and target languages is important for ensuring a relatively balanced distribution of languages.%
\footnote{I.e., such that there is not mainly one cluster of (European) languages that are similar to each other, and most other languages are both dissimilar to that cluster and to each other.}
In comparison to the studies that come closest to our work in terms of number of included source and target languages~\cite{de-vries-etal-2022-make, samardzic-etal-2022-language}, we include 
a larger variety of 
tasks, allowing us to draw conclusions across task boundaries. 

\citet{philippy-etal-2023-towards} survey contributing factors for cross-lingual transfer, including most of the works in Table~\ref{tab:related-work}.
An important take-away from their work is that prior research presents contradicting findings. 
E.g.,
some studies find lexical overlap to correlate more strongly with token rather than sentence classification tasks \cite{srinivasan2021predicting}, whereas others find the opposite \cite{ahuja-etal-2022-multi}. 
Similar contradictory examples are presented for different linguistic similarity metrics, pre-training dataset size, and model architecture. 

Based on this insight, \citeauthor{philippy-etal-2023-towards} make 
recommendations for follow-up work: \begin{enumerate*}[label=(\roman*)]
    \item focus on real natural languages 
    (instead of
    synthetic ones),
    \item examine the interaction between different contributing factors,
    \item focus on many languages, 
    \item focus on linguistic features when selecting training languages, and
    \item focus on generative tasks, given the success of generative models.
\end{enumerate*}
We focus on the first four recommendations in this work.
We include three classification tasks (POS tagging, dependency parsing, and topic classification), motivated by the availability of train and test data in a large number of languages for these tasks.

\begin{table}
\adjustbox{max width=\columnwidth}{%
\begin{tabular}{@{}l@{~}l@{}r@{}c@{}l@{}}
\toprule
Work & \# Tasks &  \multicolumn{3}{c}{\# Langs per task} \\
&& (source & $\times$ & target)\\
\midrule
\citet{de-vries-etal-2022-make}& 1 (P) & 65& $\times$ & 105 \\
\citet{rice2025untangling} & {1 (P)} & {18} & $\times$ & {21} \\
\citet{samardzic-etal-2022-language} & 1 (D) & 47& $\times$ & 62 \\
\citet{adelani-etal-2022-masakhaner} & 1 (N) & 42& $\times$ & 42 \\
\citet{adelani-etal-2024-sib} & 1 (T) & 4& $\times$ & 197 \\
\citet{pires-etal-2019-multilingual} & 2 (N\,P) & \{4--41\}& $\times$ & \{4--41\} \\
\citet{muller2023languages} & 3 (F\,N\,Q) & \{7--9\}& $\times$ & \{7--9\} \\
\citet{srinivasan2021predicting} & 3 (F\,N\,P) & \{15--40\} & $\times$ & \{15--40\}\\
\citet{lin-etal-2024-mplm} & {4 (I\,N\,P\,T)} & {6} & $\times$ & {\{44--130\}}\\
\citet{lin-etal-2019-choosing} & 4 (D\,E\,M\,P) & \{30--60\}& $\times$ & \{9--54\} \\
\citet{xia-etal-2020-predicting} & 4 (D\,E\,M\,P) & \{30--60\}& $\times$ & \{9--54\} \\
\citet{lauscher-etal-2020-zero} & 5 (D\,F\,N\,P\,Q) & 1& $\times$ & \{8--14\} \\
\citet{ahuja-etal-2022-multi} & 6 (E\,F\,N\,P\,Q\,S) & 1& $\times$ & \{7--48\} \\
\midrule
This work & 3 (D\,P\,T) & 70& $\times$ & 153 (D\,P) \\
&& 194& $\times$ & 194 (T) \\
\bottomrule
\end{tabular}}
\caption{%
\textbf{Related work} focusing on zero-shot transfer between many languages or on many tasks.
Tasks: 
D=dependency parsing, 
E=entity linking, 
F=natural language inference, 
I=intent classification,
M=machine translation, 
N=named entity recognition, 
P=part-of-speech tagging, 
Q=question answering,
S=sentence retrieval,
T=topic classification.
}
\label{tab:related-work}
\vspace{-1em}
\end{table}

\section{Methodology}
\label{sec:methodology}

We run transfer experiments on two word-level grammatical tasks with word-level annotations (\S\ref{sec:methodology_grammatical_tasks}) and a sentence-level topic classification task (\S\ref{sec:methodology_semantic_task}). 
Of the 263 languages in our experiments, 55 training and 84 test languages are shared between all tasks.
Appendix~\S\ref{sec:appendix-languages} contains details on all languages and datasets in this study.
Section~\S\ref{sec:methodology_similarity_metrics} introduces our similarity measures.

\subsection{Grammatical tasks}
\label{sec:methodology_grammatical_tasks}

\paragraph{Data}
We use POS tags and syntactic dependencies from Universal Dependencies (UD; \citealp{de-marneffe-etal-2021-universal}).
The advantage of UD is the large number of languages (153) and language families (28) in which manually annotated datasets are available that both showcase a language's specific syntactic structures and adhere to a shared set of annotation guidelines.
UD also has its drawbacks in that the different treebanks are not necessarily from the same domains, most treebanks were annotated independently by different groups of researchers, and the size of the train and test splits is not identical across treebanks. 
To mitigate the latter, we evaluate the effect of training dataset size in \S\ref{sec:results-similarity-measures}.
To account for differences in the test sets (e.g., differing complexities or sentence lengths), we focus on comparing the performance of different parsers/\allowbreak{}taggers on each test set (rather than comparing how well each parser/\allowbreak{}tagger does across test sets; \S\ref{sec:results-similarity-measures}).

We use the test splits of UD release (2.14; \citealp{ud214}): 268 treebanks in 153 target languages, from 28 language families. 
We use pretrained models (see below), trained on 124 treebanks in 70 source languages from 12 language families. 
We exclude treebanks that are without text, particularly small, focus on code-switching, or have inconsistent train/\allowbreak{}test splits (\S\ref{sec:appendix-excluded-treebanks}).

\paragraph{Models}
We use the UDPipe~2 models \citep{straka-2018-udpipe, udpipe212} that were trained on UD~2.12 data.
Each model is trained on a single treebank.
The models have so far only been systematically evaluated on their within-treebank performance, but not cross-lingually.
UDPipe~2 combines monolingual character and word embeddings
with multilingual
embeddings derived from mBERT \cite{devlin-etal-2019-bert}.
The models are trained jointly for POS tagging, dependency parsing, lemmatization and morphological feature prediction. 
We only evaluate on the first two, as many %
treebanks do not have gold standard labels for the others.
UDPipe~2 post-processes the predicted dependencies to ensure that each sentence includes a root dependency that all other nodes (in)directly depend on.

We evaluate POS tagging using accuracy, and dependency parsing using the labeled (LAS) and unlabeled attachment scores (UAS).
For LAS, we follow UD's evaluation scripts and ignore dependency label subtypes.

\subsection{Topic classification}
\label{sec:methodology_semantic_task}

\paragraph{Data}
We use the SIB-200 dataset \citep{adelani-etal-2024-sib}, a subset of FLORES-200 \citep{nllb-22} with parallel sentences in 194 languages\footnote{This excludes three Arabic varieties whose sentences are nearly identical to the Standard Arabic sentences (\S\ref{sec:appendix-excluded-sib200}).} (from 22 families) annotated with seven topic labels.
Eight languages are represented twice, but with different writing systems.
SIB-200 contains 701 training and 204 test sentences.

\paragraph{Models}
We use multi-layer perceptrons (MLPs) for topic identification, similar to the baseline models by \citet{adelani-etal-2024-sib}.
This lightweight architecture allows training and evaluating many models without prohibitive time or energy investments, and results in evaluation scores that are close to the performance of base-sized transformers like XLM-R \cite{conneau-etal-2020-unsupervised} (\S\ref{sec:appendix-results-and-heatmaps}, Table~\ref{tab:intra-vs-crosslingual-results}).
We use the scikit-learn implementation \citep{scikit-learn} and conduct hyperparameter tuning on a subset of the languages (details in Appendix~\S\ref{sec:appendix-topic-models}).
We compare different ways of representing the input data:
\begin{enumerate}[leftmargin=*,nosep]
    \item \textbf{Character $n$-gram counts} (\topicsbase).
We use character-level $n$-gram counts ($1$- to $4$-grams) to represent the input. 
This ensures that we know exactly what training data were used, and it puts all languages on equal footing.
    \item \textbf{Transliterated input} (\topicstranslit).
To remove differences between writing systems, we use uroman \cite{hermjakob-etal-2018-box} to transliterate the dataset into Latin characters and remove diacritics, and otherwise repeat the previous set-up. We exclude three datasets that were not supported by uroman (\S\ref{sec:appendix-excluded-sib200}).
    \item \textbf{Multilingual representations} (\topicsmbert). To allow for a more direct comparison with the grammatical experiments, which partially rely on multilingual mBERT representations, we follow UDPipe~2 by deriving embeddings from the mean-pooled last four layers of the frozen base-sized, uncased mBERT model \cite{devlin-etal-2019-bert}. 
We use the hidden representations of the \texttt{[CLS]} token as input to the MLP.
\end{enumerate}

\subsection{Similarity measures}
\label{sec:methodology_similarity_metrics}

We include a range of dataset-dependent and \mbox{-independent} similarity measures, which are similar to measures used in related work (\S\ref{sec:related-work}).

\paragraph{Structural similarities}

Grambank \cite{grambank_release} encodes grammatical information for several thousand languages.
Its 195 grammatical features are chosen to allow almost no logical dependencies between the values of different features, i.e.\ the value of one feature does not logically entail the value of another.

We additionally use the lang2vec tool \citep{littell-etal-2017-uriel, lang2vec_package}, which has also been used in many other studies on multilingual NLP 
(cf.\ \citealp{toossi-etal-2024-reproducibility}).
Lang2vec aggregates information on syntax, phonology, and phoneme inventories from multiple databases in the form of binary features.
Since not all sources contain full information on all languages,
we use language vectors that include information from multiple of the above sources (averaged values where sources disagree).
Some grammatical features are covered by both Grambank and lang2vec, although sometimes with different value assignments \cite{baylor-etal-2023-past}.

We use \citeposs{gower1971coefficient} coefficient to calculate similarities between language pairs by comparing their feature values.
If information is missing for a feature in one or both languages, we ignore that feature.
If two values are identical, their similarity is 1, otherwise it is 0.
The overall distance between the two languages is the mean of the (attested) feature distances.
We calculate similarity scores for Grambank's entries (\textbf{\grambank}) and for lang2vec's syntactic (\textbf{\syntax}), phonological (\textbf{\phono}), and phonetic (\textbf{\inventory}) entries.
We include only similarities for language pairs where at least half of all features could be included in the calculations. However, in practice, more features are compared in most cases. On average, 83\% of the features are included in each similarity calculation for \grambank, 63\% for \syntax, 100\% for \inventory, and 84\% for \phono.

\paragraph{Lexical similarity}
As a proxy for how similar different languages' vocabularies are,
we compare multilingual word lists from the Automated Similarity Judgment Program (ASJP; \citealp{asjp}).
\citet{jaeger2018extracting, jaeger2018global} calculated language dissimilarity scores based on ASJP, taking into account cross-linguistic phonological patterns.
Our lexical similarity score (\textbf{\asjp}) is 1 minus J\"{a}ger's dissimilarity score.
Since some languages have multiple ASJP entries, we use language-wise mean scores. 

\paragraph{Phylogenetic relatedness}

We determine whether two languages are related (and how closely) with the help of Glottolog's \cite{glottolog2024} phylogenetic trees.
For each language, we retrieve the path from its family root node to the language node. 
The relatedness of two languages (\textbf{\genetic}) is 
the ratio of shared nodes along these paths.%
\footnote{Because different trees/branches have different depths (owing to different family sizes and/or different levels of documentation), comparisons of scores can only serve as a proxy for the degree of phylogenetic relatedness. This is also noted in the metadata for lang2vec, which calculates genetic distances in a similar way: \url{http://www.cs.cmu.edu/~dmortens/projects/7_project/}.
{For constructed languages, we ignore the top-most level (e.g., ``artificial languages'') and instead retrieve phylogenetic information starting at the second level.}}

\paragraph{Geographic proximity}

We use lang2vec's location information (each language is represented as a vector of distances to a number of points on the Earth's surface) and calculate the Euclidean distances between language vectors.%
\footnote{We do not use lang2vec's precomputed distances, as the package erroneously returns the maximum distance for some languages, regardless of the comparison.}
We define geographic proximity (\textbf{\geo}) as 1 minus the distance.

\paragraph{Character and word overlap}

We measure the overlap between training and test datasets on the character level (\textbf{\overlapchar}) with the Jaccard similarity of the character sets.
We repeat this on the word level (\textbf{\overlapword}) for the UD datasets (where the data come with gold-standard word tokenization), and on the character trigram (\textbf{\overlaptri}) and mBERT subword token (\textbf{\overlapsubword}) level for SIB-200. %

\paragraph{Amount of training data}

We count the number of sentences in each training dataset (\textbf{\trainingsize}).
For the topic classification task, this is trivial as each language has the same number of training samples. %

\paragraph{Correlations between measures}
\label{sec:correlations-between-similarity-measures}

Correlations between similarity measures can influence how importance is assigned to different measures in regression analyses. %
Table~\ref{tab:correlations-between-similarity-measures} in \S\ref{sec:appendix-correlations-between-measures} shows how the different similarity measures are correlated with each other.
Lexical similarity and phylogenetic relatedness are highly correlated ($r=0.87$). %
Additionally, the two grammar-based similarity measures (\grambank{} and \syntax) are moderately strongly correlated ($r=0.61$), as are \grambank{} and lexical/phylogenetic similarity ($r=0.57$, $0.56$), and word overlap in UD and lexical similarity ($r=0.59$).

\section{Supporting Results}
\label{sec:supporting-results}

Here we present the results on which the main analyses in \S\ref{sec:main-results-analysis} build.
In~\S\ref{sec:results-within-task}, we present the POS tagging, parsing and topic classification scores within and across languages.
In~\S\ref{sec:results-across-task}, we compare how (dis)similar transfer patterns are across tasks and experiments, which helps contextualize the importance of different similarity measures for different experiments and tasks in~\S\ref{sec:results-similarity-measures}.

\subsection{Within-language vs.\ cross-lingual performance per task}
\label{sec:results-within-task}

As expected, the within-language performances are much higher than the cross-lingual performances.
Table~\ref{tab:intra-vs-crosslingual-results} shows the mean scores within and across languages and datasets.
In the following analyses, we focus on the language-level results.
For UD, which has nearly twice as many datasets as languages, we find that trends are very stable across datasets of the same language~(\S\ref{sec:appendix-robustness}).

All of our models achieve reasonable within-language results, indicating that cross-lingual transfer (to an appropriate target language) could be feasible for all models.
For the two straightforward classification tasks (POS tagging and topic classification), we construct random baselines~(\S\ref{sec:appendix-random-baselines}) that are outperformed by all models in within-language evaluations.
For our topic classification models, the within-dataset performance is comparable to the baselines in the dataset paper \citep{adelani-etal-2024-sib}, but we do not match the performance of their best model.
Our POS tagging results are similar but not identical to the ones that \citet{de-vries-etal-2022-make} obtained on UD v2.8 with fine-tuned XLM-R models for 65 training and 105 test languages.
If we consider our own POS tagging results but only select the subset of languages that was used in de \citeauthor{de-vries-etal-2022-make}'s experiments, the POS tagging accuracies are highly correlated with those that \citeauthor{de-vries-etal-2022-make} obtained (Pearson's $r$ and Spearman's $\rho=0.73$, $p<0.0001$). 
Thus while the model choice (UDPipe vs.\ XLM-R) matters for the resulting patterns, many transfer trends are similar.

\begin{table}
\centering
\adjustbox{max width=\linewidth}{%
\begin{tabular}{%
@{}l@{~~}%
r@{}r@{\hspace{5pt}}%
r@{\hspace{-2pt}}r%
r@{}r@{\hspace{5pt}}%
r@{}r@{}%
}
\toprule
 & \multicolumn{4}{c}{Dataset level} & \multicolumn{4}{c}{Language level} \\
 & \multicolumn{2}{@{}c}{Within} & \multicolumn{2}{c}{Across} & \multicolumn{4}{c}{Within~~Across}  \\
\midrule
\multicolumn{9}{@{}l@{}}{\textit{Grammatical tasks}} \\
POS accuracy  & 96.4 & \stdev{3.1} & 43.6 & \stdev{20.8} & 93.4 & \stdev{6.8} & 43.0 & \stdev{20.2} \\
POS acc.\ (\citeauthor{de-vries-etal-2022-make}) & 94.1\rlap{\smallasterisk} & \stdev{4.5} & 57.4\rlap{\smallasterisk} & \stdev{22.4} & 94.1 & \stdev{4.5} & 57.4 & \stdev{22.4} \\
UAS  & 88.7 & \stdev{6.4} & 37.3 & \stdev{19.3} & 84.8 & \stdev{10.1} & 36.8 & \stdev{18.8} \\
LAS  & 84.6 & \stdev{8.6} & 21.2 & \stdev{17.9} & 78.4 & \stdev{13.7} & 20.6 & \stdev{17.0} \\
\midrule
\multicolumn{9}{@{}l}{\textit{Topic classification (accuracy)}} \\
\topicsbase{}  & 70.3 & \stdev{4.0} & 20.7 & \stdev{8.8} & 66.7 & \stdev{14.1} & 20.1 & \stdev{8.8} \\
\topicstranslit{}  & 69.4 & \stdev{4.1} & 22.5 & \stdev{8.2} & 66.7 & \stdev{10.8} & 22.5 & \stdev{8.2} \\
\topicsmbert{}  & 60.1 & \stdev{18.7} & 42.6 & \stdev{20.4} & 58.2 & \stdev{19.9} & 42.6 & \stdev{20.4} \\
MLP (\citeauthor{adelani-etal-2024-sib}) & 62.3 & \stdev{---} & --- & & --- & & --- & \\
XLM-R\textsubscript{B} (\citeauthor{adelani-etal-2024-sib}) & 70.9 & \stdev{---} & --- & & --- & & --- & \\
XLM-R\textsubscript{L} (\citeauthor{adelani-etal-2024-sib}) & 76.1 & \stdev{---} & --- & & --- & & --- & \\
\bottomrule
\end{tabular}
}
\caption{%
\textbf{Scores of models evaluation within vs.\ across datasets and languages,} in this work and related work.
Our analyses focus on the language-level scores.
Scores are mean scores (in~\%), with standard deviations in subscripts. 
XLM-R\textsubscript{B} and \textsubscript{L} = XLM-R base/large.
*When multiple datasets were available for one language, \citeauthor{de-vries-etal-2022-make} combined them into one.
}
\label{tab:intra-vs-crosslingual-results}
\end{table}

For topic classification, we compare the different input representations.
The \textit{within-language} performance is higher for the monolingual $n$-gram-based models than the models with multilingual input representations from mBERT.
We hypothesize that this due to high \texttt{[UNK]} token rates for some languages that were not in mBERT's pre-training data.\footnote{\citet{bagheri-nezhad-agrawal-2024-drives} similarly find that the performance of multilingual models that are monolingually trained and evaluated on SIB-200 relates to how well a language is represented in a model's pre-training data.}

Although \topicsbase{} and \topicstranslit{} achieve the same within-language accuracy, \topicstranslit{} performs slightly better {cross-lingually}, having the advantage of higher $n$-gram overlap between datasets.%
\footnote{We further analyze the effect of writing systems in~\S\ref{sec:results-writing-systems}. In our correlation analysis~(\S\ref{sec:results-similarity-measures}), (lack of) character overlap serves as a proxy for writing system differences.}
In cross-lingual evaluations, \topicsmbert{} benefits from its multilingual input embeddings: its accuracy is about twice as high (42.6\%) as for the monolingual models.
Generally, transfer works well if both training and test languages are in mBERT's pre-training data (or closely related to a language that is): 
The average accuracy is 68.3\% if both languages were included in mBERT's pre-training data, 46.5\% if only the target language was included, 36.1\% if only the source language was included, and 33.2\% if neither was included.
This is consistent with results by \citet{adelani-etal-2024-sib}, who fine-tuned four XLM-R\textsubscript{large} models on one high-resource language each and found their transfer results to be very similar to each other.

\begin{figure}
    \centering
    \footnotesize
    \textit{POS accuracy \hspace{3mm} Dependency UAS \hspace{1mm} Dependency LAS \hfill\phantom{.}}\\
    \includegraphics[width=0.28\linewidth]{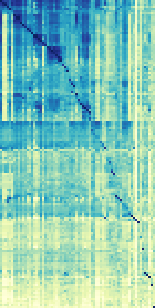}
    \hfill
    \includegraphics[width=0.28\linewidth]{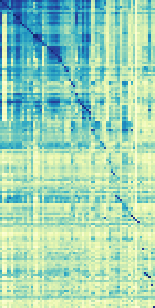}
    \hfill
    \includegraphics[width=0.28\linewidth]{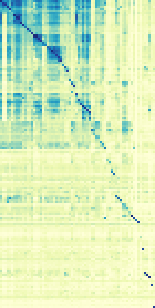}
    \hfill
    \includegraphics[width=0.085\linewidth]{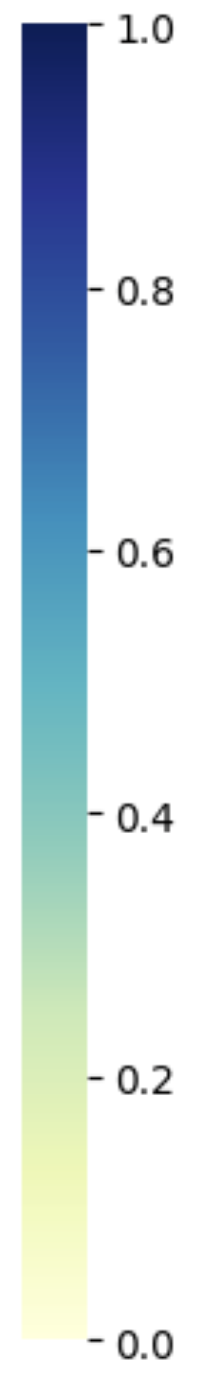}\\
    \vspace{0.3\baselineskip}
    \textit{\topicsbase\hspace{9mm}\topicstranslit\hfill\topicsmbert}\hspace{5mm}\phantom{.}\\
    \includegraphics[width=0.32\linewidth]{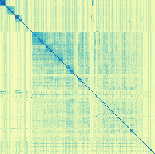}
    \hfill
    \includegraphics[width=0.32\linewidth]{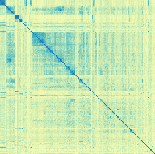}
    \hfill
    \includegraphics[width=0.32\linewidth]{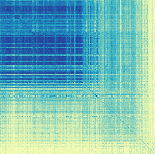}%
    \caption{\textbf{Different experiments produce different transfer patterns.} NLP transfer results for all combinations of training (columns) and test languages (rows).
    The darker a cell, the higher the score.
    The three heatmaps for the grammatical tasks are sorted in the same order, and the three heatmaps with the topic classification results are sorted in the same order.
    The darker diagonal shows the within-language scores.
    Large, labelled heatmaps are in Appendix~\S\ref{sec:appendix-heatmaps}.}
    \label{fig:heatmaps-intext}
\end{figure}

\subsection{Comparing transfer patterns across tasks}
\label{sec:results-across-task}

In Figure~\ref{fig:heatmaps-intext}, we plot heatmaps of the transfer results for the different experiments.
Darker colours correspond to higher scores.
To save space, we omit the language labels (which can be found in \S\ref{sec:appendix-heatmaps}).
We compute correlations between the results, which we summarize below (details in \S\ref{sec:appendix-correlation-tables-tasks}).

\textbf{The results of the grammatical tasks are highly correlated with each other, the correlations between other task results are weaker.}%
\footnote{Because of the very high correlation between the parsing measures LAS and UAS (Pearson's $r = 0.95$), 
we focus on LAS in the remainder of the paper as the more extensive of the two: LAS not only measures whether the dependency arcs are correct, but also whether they were labelled correctly.
}
Even when using \textit{exactly} the same underlying text data and models, evaluating on a related task results in transfer trends that are similar, but not entirely the same: 
The parsing (LAS) and POS tagging results %
by identical UDPipe models (except for the classification heads) are correlated with $r=0.86$.

The preprocessing of the input data (e.g., transliteration) and especially the choice of input representation (mono- vs.\ multilingual) affects the transfer trends: %
The results of the two $n$-gram-based topic classifiers are highly correlated with each other ($r=0.68$), but not with the results of the mBERT-based set-up ($r=0.28$ and $0.36$).
Set-ups that involve multilingual representations are more highly correlated.
For the languages that appear both in UD and SIB-200,
the results of the grammatical tasks are most strongly correlated with those of the mBERT-based topic classification models ($r=0.64$ for POS tagging, $r=0.58$ for LAS).

\section{Main Results and Analysis}
\label{sec:main-results-analysis}

Here, we investigate which factors correlate with overall transfer performance~(\S\ref{sec:results-similarity-measures}).
Then, we explore what this means for selecting a source language for cross-lingual transfer~(\S\ref{sec:results-practical-implications}).

\subsection{How do the similarity measures correlate with the tasks and input representations?}
\label{sec:results-similarity-measures}

We calculate the correlations between the similarity measures and the NLP task results.
To account for differences between test sets, we calculate correlations (Pearson's~\textit{r}) between similarity measures and transfer results on a test language level.
We compare the performance of different NLP systems on the same test language, but we do \textit{not} compare the performance of a single system across multiple test sets.
This decision is motivated especially by the challenges of comparing parsing performance across treebanks:
Sentence length influences parsing difficulty as it determines the available search space (cf.\ \citealp{mcdonald-nivre-2011-analyzing}, \citealp{choi-etal-2015-depends}), and morphological differences between languages hamper the comparability of parsers \cite{nivre-fang-2017-universal}.

We analyze overall correlation scores by averaging across test languages.
We treat correlation coefficients with \textit{p}-values of at least 0.05 as 0, and exclude items where we could not calculate similarity scores due to missing linguistic information.
We use language-level averages
so that those languages with multiple training and test datasets do not have artificially high correlations,
and languages with multiple test datasets do not have greater influence on the overall averages.
However, if we instead consider treebank-level correlations, we observe that test sets that belong to the same language show very similar correlation patterns and scores~{(\S\ref{sec:appendix-robustness})}.

\begin{figure}
    \centering
    \includegraphics[width=\linewidth]{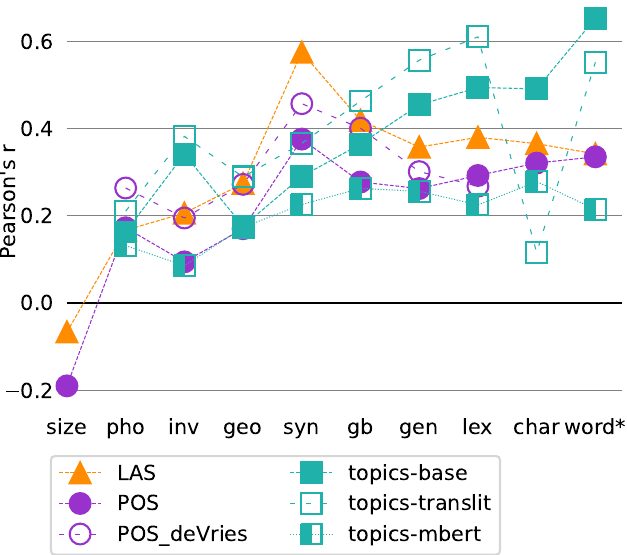}
    \caption{\textbf{Mean correlation scores between task results and similarity measures.} 
    ``Word*'' = overlap between words (\overlapword; UD tasks) trigrams (\overlaptri; \topicsbase{}/\texttt{translit}), and subword tokens (\overlapsubword; \topicsmbert).
    Dotted %
    lines are added for intelligibility.}
    \label{fig:correlations-rows}
\vspace{-1em}
\end{figure}

The correlations between task results and similarity measures vary across our experiments.
Figure~\ref{fig:correlations-rows} shows the correlations, which we summarize below.
Unaggregated correlations and 95\% confidence intervals of the mean scores can be found in \S\ref{sec:appendix-correlation-tables-similarity-and-tasks}.
These unaggregated correlations show that, while there are some outliers, the overall trends we report below accurately reflect the language-level trends.

We observe some \textbf{common trends across experiments:}
Training dataset size does not matter much: 
it
is the same across all topic classification experiments, and the correlations between training set size and performance in the grammatical tasks are close to zero. %
The phonological and phonetic similarity measures (\phono, \inventory) also generally have low correlation scores.%
\footnote{%
The correlations with \inventory{} are comparatively high for the $n$-gram-based topic classification models.
We assume that this is due to \inventory{} being moderately correlated with string similarity measures like trigram overlap~(\S\ref{sec:appendix-correlations-between-measures}), which is in turn strongly correlated with the topic classification results.}

\textbf{The strongest predictor for parsing performance is syntactic similarity {\normalfont{}(\syntax) as determined by lang2vec (\ravg{\syntax}$=0.57$),} followed by the similarity of Grambank features {\normalfont{}(\grambank; \ravg{\grambank}$=0.42$).}}
These correlations are even stronger when we only consider the test languages that mBERT was pretrained on (\ravg{\syntax}$=0.69$ and \ravg{\grambank}$=0.60$, respectively; Appendix~\S\ref{sec:appendix-correlation-tables-similarity-and-tasks}).

\textbf{The POS tagging outputs show correlation patterns similar to the ones for parsing, albeit weaker.}
Although the correlation strengths are generally weaker for POS tagging, e.g., the strongest predictor is lang2vec's \textbf{syntactic similarity} (\syntax) with \ravg{\syntax}$=0.37$, \textbf{word overlap} (\overlapword) is as good a predictor for POS accuracy as for LAS (\ravg{\overlapword}=0.33 and \ravg{\overlapword}=0.34).
Again, the correlation strengths are higher when only considering test languages that were in mBERT's pre-training data (\ravg{\syntax}$=0.46$).\footnote{%
The correlation strengths are also higher when we analyze \citeauthor{de-vries-etal-2022-make}'s results (\ravg{\syntax}$=0.46$). 
We hypothesize that this is due to the smaller amount of test languages in their experiments, a higher proportion of which is in XLM-R's pre-training data.}
The correlations we observe for the grammatical task's transfer results partially align with prior research: \citet{lauscher-etal-2020-zero}, \citet{samardzic-etal-2022-language} and \citet{pires-etal-2019-multilingual} also find syntactic similarity to be important, and \citet{lin-etal-2019-choosing} and \citet{xia-etal-2020-predicting} also find word overlap to be a good predictor for POS tagging.
However, contrary to our results, the latter two find syntactic similarity to be relatively unimportant for both tasks. 
We hypothesize that such inconsistencies are due to differences in the language sets studied and other experimental differences (e.g., \citet{lin-etal-2019-choosing} add data from the test language to the training set when possible).

\textbf{The results of the $n$-gram-based models are most highly correlated with measures of string similarity and lexical similarity.}
This is expected based on their input representations.
For \topicsbase, the \textbf{trigram overlap} (\overlaptri) between training and test data shows the highest correlation (\ravg{\overlaptri}=0.65), followed by \textbf{character overlap} (\overlapchar; \ravg{\overlapchar}=0.49) and \textbf{lexical similarity} (\asjp; \ravg{\asjp}=0.49). 
Trigram overlap and lexical similarity (and the highly correlated genetic relatedness, \genetic) are also the strongest predictors for \topicstranslit{} (\ravg{\overlaptri}=0.55, \ravg{\asjp}=0.61, \ravg{\genetic}=0.55).
However, for this model, character overlap only plays a very small role (\ravg{\overlapchar}=0.11), as all datasets use the same character inventory due to the transliteration.

\textbf{For the model using mBERT representations (\topicsmbert), none of the correlations are strong.}
The correlations peak at \ravg{\overlapchar}=0.27 for character overlap (\overlapchar).
Model performance depends instead on the inclusion of the source and especially target languages in mBERT's pre-training data (\S\ref{sec:results-within-task}).

\textbf{The transfer results cannot be predicted by any one factor alone.}
We fit a linear mixed effects model for each experiment, with the NLP score as the dependent variable and the source and target languages as random effects. The fixed effects are the similarity scores, whether the training and test data use the same writing systems, and whether the test language was included in mBERT's pre-training data.
We use R \cite{r2024} and the lme4 package \cite{bates2015fitting}.
This analysis 
involves
a reduced set of languages, as for many language pairs in our data at least one similarity metric is not defined and we do not perform any data imputation.
The analysis shows similar trends to the correlations described above. 
Additionally, 
for each experiment, most of the variables 
in our analyses are significant predictors of the transfer score (details in \S\ref{sec:appendix-correlation-tables-mixed-effects}). 
This even sometimes applies to measures that capture similarity at the same 
linguistic level: e.g., both grammatical similarity measures (\syntax, \grambank) are independently among the highest predictors for parsing performance.
The finding that multiple measures contribute to predicting the transfer score confirms previous work \cite{lin-etal-2019-choosing, de-vries-etal-2022-make}.

\subsection{Practical implications}
\label{sec:results-practical-implications}

So far we have investigated correlations on a global level. In this section we investigate how these findings can be used in practice: Do the overall correlation patterns also apply when picking a source language for a given target language according to simple heuristics derived from the global patterns? %

We intentionally pick very simple heuristics, as we believe them to be more realistic to how practitioners choose source language candidates in practice.
Additionally, we choose them to be easily generalizable and not be constrained to the languages in our experiments.%
\footnote{For instance, while LangRank \cite{lin-etal-2019-choosing} is an easy-to-use tool for transfer recommendations, it is limited to the tasks and languages it was trained on and thus recommends one of 30 possible source languages for dependency parsing or one of 60 for POS tagging.}

\begin{table}
\setlength{\tabcolsep}{1pt} %
\centering
\adjustbox{max width=\linewidth}{%
\begin{tabular}{@{}lrrrrrrrrrr@{}}
\toprule
 & \trainingsize & \phono & \inventory & \geo & \syntax & \grambank & \genetic & \asjp & \overlapchar & word\rlap{*} \\
\midrule
\multicolumn{11}{@{}l@{}}{\textit{Top-1 source candidate (= most similar language)}} \\
\midrule
POS \rlap{deV} & --- & \cellcolor[HTML]{fde2d2} 8\stdev{8} & \cellcolor[HTML]{fddbc7} 10\stdev{11} & \cellcolor[HTML]{fddbc7} 10\stdev{11} & \cellcolor[HTML]{fee6d9} 7\stdev{8} & \cellcolor[HTML]{feede4} 5\stdev{7} & \cellcolor[HTML]{fde2d2} 8\stdev{11} & \cellcolor[HTML]{fee9dd} 6\stdev{8} & --- & --- \\
POS & \cellcolor[HTML]{d86551} 29\stdev{12} & \cellcolor[HTML]{f8bfa4} 15\stdev{13} & \cellcolor[HTML]{f9c6ac} 14\stdev{12} & \cellcolor[HTML]{f8bfa4} 15\stdev{15} & \cellcolor[HTML]{fddbc7} 10\stdev{12} & \cellcolor[HTML]{fbd0b9} 12\stdev{12} & \cellcolor[HTML]{fddecb} \phantom{0}9\stdev{11} & \cellcolor[HTML]{fddbc7} 10\stdev{10} & \cellcolor[HTML]{f8bfa4} 15\stdev{14} & \cellcolor[HTML]{fbd0b9} 12\stdev{13} \\
LAS & \cellcolor[HTML]{f19e7d} 21\stdev{14} & \cellcolor[HTML]{facab1} 13\stdev{12} & \cellcolor[HTML]{facab1} 13\stdev{13} & \cellcolor[HTML]{facab1} 13\stdev{16} & \cellcolor[HTML]{fee6d9} 7\stdev{10} & \cellcolor[HTML]{fddbc7} 10\stdev{\phantom{0}9} & \cellcolor[HTML]{fde2d2} 8\stdev{10} & \cellcolor[HTML]{fde2d2} 8\stdev{\phantom{0}9} & \cellcolor[HTML]{f8bb9e} 16\stdev{16} & \cellcolor[HTML]{fcd5bf} 11\stdev{13} \\
UAS & \cellcolor[HTML]{de735c} 27\stdev{12} & \cellcolor[HTML]{f8bb9e} 16\stdev{13} & \cellcolor[HTML]{f8bfa4} 15\stdev{12} & \cellcolor[HTML]{f8bb9e} 16\stdev{15} & \cellcolor[HTML]{fde2d2} 8\stdev{\phantom{0}9} & \cellcolor[HTML]{facab1} 13\stdev{11} & \cellcolor[HTML]{fddecb} 9\stdev{10} & \cellcolor[HTML]{fddbc7} 10\stdev{11} & \cellcolor[HTML]{f7b596} 17\stdev{14} & \cellcolor[HTML]{f9c6ac} 14\stdev{15} \\
\texttt{top.-}\rlap{\texttt{b.}} & --- & \cellcolor[HTML]{f7b596} 17\stdev{12} & \cellcolor[HTML]{f7b596} 17\stdev{14} & \cellcolor[HTML]{facab1} 13\stdev{14} & \cellcolor[HTML]{f8bfa4} 15\stdev{12} & \cellcolor[HTML]{f9c6ac} 14\stdev{13} & \cellcolor[HTML]{fddecb} 9\stdev{12} & \cellcolor[HTML]{fddecb} 9\stdev{11} & \cellcolor[HTML]{facab1} 13\stdev{11} & \cellcolor[HTML]{fef0e8} 4\stdev{\phantom{0}5} \\
\texttt{top.-}\rlap{\texttt{tr.}} & --- & \cellcolor[HTML]{facab1} 13\stdev{10} & \cellcolor[HTML]{facab1} 13\stdev{11} & \cellcolor[HTML]{fcd5bf} 11\stdev{11} & \cellcolor[HTML]{fcd5bf} 11\stdev{\phantom{0}9} & \cellcolor[HTML]{fddbc7} 10\stdev{\phantom{0}9} & \cellcolor[HTML]{fee6d9} 7\stdev{\phantom{0}9} & \cellcolor[HTML]{fee6d9} 7\stdev{\phantom{0}8} & \cellcolor[HTML]{f3a481} 20\stdev{13} & \cellcolor[HTML]{fef4ef} 3\stdev{\phantom{0}4} \\
\texttt{top.-}\rlap{\texttt{m.}} & --- & \cellcolor[HTML]{fbd0b9} 12\stdev{13} & \cellcolor[HTML]{fcd5bf} 11\stdev{13} & \cellcolor[HTML]{fddbc7} 10\stdev{11} & \cellcolor[HTML]{fddecb} 9\stdev{10} & \cellcolor[HTML]{fde2d2} 8\stdev{\phantom{0}9} & \cellcolor[HTML]{fde2d2} 8\stdev{\phantom{0}9} & \cellcolor[HTML]{fde2d2} 8\stdev{\phantom{0}9} & \cellcolor[HTML]{fbd0b9} 12\stdev{10} & \cellcolor[HTML]{fddecb} 9\stdev{13} \\
\midrule
\multicolumn{8}{@{}l@{}}{\textit{Top-3 source candidates}} \\
\midrule
POS \rlap{deV} & --- & \cellcolor[HTML]{fef4ef} 3\stdev{4} & \cellcolor[HTML]{fef0e8} 4\stdev{5} & \cellcolor[HTML]{fef0e8} 4\stdev{5} & \cellcolor[HTML]{fff7f3} 2\stdev{5} & \cellcolor[HTML]{fff7f3} 2\stdev{4} & \cellcolor[HTML]{feede4} 5\stdev{8} & \cellcolor[HTML]{fff7f3} 2\stdev{4} & --- & --- \\
POS & \cellcolor[HTML]{e48066} 25\stdev{13} & \cellcolor[HTML]{fee6d9} 7\stdev{\phantom{0}8} & \cellcolor[HTML]{fee6d9} 7\stdev{\phantom{0}8} & \cellcolor[HTML]{feede4} 5\stdev{\phantom{0}7} & \cellcolor[HTML]{fef4ef} 3\stdev{\phantom{0}4} & \cellcolor[HTML]{fef0e8} 4\stdev{\phantom{0}6} & \cellcolor[HTML]{feede4} 5\stdev{\phantom{0}7} & \cellcolor[HTML]{fef0e8} 4\stdev{\phantom{0}6} & \cellcolor[HTML]{fee6d9} 7\stdev{\phantom{0}8} & \cellcolor[HTML]{feede4} 5\stdev{\phantom{0}7} \\
LAS & \cellcolor[HTML]{f6af8e} 18\stdev{14} & \cellcolor[HTML]{fde2d2} 8\stdev{\phantom{0}9} & \cellcolor[HTML]{fee6d9} 7\stdev{\phantom{0}9} & \cellcolor[HTML]{feede4} 5\stdev{\phantom{0}8} & \cellcolor[HTML]{fef4ef} 3\stdev{\phantom{0}4} & \cellcolor[HTML]{fef0e8} 4\stdev{\phantom{0}6} & \cellcolor[HTML]{fef0e8} 4\stdev{\phantom{0}6} & \cellcolor[HTML]{fef4ef} 3\stdev{\phantom{0}5} & \cellcolor[HTML]{fde2d2} 8\stdev{11} & \cellcolor[HTML]{fee9dd} 6\stdev{\phantom{0}9} \\
UAS & \cellcolor[HTML]{e8896c} 24\stdev{13} & \cellcolor[HTML]{fde2d2} 8\stdev{\phantom{0}8} & \cellcolor[HTML]{fee6d9} 7\stdev{\phantom{0}8} & \cellcolor[HTML]{fee9dd} 6\stdev{\phantom{0}8} & \cellcolor[HTML]{fef4ef} 3\stdev{\phantom{0}5} & \cellcolor[HTML]{feede4} 5\stdev{\phantom{0}6} & \cellcolor[HTML]{feede4} 5\stdev{\phantom{0}7} & \cellcolor[HTML]{feede4} 5\stdev{\phantom{0}7} & \cellcolor[HTML]{fddecb} 9\stdev{10} & \cellcolor[HTML]{fee6d9} 7\stdev{\phantom{0}9} \\
\texttt{top.-}\rlap{\texttt{b.}} & --- & \cellcolor[HTML]{fddbc7} 10\stdev{10} & \cellcolor[HTML]{fddecb} 9\stdev{11} & \cellcolor[HTML]{feede4} 5\stdev{\phantom{0}9} & \cellcolor[HTML]{fee9dd} 6\stdev{\phantom{0}7} & \cellcolor[HTML]{fee9dd} 6\stdev{\phantom{0}9} & \cellcolor[HTML]{feede4} 5\stdev{\phantom{0}7} & \cellcolor[HTML]{fef4ef} 3\stdev{\phantom{0}6} & \cellcolor[HTML]{fde2d2} 8\stdev{\phantom{0}9} & \cellcolor[HTML]{fff7f3} 2\stdev{\phantom{0}4} \\
\texttt{top.-}\rlap{\texttt{tr.}} & --- & \cellcolor[HTML]{fde2d2} 8\stdev{\phantom{0}8} & \cellcolor[HTML]{fee6d9} 7\stdev{\phantom{0}9} & \cellcolor[HTML]{fef0e8} 4\stdev{\phantom{0}8} & \cellcolor[HTML]{feede4} 5\stdev{\phantom{0}6} & \cellcolor[HTML]{feede4} 5\stdev{\phantom{0}6} & \cellcolor[HTML]{fef4ef} 3\stdev{\phantom{0}5} & \cellcolor[HTML]{fef4ef} 3\stdev{\phantom{0}5} & \cellcolor[HTML]{f9c6ac} 14\stdev{11} & \cellcolor[HTML]{fffbfa} 1\stdev{\phantom{0}3} \\
\texttt{top.-}\rlap{\texttt{m.}} & --- & \cellcolor[HTML]{fee9dd} 6\stdev{\phantom{0}7} & \cellcolor[HTML]{feede4} 5\stdev{\phantom{0}5} & \cellcolor[HTML]{fef0e8} 4\stdev{\phantom{0}5} & \cellcolor[HTML]{fef0e8} 4\stdev{\phantom{0}4} & \cellcolor[HTML]{fef0e8} 4\stdev{\phantom{0}5} & \cellcolor[HTML]{fef0e8} 4\stdev{\phantom{0}4} & \cellcolor[HTML]{fef0e8} 4\stdev{\phantom{0}5} & \cellcolor[HTML]{feede4} 5\stdev{\phantom{0}4} & \cellcolor[HTML]{fee9dd} 6\stdev{11} \\
\bottomrule
\end{tabular}
}
\caption{%
\textbf{Mean performance loss if picking source languages based solely on a given metric}
(or based on training dataset size).
Performance loss in percentage points.
Subscripts = standard deviations;
``word*'' = \overlapword{}, \overlaptri{}, \overlapsubword{}; deV = \citet{de-vries-etal-2022-make}.}
\label{tab:effect_of_language_choice_metric}
\end{table}

\subsubsection{Should I pick the most similar language according to one similarity measure?}
\label{sec:results-choosing-most-similar}

For each target language and each similarity measure, we calculate the difference between the best score obtained by any source language and the score obtained by the most similar language per the measure.
We only consider the source language candidates for which we can compute a similarity score for the target.
For \trainingsize, we select the language with the most training data. 
The patterns for the most strongly correlated similarity measures are similar --
\textbf{choosing a source language based on the pertinent similarity measure incurs relatively small losses.}
E.g., parsing performance is most strongly correlated with syntactic similarity, and picking the syntactically most similar language as the source also results in the lowest performance loss for parsing (Table~\ref{tab:effect_of_language_choice_metric}, top).
Our findings suggest choosing a syntactically similar language for POS tagging and dependency parsing, and a dataset with high trigram overlap for the $n$-gram-based topic classification experiments.

\begin{figure}
    \centering
    \includegraphics[width=\linewidth]{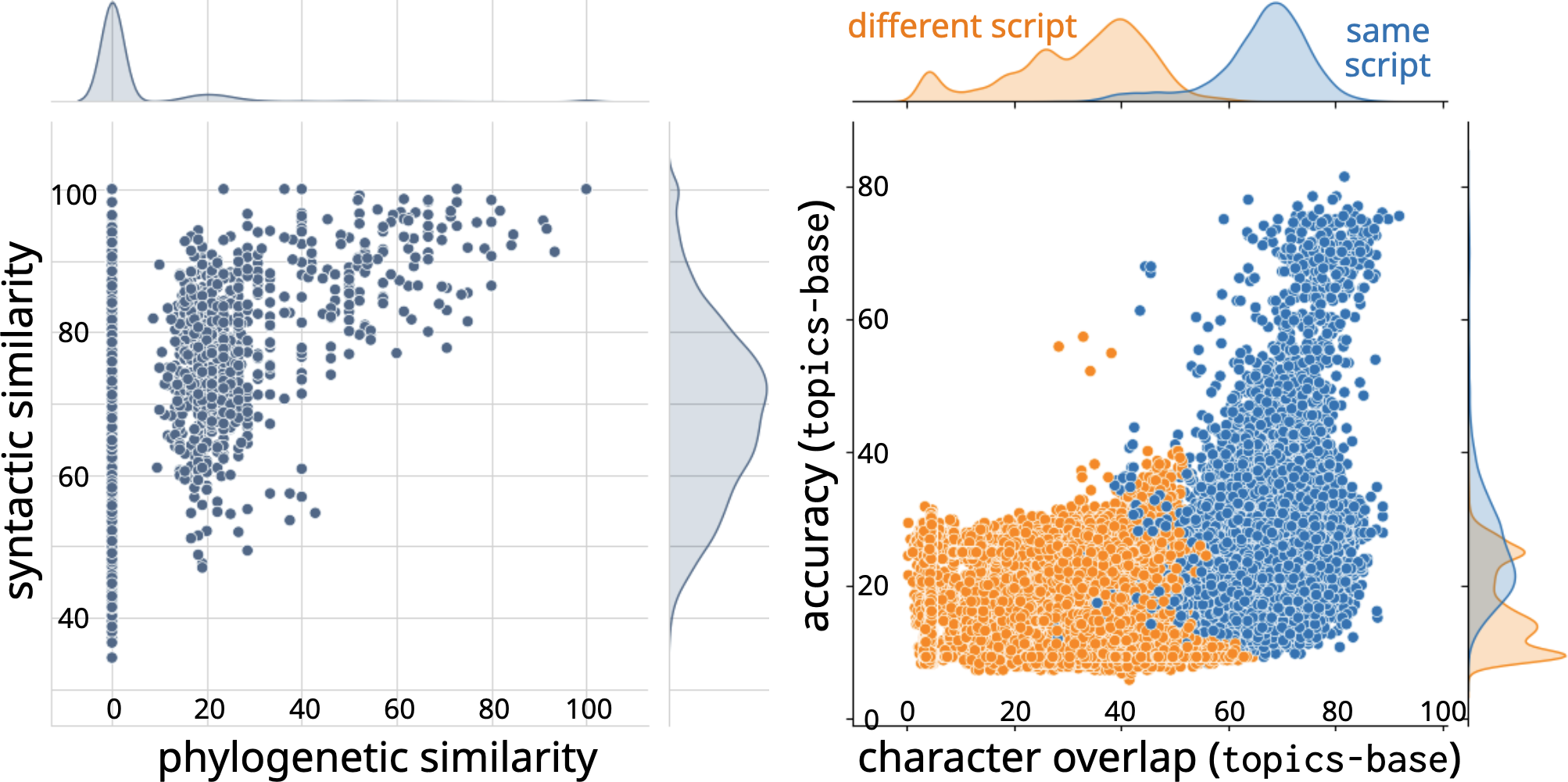}
    \caption{%
    \textbf{Left:} Relationship between phylogenetic and syntactic similarity -- unrelated or distantly related languages can be syntactically similar or dissimilar, but all closely related languages are syntactically similar.
    \textbf{Right:} Relationship between character overlap (between training and test sets) and the classification scores of the \topicsbase{} model -- transfer between languages with low character overlap works poorly, but high overlap does not guarantee good transfer.
    }
    \label{fig:scatterplots}
\end{figure}

However, some of the less strong predictors in the global correlations are nearly as good for picking a source language, e.g., genetic and lexical similarity for parsing. 
We hypothesize that this is due some linguistic measures being more correlated when similarity is high.\footnote{E.g., very closely related languages tend to be syntactically similar, even if the overall correlation between genetic and syntactic similarity is only moderately high due the existence of languages that are unrelated but share syntactic similarities or that are (more distantly) related and syntactically dissimilar 
(Figure~\ref{fig:scatterplots}, left).
}
Conversely, character overlap is a  worse measure for selecting a source language than the overall correlations would suggest.
This is likely due to transfer between languages with low character overlap generally working poorly, while high overlap does not guarantee good transfer (e.g.,  
Figure~\ref{fig:scatterplots}, right).

We additionally simulate a setup where a researcher has data in multiple source language candidates at their disposal and can afford training and comparing a small selection of models.
We select the top three most similar languages according to each measure and take the highest transfer score produced by any of them 
(Table~\ref{tab:effect_of_language_choice_metric}, bottom).
The same trends still hold as when picking only one 
candidate.
However, the overall results are much better, and the gaps between the performance losses between the different measures become smaller.
\textbf{Comparing the results of multiple top source language candidates often yields better results than only considering the most similar one.}

\begin{table}
\setlength{\tabcolsep}{3pt} %
\centering
\adjustbox{max width=\linewidth}{%
\begin{tabular}{lrrrrrrr}
\toprule
 & \llap{POS} deV & POS & LAS & UAS & \texttt{to.\hspace{-2pt}-b}\rlap{.} & \texttt{to.\hspace{-2pt}-t}\rlap{.} & \texttt{to.\hspace{-2pt}-m}\rlap{.} \\
\midrule
\multicolumn{8}{@{}l@{}}{\textit{Top-1 source candidate (= best src in another experiment)}} \\
\midrule
POS deV & ---~~ & \cellcolor[HTML]{fef4ef} 3\stdev{\phantom{0}4} & \cellcolor[HTML]{fffefe} \cellcolor[HTML]{fef0e8} 4\stdev{\phantom{0}6} & \cellcolor[HTML]{feede4} 5\stdev{\phantom{0}6} & \cellcolor[HTML]{fee9dd} 6\stdev{10} & \cellcolor[HTML]{fee6d9} 7\stdev{12} & \cellcolor[HTML]{fee9dd} 6\stdev{\phantom{0}6} \\
POS & \cellcolor[HTML]{fee9dd} 6\stdev{\phantom{0}9} & \cellcolor[HTML]{fffefe} ---~~ & \cellcolor[HTML]{fff7f3} 2\stdev{\phantom{0}5} & \cellcolor[HTML]{fef4ef} 3\stdev{\phantom{0}5} & \cellcolor[HTML]{fddecb} 9\stdev{11} & \cellcolor[HTML]{fddbc7} 10\stdev{12} & \cellcolor[HTML]{f8bb9e} 16\stdev{14} \\
LAS & \cellcolor[HTML]{fee6d9} 7\stdev{\phantom{0}9} & \cellcolor[HTML]{fffbfa} 1\stdev{\phantom{0}2} &\cellcolor[HTML]{fffefe} ---~~ & \cellcolor[HTML]{fffbfa} 1\stdev{\phantom{0}2} & \cellcolor[HTML]{fcd5bf} 11\stdev{14} & \cellcolor[HTML]{fddbc7} 10\stdev{13} & \cellcolor[HTML]{f6af8e} 18\stdev{14} \\
UAS & \cellcolor[HTML]{fde2d2} 8\stdev{10} & \cellcolor[HTML]{fef4ef} 3\stdev{\phantom{0}5} & \cellcolor[HTML]{fffbfa} 1\stdev{\phantom{0}2} & \cellcolor[HTML]{fffefe} ---~~ & \cellcolor[HTML]{fbd0b9} 12\stdev{15} & \cellcolor[HTML]{fddbc7} 10\stdev{12} & \cellcolor[HTML]{f7b596} 17\stdev{14} \\
\texttt{top.-}\rlap{\texttt{b.}} & \cellcolor[HTML]{fbd0b9} 12\stdev{11} & \cellcolor[HTML]{fddbc7} 10\stdev{11} & \cellcolor[HTML]{fcd5bf} 11\stdev{12} & \cellcolor[HTML]{facab1} 13\stdev{13} & \cellcolor[HTML]{fffefe} ---~~ & \cellcolor[HTML]{fef4ef} 3\stdev{\phantom{0}7} & \cellcolor[HTML]{f5aa89} 19\stdev{15} \\
\texttt{top.-}\rlap{\texttt{tr.}} & \cellcolor[HTML]{fddbc7} 10\stdev{\phantom{0}9} & \cellcolor[HTML]{fde2d2} 8\stdev{\phantom{0}8} & \cellcolor[HTML]{fde2d2} 8\stdev{\phantom{0}8} & \cellcolor[HTML]{fddecb} 9\stdev{\phantom{0}9} & \cellcolor[HTML]{fef4ef} 3\stdev{\phantom{0}6} & \cellcolor[HTML]{fffefe} ---~~ & \cellcolor[HTML]{f6af8e} 18\stdev{13} \\
\texttt{top.-}\rlap{\texttt{m.}} & \cellcolor[HTML]{feede4} 5\stdev{\phantom{0}6} & \cellcolor[HTML]{fef0e8} 4\stdev{\phantom{0}3} & \cellcolor[HTML]{feede4} 5\stdev{\phantom{0}5} & \cellcolor[HTML]{feede4} 5\stdev{\phantom{0}5} & \cellcolor[HTML]{fee6d9} 7\stdev{\phantom{0}7} & \cellcolor[HTML]{fee9dd} 6\stdev{\phantom{0}7} & \cellcolor[HTML]{fffefe} ---~~ \\
\midrule
\multicolumn{8}{@{}l@{}}{\textit{Top-3 source candidates}} \\
\midrule
POS deV & \cellcolor[HTML]{fffefe} ---~~ &  \cellcolor[HTML]{fffbfa} 1\stdev{\phantom{0}3} & \cellcolor[HTML]{fff7f3} 2\stdev{\phantom{0}4} & \cellcolor[HTML]{fef4ef} 3\stdev{\phantom{0}4} & \cellcolor[HTML]{fef4ef} 3\stdev{\phantom{0}5} & \cellcolor[HTML]{fff7f3} 2\stdev{\phantom{0}3} & \cellcolor[HTML]{fef0e8} 4\stdev{\phantom{0}4} \\
POS & \cellcolor[HTML]{fef4ef} 3\stdev{\phantom{0}6} & \cellcolor[HTML]{fffefe} ---~~ & \cellcolor[HTML]{fffbfa} 1\stdev{\phantom{0}2} & \cellcolor[HTML]{fffbfa} 1\stdev{\phantom{0}3} & \cellcolor[HTML]{fef0e8} 4\stdev{\phantom{0}6} & \cellcolor[HTML]{fef0e8} 4\stdev{\phantom{0}6} & \cellcolor[HTML]{fee6d9} 7\stdev{\phantom{0}7} \\
LAS & \cellcolor[HTML]{fef4ef} 3\stdev{\phantom{0}5} & \cellcolor[HTML]{fffefe} 0\stdev{\phantom{0}1} & \cellcolor[HTML]{fffefe} ---~~ & \cellcolor[HTML]{fffefe} ---~~ & \cellcolor[HTML]{feede4} 5\stdev{\phantom{0}8} & \cellcolor[HTML]{feede4} 5\stdev{\phantom{0}8} & \cellcolor[HTML]{fddecb} 9\stdev{10} \\
UAS & \cellcolor[HTML]{fef0e8} 4\stdev{\phantom{0}6} & \cellcolor[HTML]{fffbfa} 1\stdev{\phantom{0}3} & \cellcolor[HTML]{fffefe} 0\stdev{\phantom{0}1} & \cellcolor[HTML]{fffefe} ---~~ & \cellcolor[HTML]{feede4} 5\stdev{\phantom{0}8} & \cellcolor[HTML]{feede4} 5\stdev{\phantom{0}7} & \cellcolor[HTML]{fde2d2} 8\stdev{\phantom{0}9} \\
\texttt{top.-}\rlap{\texttt{b.}} & \cellcolor[HTML]{fee9dd} 6\stdev{\phantom{0}8} & \cellcolor[HTML]{feede4} 5\stdev{\phantom{0}7} & \cellcolor[HTML]{feede4} 5\stdev{\phantom{0}8} & \cellcolor[HTML]{feede4} 5\stdev{\phantom{0}8} & \cellcolor[HTML]{fffefe} ---~~ & \cellcolor[HTML]{fffbfa} 1\stdev{\phantom{0}3} & \cellcolor[HTML]{fbd0b9} 12\stdev{11} \\
\texttt{top.-}\rlap{\texttt{tr.}} & \cellcolor[HTML]{feede4} 5\stdev{\phantom{0}6} & \cellcolor[HTML]{feede4} 5\stdev{\phantom{0}6} & \cellcolor[HTML]{fef0e8} 4\stdev{\phantom{0}5} & \cellcolor[HTML]{feede4} 5\stdev{\phantom{0}6} & \cellcolor[HTML]{fff7f3} 2\stdev{\phantom{0}5} & \cellcolor[HTML]{fffefe} ---~~ & \cellcolor[HTML]{fbd0b9} 12\stdev{10} \\
\texttt{top.-}\rlap{\texttt{m.}} & \cellcolor[HTML]{fff7f3} 2\stdev{\phantom{0}2} & \cellcolor[HTML]{fff7f3} 2\stdev{\phantom{0}2} & \cellcolor[HTML]{fff7f3} 2\stdev{\phantom{0}2} & \cellcolor[HTML]{fff7f3} 2\stdev{\phantom{0}2} & \cellcolor[HTML]{fef4ef} 3\stdev{\phantom{0}3} & \cellcolor[HTML]{fef4ef} 3\stdev{\phantom{0}4} & \cellcolor[HTML]{fffefe} ---~~ \\
\bottomrule
\end{tabular}
}
\caption{%
\textbf{Mean performance loss if picking source languages based solely on performance on the task in the column.}
Performance loss in percentage points.
Subscripts = standard deviations;
``word*'' = \overlapword{}, \overlaptri{}, \overlapsubword{}; deV = \citet{de-vries-etal-2022-make}.
}
\label{tab:effect_of_language_choice_task}
\end{table}

\subsubsection{Should I choose a source language based on another experiment?}
\label{sec:results-choosing-based-on-experiment}

We repeat these analyses but select source languages based on how well they served as source languages for the target language in other experiments 
Table~\ref{tab:effect_of_language_choice_task}).

\textbf{For the tasks included in our experiments, choosing a source language based on the results for a similar task with similar input representations leads to only small losses:} 
POS tagging results serve as a good predictor for parsing and vice versa, but not for topic classification. Similarly, the results for some topic classification settings are good predictors for each other, but not for POS tagging and parsing.
These losses are often even smaller than when selecting source languages based on similarity measures 
(cf.\ Tables \ref{tab:effect_of_language_choice_metric} and~\ref{tab:effect_of_language_choice_task}).

However, using other input representations weakens the predictive effect somewhat: e.g., for picking source languages for POS tagging with UDPipe, it is worse to choose based on \citeauthor{de-vries-etal-2022-make}'s XLM-R results than to choose based on UDPipe's parsing results. 
\Topicsmbert{} is an especially poor predictor -- this method likely suggests a fairly arbitrary mBERT language, which might or might not be similar to the target.

\section{Conclusion}
\label{sec:conclusion}

We investigated how linguistic and dataset level similarities impact cross-lingual transfer, and  
find that the most predictive similarity measures differ across tasks and input representations.
For practitioners working on the tasks included in our study, we recommend choosing a source language based on a pertinent similarity measure, or, if results for an extensive transfer experiment involving a similar task and similar input representations are available, based on the transfer results in that experiment.
Future work should investigate to what extent these patterns hold across even more languages, tasks, and NLP models.

\section*{Acknowledgements}

We would like to thank Katherine Metcalf, Maureen de Seyssel, Sinead Williamson, Skyler Seto, Stéphane Aroca-Ouellette, and the anonymous reviewers for their helpful comments, discussions, and feedback.

\section*{Limitations}
\label{sec:limitations}

Throughout our study we have made a number of pragmatically motivated decisions. Although well motivated, these decisions come with some limitations that we address below.

\paragraph{Data} Despite the fact that our study includes more languages than any prior work, we were able to include only 263 of the world's $\sim$7000 languages. Moreover, the usual high-resource languages are also over-represented in our work. We encourage data collection in more languages, especially those that are currently extremely low-resource or not available at all.

Despite our best efforts to mitigate any confounding effects, it is impossible to entirely avoid these.
For instance, while the parallel nature of the SIB-200 datasets avoids some of the confounders related to UD (datasets containing texts from many different sources, genres, and domains), parallel datasets can also show translation artifacts \cite{artetxe-etal-2020-translation}.

\paragraph{Models} 
We investigate one model type per task. 
Our model choices are motivated by practical considerations.
We use UDPipe in part because of its state-of-the-art status for many languages and its wide coverage of languages, making it a likely out-of-the-box tool to be used for (morpho)syntactic annotations. 
An important motivation for us to choose MLPs for our topic classification experiments is the lightweight architecture that allows us to quickly train and compare many models, and we show that the performance is not far off from larger, more resource-intensive models~(Table~\ref{tab:intra-vs-crosslingual-results}).
We include $n$-gram-based MLPs in order to be able to have full control over the language model training data.
Despite these models not having access to word order or syntactic information of the input data (and being more affected by exact word choices than a model with (sub)word embeddings would be),  we include them because of their computational cheapness, and because there are no (comparable) monolingual embeddings for all $\sim$200 languages in the topic classification data.

We cannot guarantee that our findings hold for other models that could be used for the task, which is an avenue for investigation in future work.
We also do not consider model- or tokenization-specific biases that may have different effects based on the 
word order \cite{white-cotterell-2021-examining},
morphology \cite{park-etal-2021-morphology},
or writing system and orthography \cite{sproat-gutkin-2021-taxonomy}.

\paragraph{Similarity measures} Within linguistics, there is a rich literature on defining similarity measures (cf.~\citealp{borin2013why}). We chose to include similarity measures that are commonly used in NLP studies, and we adapt them where necessary. However, we acknowledge that there are many other similarity measures that are interesting to investigate.

Recent work has focused on extending lang2vec \cite{khan-etal-2025-uriel, amirzadeh-etal-2025-data2lang2vec} -- we do not include these extensions in this paper as they were released concurrently to our work.

\paragraph{Tasks} We included POS tagging, dependency parsing, and topic classification in our study. An important motivation for this choice was the availability of data in a significant number of languages for these tasks. However, many more NLP tasks exist that are important to investigate. We echo~\citet{philippy-etal-2023-towards}: it is especially important to extend this work to more generative tasks, such as language modeling.

\paragraph{Analysis}
The transfer results are overall rather symmetric (i.e., the scores when training on language \textit{A} and evaluating on language \textit{B} tend to be similar to when training on \textit{B} and testing on \textit{A}; compare the upper right and lower left triangles of the result heatmaps in Figure~\ref{fig:heatmaps-intext}).
However, these trends are not perfectly symmetrical.
Transfer asymmetries have also been observed by other researchers \cite{malkin-etal-2022-balanced, protasov-etal-2024-super}.
However, we do not assume that these asymmetries mean that certain languages are intrinsically well-suited as source or target languages, but rather that they reflect peculiarities of the data sets (cf.\ \citealp{bjerva-2024-role}). 
We encourage further research on the (a)symmetries of cross-lingual transfer patterns.

\bibliography{anthology,custom}

\begin{thebibliography}{60}
\providecommand{\natexlab}[1]{#1}

\bibitem[{Adelani et~al.(2024)Adelani, Liu, Shen, Vassilyev, Alabi, Mao, Gao, and Lee}]{adelani-etal-2024-sib}
David~Ifeoluwa Adelani, Hannah Liu, Xiaoyu Shen, Nikita Vassilyev, Jesujoba~O. Alabi, Yanke Mao, Haonan Gao, and En-Shiun~Annie Lee. 2024.
\newblock \href {https://aclanthology.org/2024.eacl-long.14/} {{SIB}-200: A simple, inclusive, and big evaluation dataset for topic classification in 200+ languages and dialects}.
\newblock In \emph{Proceedings of the 18th Conference of the European Chapter of the Association for Computational Linguistics (Volume 1: Long Papers)}, pages 226--245, St. Julian{'}s, Malta. Association for Computational Linguistics.

\bibitem[{Adelani et~al.(2022)Adelani, Neubig, Ruder, Rijhwani, Beukman, Palen-Michel, Lignos, Alabi, Muhammad, Nabende, Dione, Bukula, Mabuya, Dossou, Sibanda, Buzaaba, Mukiibi, Kalipe, Mbaye, Taylor, Kabore, Emezue, Aremu, Ogayo, Gitau, Munkoh-Buabeng, Memdjokam~Koagne, Tapo, Macucwa, Marivate, Mboning, Gwadabe, Adewumi, Ahia, Nakatumba-Nabende, Mokono, Ezeani, Chukwuneke, Adeyemi, Hacheme, Abdulmumim, Ogundepo, Yousuf, Moteu~Ngoli, and Klakow}]{adelani-etal-2022-masakhaner}
David~Ifeoluwa Adelani, Graham Neubig, Sebastian Ruder, Shruti Rijhwani, Michael Beukman, Chester Palen-Michel, Constantine Lignos, Jesujoba~O. Alabi, Shamsuddeen~H. Muhammad, Peter Nabende, Cheikh M.~Bamba Dione, Andiswa Bukula, Rooweither Mabuya, Bonaventure F.~P. Dossou, Blessing Sibanda, Happy Buzaaba, Jonathan Mukiibi, Godson Kalipe, Derguene Mbaye, and 26 others. 2022.
\newblock \href {https://doi.org/10.18653/v1/2022.emnlp-main.298} {{M}asakha{NER} 2.0: {A}frica-centric transfer learning for named entity recognition}.
\newblock In \emph{Proceedings of the 2022 Conference on Empirical Methods in Natural Language Processing}, pages 4488--4508, Abu Dhabi, United Arab Emirates. Association for Computational Linguistics.

\bibitem[{Ahuja et~al.(2022)Ahuja, Kumar, Dandapat, and Choudhury}]{ahuja-etal-2022-multi}
Kabir Ahuja, Shanu Kumar, Sandipan Dandapat, and Monojit Choudhury. 2022.
\newblock \href {https://doi.org/10.18653/v1/2022.acl-long.374} {Multi task learning for zero shot performance prediction of multilingual models}.
\newblock In \emph{Proceedings of the 60th Annual Meeting of the Association for Computational Linguistics (Volume 1: Long Papers)}, pages 5454--5467, Dublin, Ireland. Association for Computational Linguistics.

\bibitem[{AI4Bharat et~al.(2023)AI4Bharat, Gala, Chitale, AK, Doddapaneni, Gumma, Kumar, Nawale, Sujatha, Puduppully, Raghavan, Kumar, Khapra, Dabre, and Kunchukuttan}]{indictrans2-23}
AI4Bharat, Jay Gala, Pranjal~A. Chitale, Raghavan AK, Sumanth Doddapaneni, Varun Gumma, Aswanth Kumar, Janki Nawale, Anupama Sujatha, Ratish Puduppully, Vivek Raghavan, Pratyush Kumar, Mitesh~M. Khapra, Raj Dabre, and Anoop Kunchukuttan. 2023.
\newblock \href {https://arxiv.org/abs/arXiv:2305.16307} {{IndicTrans2}: Towards high-quality and accessible machine translation models for all 22 scheduled {Indian} languages}.

\bibitem[{Amirzadeh et~al.(2025)Amirzadeh, Jafari, Harju, and van~der Goot}]{amirzadeh-etal-2025-data2lang2vec}
Hamidreza Amirzadeh, Sadegh Jafari, Anika Harju, and Rob van~der Goot. 2025.
\newblock \href {https://aclanthology.org/2025.coling-main.435/} {data2lang2vec: Data driven typological features completion}.
\newblock In \emph{Proceedings of the 31st International Conference on Computational Linguistics}, pages 6520--6529, Abu Dhabi, UAE. Association for Computational Linguistics.

\bibitem[{Artetxe et~al.(2020)Artetxe, Labaka, and Agirre}]{artetxe-etal-2020-translation}
Mikel Artetxe, Gorka Labaka, and Eneko Agirre. 2020.
\newblock \href {https://doi.org/10.18653/v1/2020.emnlp-main.618} {Translation artifacts in cross-lingual transfer learning}.
\newblock In \emph{Proceedings of the 2020 Conference on Empirical Methods in Natural Language Processing (EMNLP)}, pages 7674--7684, Online. Association for Computational Linguistics.

\bibitem[{Bagheri~Nezhad and Agrawal(2024)}]{bagheri-nezhad-agrawal-2024-drives}
Sina Bagheri~Nezhad and Ameeta Agrawal. 2024.
\newblock \href {https://doi.org/10.18653/v1/2024.vardial-1.2} {What drives performance in multilingual language models?}
\newblock In \emph{Proceedings of the Eleventh Workshop on NLP for Similar Languages, Varieties, and Dialects (VarDial 2024)}, pages 16--27, Mexico City, Mexico. Association for Computational Linguistics.

\bibitem[{Bates et~al.(2015)Bates, M{\"a}chler, Bolker, and Walker}]{bates2015fitting}
Douglas Bates, Martin M{\"a}chler, Ben Bolker, and Steve Walker. 2015.
\newblock \href {https://doi.org/10.18637/jss.v067.i01} {Fitting linear mixed-effects models using {lme4}}.
\newblock \emph{Journal of Statistical Software}, 67(1):1--48.

\bibitem[{Baylor et~al.(2023)Baylor, Ploeger, and Bjerva}]{baylor-etal-2023-past}
Emi Baylor, Esther Ploeger, and Johannes Bjerva. 2023.
\newblock \href {https://doi.org/10.18653/v1/2023.findings-emnlp.82} {The past, present, and future of typological databases in {NLP}}.
\newblock In \emph{Findings of the Association for Computational Linguistics: EMNLP 2023}, pages 1163--1169, Singapore. Association for Computational Linguistics.

\bibitem[{Bjerva(2024)}]{bjerva-2024-role}
Johannes Bjerva. 2024.
\newblock \href {https://doi.org/10.1162/coli_a_00498} {The role of typological feature prediction in {NLP} and linguistics}.
\newblock \emph{Computational Linguistics}, 50(2):781--794.

\bibitem[{Borin(2013)}]{borin2013why}
Lars Borin. 2013.
\newblock \href {https://doi.org/doi:10.1515/9783110305258.3} {The why and how of measuring linguistic differences}.
\newblock In Lars Borin and Anju Saxena, editors, \emph{Approaches to Measuring Linguistic Differences}, pages 3--26. De Gruyter Mouton, Berlin, Boston.

\bibitem[{Choi et~al.(2015)Choi, Tetreault, and Stent}]{choi-etal-2015-depends}
Jinho~D. Choi, Joel Tetreault, and Amanda Stent. 2015.
\newblock \href {https://doi.org/10.3115/v1/P15-1038} {It depends: Dependency parser comparison using a web-based evaluation tool}.
\newblock In \emph{Proceedings of the 53rd Annual Meeting of the Association for Computational Linguistics and the 7th International Joint Conference on Natural Language Processing (Volume 1: Long Papers)}, pages 387--396, Beijing, China. Association for Computational Linguistics.

\bibitem[{Collins and Kayne(2011)}]{sswl}
Chris Collins and Richard Kayne. 2011.
\newblock Syntactic structures of the world's languages.
\newblock New York University, New York.

\bibitem[{Conneau et~al.(2020)Conneau, Khandelwal, Goyal, Chaudhary, Wenzek, Guzm{\'a}n, Grave, Ott, Zettlemoyer, and Stoyanov}]{conneau-etal-2020-unsupervised}
Alexis Conneau, Kartikay Khandelwal, Naman Goyal, Vishrav Chaudhary, Guillaume Wenzek, Francisco Guzm{\'a}n, Edouard Grave, Myle Ott, Luke Zettlemoyer, and Veselin Stoyanov. 2020.
\newblock \href {https://doi.org/10.18653/v1/2020.acl-main.747} {Unsupervised cross-lingual representation learning at scale}.
\newblock In \emph{Proceedings of the 58th Annual Meeting of the Association for Computational Linguistics}, pages 8440--8451, Online. Association for Computational Linguistics.

\bibitem[{de~Marneffe et~al.(2021)de~Marneffe, Manning, Nivre, and Zeman}]{de-marneffe-etal-2021-universal}
Marie-Catherine de~Marneffe, Christopher~D. Manning, Joakim Nivre, and Daniel Zeman. 2021.
\newblock \href {https://doi.org/10.1162/coli_a_00402} {{U}niversal {D}ependencies}.
\newblock \emph{Computational Linguistics}, 47(2):255--308.

\bibitem[{de~Vries et~al.(2022)de~Vries, Wieling, and Nissim}]{de-vries-etal-2022-make}
Wietse de~Vries, Martijn Wieling, and Malvina Nissim. 2022.
\newblock \href {https://doi.org/10.18653/v1/2022.acl-long.529} {Make the best of cross-lingual transfer: Evidence from {POS} tagging with over 100 languages}.
\newblock In \emph{Proceedings of the 60th Annual Meeting of the Association for Computational Linguistics (Volume 1: Long Papers)}, pages 7676--7685, Dublin, Ireland. Association for Computational Linguistics.

\bibitem[{Devlin et~al.(2019)Devlin, Chang, Lee, and Toutanova}]{devlin-etal-2019-bert}
Jacob Devlin, Ming-Wei Chang, Kenton Lee, and Kristina Toutanova. 2019.
\newblock \href {https://doi.org/10.18653/v1/N19-1423} {{BERT}: Pre-training of deep bidirectional transformers for language understanding}.
\newblock In \emph{Proceedings of the 2019 Conference of the North {A}merican Chapter of the Association for Computational Linguistics: Human Language Technologies, Volume 1 (Long and Short Papers)}, pages 4171--4186, Minneapolis, Minnesota. Association for Computational Linguistics.

\bibitem[{Doumbouya et~al.(2023)Doumbouya, Diané, Cissé, Diané, Sow, Doumbouya, Bangoura, Bayo, Condé, Diané, Piech, and Manning}]{mt4nko-23}
Moussa Doumbouya, Baba~Mamadi Diané, Solo~Farabado Cissé, Djibrila Diané, Abdoulaye Sow, Séré~Moussa Doumbouya, Daouda Bangoura, Fodé~Moriba Bayo, Ibrahima Sory~2. Condé, Kalo~Mory Diané, Chris Piech, and Christopher Manning. 2023.
\newblock \href {https://aclanthology.org/2023.wmt-1.34} {Machine translation for {Nko}: Tools, corpora, and baseline results}.
\newblock In \emph{Proceedings of the Eighth Conference on Machine Translation}, pages 312--343, Singapore. Association for Computational Linguistics.

\bibitem[{Dryer and Haspelmath(2013)}]{wals}
Matthew~S. Dryer and Martin Haspelmath, editors. 2013.
\newblock \emph{The World Atlas of Language Structures Online}.
\newblock Max Planck Institute for Evolutionary Anthropology.

\bibitem[{Gower(1971)}]{gower1971coefficient}
J.~C. Gower. 1971.
\newblock \href {http://www.jstor.org/stable/2528823} {A general coefficient of similarity and some of its properties}.
\newblock \emph{Biometrics}, 27(4):857--871.

\bibitem[{Goyal et~al.(2022)Goyal, Gao, Chaudhary, Chen, Wenzek, Ju, Krishnan, Ranzato, Guzm{\'a}n, and Fan}]{goyal-etal-2022-flores}
Naman Goyal, Cynthia Gao, Vishrav Chaudhary, Peng-Jen Chen, Guillaume Wenzek, Da~Ju, Sanjana Krishnan, Marc{'}Aurelio Ranzato, Francisco Guzm{\'a}n, and Angela Fan. 2022.
\newblock \href {https://doi.org/10.1162/tacl_a_00474} {The {F}lores-101 evaluation benchmark for low-resource and multilingual machine translation}.
\newblock \emph{Transactions of the Association for Computational Linguistics}, 10:522--538.

\bibitem[{Guzmán et~al.(2019)Guzmán, Chen, Ott, Pino, Lample, Koehn, Chaudhary, and Ranzato}]{flores1-19}
Francisco Guzmán, Peng-Jen Chen, Myle Ott, Juan Pino, Guillaume Lample, Philipp Koehn, Vishrav Chaudhary, and Marc’Aurelio Ranzato. 2019.
\newblock \href {https://aclanthology.org/D19-1632} {The {FLORES} evaluation datasets for low-resource machine translation: {N}epali{--}{E}nglish and {S}inhala{--}{E}nglish}.
\newblock In \emph{Proceedings of the 2019 Conference on Empirical Methods in Natural Language Processing and the 9th International Joint Conference on Natural Language Processing (EMNLP-IJCNLP)}, pages 6098--6111, Hong Kong, China. Association for Computational Linguistics.

\bibitem[{Hammarstr\"{o}m et~al.(2024)Hammarstr\"{o}m, Forkel, Haspelmath, and Bank}]{glottolog2024}
Harald Hammarstr\"{o}m, Robert Forkel, Martin Haspelmath, and Sebastian Bank. 2024.
\newblock Glottolog 5.0.
\newblock Max Planck Institute for Evolutionary Anthropology, Leipzig. Available online at \url{https://glottolog.org}.

\bibitem[{Hermjakob et~al.(2018)Hermjakob, May, and Knight}]{hermjakob-etal-2018-box}
Ulf Hermjakob, Jonathan May, and Kevin Knight. 2018.
\newblock \href {https://doi.org/10.18653/v1/P18-4003} {Out-of-the-box universal {R}omanization tool uroman}.
\newblock In \emph{Proceedings of {ACL} 2018, System Demonstrations}, pages 13--18, Melbourne, Australia. Association for Computational Linguistics.

\bibitem[{J\"{a}ger(2018{\natexlab{a}})}]{jaeger2018extracting}
Gerhard J\"{a}ger. 2018{\natexlab{a}}.
\newblock \href {https://doi.org/10.17605/OSF.IO/CUFV7} {Extracting language distances and character matrices from {ASJP} data}.
\newblock OSF.

\bibitem[{J\"{a}ger(2018{\natexlab{b}})}]{jaeger2018global}
Gerhard J\"{a}ger. 2018{\natexlab{b}}.
\newblock \href {https://doi.org/10.1038/sdata.2018.189} {Global-scale phylogenetic linguistic inference from lexical resources}.
\newblock \emph{Scientific data}, 5(180189).

\bibitem[{Kargaran et~al.(2024)Kargaran, Yvon, and Sch{\"u}tze}]{kargaran-etal-2024-glotscript}
Amir~Hossein Kargaran, Fran{\c{c}}ois Yvon, and Hinrich Sch{\"u}tze. 2024.
\newblock \href {https://aclanthology.org/2024.lrec-main.687/} {{G}lot{S}cript: A resource and tool for low resource writing system identification}.
\newblock In \emph{Proceedings of the 2024 Joint International Conference on Computational Linguistics, Language Resources and Evaluation (LREC-COLING 2024)}, pages 7774--7784, Torino, Italia. ELRA and ICCL.

\bibitem[{Khan et~al.(2025)Khan, Shipton, Anugraha, Duan, Hoang, Khiu, Do{\u{g}}ru{\"o}z, and Lee}]{khan-etal-2025-uriel}
Aditya Khan, Mason Shipton, David Anugraha, Kaiyao Duan, Phuong~H. Hoang, Eric Khiu, A.~Seza Do{\u{g}}ru{\"o}z, and En-Shiun~Annie Lee. 2025.
\newblock \href {https://aclanthology.org/2025.coling-main.463/} {{URIEL}+: Enhancing linguistic inclusion and usability in a typological and multilingual knowledge base}.
\newblock In \emph{Proceedings of the 31st International Conference on Computational Linguistics}, pages 6937--6952, Abu Dhabi, UAE. Association for Computational Linguistics.

\bibitem[{Lauscher et~al.(2020)Lauscher, Ravishankar, Vuli{\'c}, and Glava{\v{s}}}]{lauscher-etal-2020-zero}
Anne Lauscher, Vinit Ravishankar, Ivan Vuli{\'c}, and Goran Glava{\v{s}}. 2020.
\newblock \href {https://doi.org/10.18653/v1/2020.emnlp-main.363} {From zero to hero: {O}n the limitations of zero-shot language transfer with multilingual {T}ransformers}.
\newblock In \emph{Proceedings of the 2020 Conference on Empirical Methods in Natural Language Processing (EMNLP)}, pages 4483--4499, Online. Association for Computational Linguistics.

\bibitem[{Lewis et~al.(2015)Lewis, Simons, and Fennig}]{ethnologue2015}
M.~Paul Lewis, Gary~F. Simons, and Charles~D. Fennig. 2015.
\newblock Ethnologue: Languages of the world, eighteenth edition.
\newblock SIL International, Dallas, Texas.

\bibitem[{Lin et~al.(2024)Lin, Hu, Zhang, Martins, and Schuetze}]{lin-etal-2024-mplm}
Peiqin Lin, Chengzhi Hu, Zheyu Zhang, Andre Martins, and Hinrich Schuetze. 2024.
\newblock \href {https://aclanthology.org/2024.findings-eacl.20/} {m{PLM}-sim: Better cross-lingual similarity and transfer in multilingual pretrained language models}.
\newblock In \emph{Findings of the Association for Computational Linguistics: EACL 2024}, pages 276--310, St. Julian{'}s, Malta. Association for Computational Linguistics.

\bibitem[{Lin et~al.(2019)Lin, Chen, Lee, Li, Zhang, Xia, Rijhwani, He, Zhang, Ma, Anastasopoulos, Littell, and Neubig}]{lin-etal-2019-choosing}
Yu-Hsiang Lin, Chian-Yu Chen, Jean Lee, Zirui Li, Yuyan Zhang, Mengzhou Xia, Shruti Rijhwani, Junxian He, Zhisong Zhang, Xuezhe Ma, Antonios Anastasopoulos, Patrick Littell, and Graham Neubig. 2019.
\newblock \href {https://doi.org/10.18653/v1/P19-1301} {Choosing transfer languages for cross-lingual learning}.
\newblock In \emph{Proceedings of the 57th Annual Meeting of the Association for Computational Linguistics}, pages 3125--3135, Florence, Italy. Association for Computational Linguistics.

\bibitem[{Littell et~al.(2019)Littell, Mortensen, and Anastasopoulos}]{lang2vec_package}
Patrick Littell, David Mortensen, and Antonis Anastasopoulos. 2019.
\newblock \href {https://github.com/antonisa/lang2vec} {lang2vec 1.1.6}.

\bibitem[{Littell et~al.(2017)Littell, Mortensen, Lin, Kairis, Turner, and Levin}]{littell-etal-2017-uriel}
Patrick Littell, David~R. Mortensen, Ke~Lin, Katherine Kairis, Carlisle Turner, and Lori Levin. 2017.
\newblock \href {https://aclanthology.org/E17-2002/} {{URIEL} and lang2vec: Representing languages as typological, geographical, and phylogenetic vectors}.
\newblock In \emph{Proceedings of the 15th Conference of the {E}uropean Chapter of the Association for Computational Linguistics: Volume 2, Short Papers}, pages 8--14, Valencia, Spain. Association for Computational Linguistics.

\bibitem[{Malkin et~al.(2022)Malkin, Limisiewicz, and Stanovsky}]{malkin-etal-2022-balanced}
Dan Malkin, Tomasz Limisiewicz, and Gabriel Stanovsky. 2022.
\newblock \href {https://doi.org/10.18653/v1/2022.naacl-main.361} {A balanced data approach for evaluating cross-lingual transfer: Mapping the linguistic blood bank}.
\newblock In \emph{Proceedings of the 2022 Conference of the North American Chapter of the Association for Computational Linguistics: Human Language Technologies}, pages 4903--4915, Seattle, United States. Association for Computational Linguistics.

\bibitem[{McDonald and Nivre(2011)}]{mcdonald-nivre-2011-analyzing}
Ryan McDonald and Joakim Nivre. 2011.
\newblock \href {https://doi.org/10.1162/coli_a_00039} {Analyzing and integrating dependency parsers}.
\newblock \emph{Computational Linguistics}, 37(1):197--230.

\bibitem[{Moran et~al.(2014)Moran, McCloy, and Wright}]{phoible}
Steven Moran, Daniel McCloy, and Richard Wright. 2014.
\newblock {PHOIBLE} online.
\newblock Max Planck Institute for Evolutionary Anthropology, Leipzig.

\bibitem[{Muller et~al.(2023)Muller, Gupta, Fauconnier, Patwardhan, Vandyke, and Agarwal}]{muller2023languages}
Benjamin Muller, Deepanshu Gupta, Jean-Philippe Fauconnier, Siddharth Patwardhan, David Vandyke, and Sachin Agarwal. 2023.
\newblock \href {https://proceedings.mlr.press/v203/muller23a.html} {Languages you know influence those you learn: {Impact} of language characteristics on multi-lingual text-to-text transfer}.
\newblock In \emph{Proceedings of The 1st Transfer Learning for Natural Language Processing Workshop}, volume 203 of \emph{Proceedings of Machine Learning Research}, pages 88--102. PMLR.

\bibitem[{Nivre and Fang(2017)}]{nivre-fang-2017-universal}
Joakim Nivre and Chiao-Ting Fang. 2017.
\newblock \href {https://aclanthology.org/W17-0411/} {{U}niversal {D}ependency evaluation}.
\newblock In \emph{Proceedings of the {N}o{D}a{L}i{D}a 2017 Workshop on Universal Dependencies ({UDW} 2017)}, pages 86--95, Gothenburg, Sweden. Association for Computational Linguistics.

\bibitem[{{NLLB Team} et~al.(2022){NLLB Team}, Costa-jussà, Cross, Çelebi, Elbayad, Heafield, Heffernan, Kalbassi, Lam, Licht, Maillard, Sun, Wang, Wenzek, Youngblood, Akula, Barrault, Mejia-Gonzalez, Hansanti, Hoffman, Jarrett, Sadagopan, Rowe, Spruit, Tran, Andrews, Ayan, Bhosale, Edunov, Fan, Gao, Goswami, Guzmán, Koehn, Mourachko, Ropers, Saleem, Schwenk, and Wang}]{nllb-22}
{NLLB Team}, Marta~R. Costa-jussà, James Cross, Onur Çelebi, Maha Elbayad, Kenneth Heafield, Kevin Heffernan, Elahe Kalbassi, Janice Lam, Daniel Licht, Jean Maillard, Anna Sun, Skyler Wang, Guillaume Wenzek, Al~Youngblood, Bapi Akula, Loic Barrault, Gabriel Mejia-Gonzalez, Prangthip Hansanti, and 20 others. 2022.
\newblock \href {https://arxiv.org/abs/arXiv:1902.01382} {No language left behind: Scaling human-centered machine translation}.

\bibitem[{Park et~al.(2021)Park, Zhang, Haley, Steimel, Liu, and Schwartz}]{park-etal-2021-morphology}
Hyunji~Hayley Park, Katherine~J. Zhang, Coleman Haley, Kenneth Steimel, Han Liu, and Lane Schwartz. 2021.
\newblock \href {https://doi.org/10.1162/tacl_a_00365} {Morphology matters: A multilingual language modeling analysis}.
\newblock \emph{Transactions of the Association for Computational Linguistics}, 9:261--276.

\bibitem[{Pedregosa et~al.(2011)Pedregosa, Varoquaux, Gramfort, Michel, Thirion, Grisel, Blondel, Prettenhofer, Weiss, Dubourg, Vanderplas, Passos, Cournapeau, Brucher, Perrot, and Duchesnay}]{scikit-learn}
F.~Pedregosa, G.~Varoquaux, A.~Gramfort, V.~Michel, B.~Thirion, O.~Grisel, M.~Blondel, P.~Prettenhofer, R.~Weiss, V.~Dubourg, J.~Vanderplas, A.~Passos, D.~Cournapeau, M.~Brucher, M.~Perrot, and E.~Duchesnay. 2011.
\newblock Scikit-learn: Machine learning in {P}ython.
\newblock \emph{Journal of Machine Learning Research}, 12:2825--2830.

\bibitem[{Philippy et~al.(2023)Philippy, Guo, and Haddadan}]{philippy-etal-2023-towards}
Fred Philippy, Siwen Guo, and Shohreh Haddadan. 2023.
\newblock \href {https://doi.org/10.18653/v1/2023.acl-long.323} {Towards a common understanding of contributing factors for cross-lingual transfer in multilingual language models: A review}.
\newblock In \emph{Proceedings of the 61st Annual Meeting of the Association for Computational Linguistics (Volume 1: Long Papers)}, pages 5877--5891, Toronto, Canada. Association for Computational Linguistics.

\bibitem[{Pires et~al.(2019)Pires, Schlinger, and Garrette}]{pires-etal-2019-multilingual}
Telmo Pires, Eva Schlinger, and Dan Garrette. 2019.
\newblock \href {https://doi.org/10.18653/v1/P19-1493} {How multilingual is multilingual {BERT}?}
\newblock In \emph{Proceedings of the 57th Annual Meeting of the Association for Computational Linguistics}, pages 4996--5001, Florence, Italy. Association for Computational Linguistics.

\bibitem[{Protasov et~al.(2024)Protasov, Stakovskii, Voloshina, Shavrina, and Panchenko}]{protasov-etal-2024-super}
Vitaly Protasov, Elisei Stakovskii, Ekaterina Voloshina, Tatiana Shavrina, and Alexander Panchenko. 2024.
\newblock \href {https://doi.org/10.18653/v1/2024.loresmt-1.10} {Super donors and super recipients: Studying cross-lingual transfer between high-resource and low-resource languages}.
\newblock In \emph{Proceedings of the Seventh Workshop on Technologies for Machine Translation of Low-Resource Languages (LoResMT 2024)}, pages 94--108, Bangkok, Thailand. Association for Computational Linguistics.

\bibitem[{{R Core Team}(2024)}]{r2024}
{R Core Team}. 2024.
\newblock \href {https://www.R-project.org/} {{R:} {A} language and environment for statistical computing}.

\bibitem[{Rice et~al.(2025)Rice, Marashian, Haynie, von~der Wense, and Palmer}]{rice2025untangling}
Enora Rice, Ali Marashian, Hannah Haynie, Katharina von~der Wense, and Alexis Palmer. 2025.
\newblock \href {https://arxiv.org/abs/2503.19979} {Untangling the influence of typology, data and model architecture on ranking transfer languages for cross-lingual {POS} tagging}.
\newblock \emph{Preprint}, arXiv:2503.19979.

\bibitem[{Samard{\v{z}}i{\'c} et~al.(2022)Samard{\v{z}}i{\'c}, Gutierrez-Vasques, van~der Goot, M{\"u}ller-Eberstein, Pelloni, and Plank}]{samardzic-etal-2022-language}
Tanja Samard{\v{z}}i{\'c}, Ximena Gutierrez-Vasques, Rob van~der Goot, Max M{\"u}ller-Eberstein, Olga Pelloni, and Barbara Plank. 2022.
\newblock \href {https://doi.org/10.18653/v1/2022.conll-1.18} {On language spaces, scales and cross-lingual transfer of {UD} parsers}.
\newblock In \emph{Proceedings of the 26th Conference on Computational Natural Language Learning (CoNLL)}, pages 266--281, Abu Dhabi, United Arab Emirates (Hybrid). Association for Computational Linguistics.

\bibitem[{Skirg{\aa}rd et~al.(2023)Skirg{\aa}rd, Haynie, Blasi, Hammarstr{\"o}m, Collins, Latarche, Lesage, Weber, Witzlack-Makarevich, Passmore, Chira, Maurits, Dinnage, Dunn, Reesink, Singer, Bowern, Epps, Hill, Vesakoski, Robbeets, Abbas, Auer, Bakker, Barbos, Borges, Danielsen, Dorenbusch, Dorn, Elliott, Falcone, Fischer, Ghanggo~Ate, Gibson, G{\"o}bel, Goodall, Gruner, Harvey, Hayes, Heer, Herrera~Miranda, H{\"u}bler, Huntington-Rainey, Ivani, Johns, Just, Kashima, Kipf, Klingenberg, K{\"o}nig, Koti, Kowalik, Krasnoukhova, Lindvall, Lorenzen, Lutzenberger, Martins, Mata~German, van~der Meer, Montoya~Samam{\'e}, M{\"u}ller, Murado$\breve{g}$lu, Neely, Nickel, Norvik, Oluoch, Peacock, Pearey, Peck, Petit, Pieper, Poblete, Prestipino, Raabe, Raja, Reimringer, Rey, Rizaew, Ruppert, Salmon, Sammet, Schembri, Schlabbach, Schmidt, Skilton, Smith, de~Sousa, Sverredal, Valle, Vera, Vo{\ss}, Witte, Wu, Yam, Ye, Yong, Yuditha, Zariquiey, Forkel, Evans, Levinson, Haspelmath, Greenhill, Atkinson, and
  Gray}]{grambank_release}
Hedvig Skirg{\aa}rd, Hannah~J. Haynie, Dami{\'a}n~E. Blasi, Harald Hammarstr{\"o}m, Jeremy Collins, Jay~J. Latarche, Jakob Lesage, Tobias Weber, Alena Witzlack-Makarevich, Sam Passmore, Angela Chira, Luke Maurits, Russell Dinnage, Michael Dunn, Ger Reesink, Ruth Singer, Claire Bowern, Patience Epps, Jane Hill, and 86 others. 2023.
\newblock \href {https://doi.org/10.1126/sciadv.adg6175} {Grambank reveals global patterns in the structural diversity of the world's languages}.
\newblock \emph{Science Advances}, 9.

\bibitem[{Sproat and Gutkin(2021)}]{sproat-gutkin-2021-taxonomy}
Richard Sproat and Alexander Gutkin. 2021.
\newblock \href {https://doi.org/10.1162/coli_a_00409} {The taxonomy of writing systems: How to measure how logographic a system is}.
\newblock \emph{Computational Linguistics}, 47(3):477--528.

\bibitem[{Srinivasan et~al.(2021)Srinivasan, Sitaram, Ganu, Dandapat, Bali, and Choudhury}]{srinivasan2021predicting}
Anirudh Srinivasan, Sunayana Sitaram, Tanuja Ganu, Sandipan Dandapat, Kalika Bali, and Monojit Choudhury. 2021.
\newblock \href {https://arxiv.org/abs/2110.08875} {Predicting the performance of multilingual {NLP} models}.
\newblock \emph{Computing Research Repository}, arXiv:2110.08875.

\bibitem[{Straka(2018)}]{straka-2018-udpipe}
Milan Straka. 2018.
\newblock \href {https://doi.org/10.18653/v1/K18-2020} {{UDP}ipe 2.0 prototype at {C}o{NLL} 2018 {UD} shared task}.
\newblock In \emph{Proceedings of the {C}o{NLL} 2018 Shared Task: Multilingual Parsing from Raw Text to Universal Dependencies}, pages 197--207, Brussels, Belgium. Association for Computational Linguistics.

\bibitem[{Straka(2023)}]{udpipe212}
Milan Straka. 2023.
\newblock \href {http://hdl.handle.net/11234/1-5200} {{Universal Dependencies} 2.12 models for {UDPipe} 2 (2023-07-17)}.
\newblock {LINDAT}/{CLARIAH}-{CZ} digital library at the Institute of Formal and Applied Linguistics ({{\'U}FAL}), Faculty of Mathematics and Physics, Charles University.

\bibitem[{Toossi et~al.(2024)Toossi, Huai, Liu, Khiu, Do{\u{g}}ru{\"o}z, and Lee}]{toossi-etal-2024-reproducibility}
Hasti Toossi, Guo Huai, Jinyu Liu, Eric Khiu, A.~Seza Do{\u{g}}ru{\"o}z, and En-Shiun Lee. 2024.
\newblock \href {https://doi.org/10.18653/v1/2024.naacl-srw.25} {A reproducibility study on quantifying language similarity: The impact of missing values in the {URIEL} knowledge base}.
\newblock In \emph{Proceedings of the 2024 Conference of the North American Chapter of the Association for Computational Linguistics: Human Language Technologies (Volume 4: Student Research Workshop)}, pages 233--241, Mexico City, Mexico. Association for Computational Linguistics.

\bibitem[{Virtanen et~al.(2020)Virtanen, Gommers, Oliphant, Haberland, Reddy, Cournapeau, Burovski, Peterson, Weckesser, Bright, {van der Walt}, Brett, Wilson, Millman, Mayorov, Nelson, Jones, Kern, Larson, Carey, Polat, Feng, Moore, {VanderPlas}, Laxalde, Perktold, Cimrman, Henriksen, Quintero, Harris, Archibald, Ribeiro, Pedregosa, {van Mulbregt}, and {SciPy 1.0 Contributors}}]{2020SciPy-NMeth}
Pauli Virtanen, Ralf Gommers, Travis~E. Oliphant, Matt Haberland, Tyler Reddy, David Cournapeau, Evgeni Burovski, Pearu Peterson, Warren Weckesser, Jonathan Bright, St{\'e}fan~J. {van der Walt}, Matthew Brett, Joshua Wilson, K.~Jarrod Millman, Nikolay Mayorov, Andrew R.~J. Nelson, Eric Jones, Robert Kern, Eric Larson, and 16 others. 2020.
\newblock \href {https://doi.org/10.1038/s41592-019-0686-2} {{SciPy} 1.0: {Fundamental} algorithms for scientific computing in {Python}}.
\newblock \emph{Nature Methods}, 17:261--272.

\bibitem[{Ward(1963)}]{ward1963hierarchical}
Joe~H. Ward, Jr. 1963.
\newblock \href {https://doi.org/10.1080/01621459.1963.10500845} {Hierarchical grouping to optimize an objective function}.
\newblock \emph{Journal of the American Statistical Association}, 58(301):236--244.

\bibitem[{White and Cotterell(2021)}]{white-cotterell-2021-examining}
Jennifer~C. White and Ryan Cotterell. 2021.
\newblock \href {https://doi.org/10.18653/v1/2021.acl-long.38} {Examining the inductive bias of neural language models with artificial languages}.
\newblock In \emph{Proceedings of the 59th Annual Meeting of the Association for Computational Linguistics and the 11th International Joint Conference on Natural Language Processing (Volume 1: Long Papers)}, pages 454--463, Online. Association for Computational Linguistics.

\bibitem[{Wichmann et~al.(2016)Wichmann, Holman, and Brown}]{asjp}
Søren Wichmann, Eric~W. Holman, and Cecil~H. Brown. 2016.
\newblock \href {https://asjp.clld.org/} {The {ASJP} database (version 17)}.
\newblock Zenodo.

\bibitem[{Xia et~al.(2020)Xia, Anastasopoulos, Xu, Yang, and Neubig}]{xia-etal-2020-predicting}
Mengzhou Xia, Antonios Anastasopoulos, Ruochen Xu, Yiming Yang, and Graham Neubig. 2020.
\newblock \href {https://doi.org/10.18653/v1/2020.acl-main.764} {Predicting performance for natural language processing tasks}.
\newblock In \emph{Proceedings of the 58th Annual Meeting of the Association for Computational Linguistics}, pages 8625--8646, Online. Association for Computational Linguistics.

\bibitem[{Zeman et~al.(2024)Zeman, Nivre, Abrams, Ackermann, Aepli, Aghaei, Agi{\'c}, Ahmadi, Ahrenberg, Ajede, Akkurt, Aleksandravi{\v c}i{\=u}t{\.e}, Alfina, Algom, Alnajjar, Alzetta, Andersen, Antonsen, Aoyama, Aplonova, Aquino, Aragon, Aranes, Aranzabe, Ar{\i}can, Arnard{\'o}ttir, Arutie, Arwidarasti, Asahara, {\'A}sgeirsd{\'o}ttir, Aslan, Asmazo{\u g}lu, Ateyah, Atmaca, Attia, Atutxa, Augustinus, Avel{\~a}s, Badmaeva, Balasubramani, Ballesteros, Banerjee, Bank, Barbu~Mititelu, Barkarson, Basile, Basmov, Batchelor, Bauer, Bedir, Behzad, Belieni, Bengoetxea, Benli, Ben~Moshe, Berg, Berk, Bhat, Biagetti, Bick, Bielinskien{\.e}, Bilgin~Ta{\c s}demir, Bjarnad{\'o}ttir, Blaschke, Blokland, Bobicev, Boizou, Bonilla, Borges~V{\"o}lker, B{\"o}rstell, Bosco, Bouma, Bowman, Boyd, Braggaar, Branco, Brokait{\.e}, Burchardt, Campos, Candito, Caron, Caron, Carvalheiro, Carvalho, Cassidy, Castro, Castro, Cavalcanti, Cebiro{\u g}lu~Eryi{\u g}it, Cecchini, Celano, {\v C}{\'e}pl{\"o}, Cesur, Cetin, {\c C}etino{\u g}lu,
  Chalub, Chamila, Chauhan, Chen, Chi, Chika, Cho, Choi, Chontaeva, Chun, Chung, Cignarella, Cinkov{\'a}, Collomb, {\c C}{\"o}ltekin, Connor, Corbetta, Corbetta, Costa, Courtin, Crabb{\'e}, Cristescu, Cvetkoski, Dale, Daniel, Davidson, de~Alencar, Dehouck, de~Laurentiis, de~Marneffe, de~Paiva, Derin, de~Souza, Diaz~de Ilarraza, D{\'{\i}}az~Hern{\'a}ndez, Dickerson, Dinakaramani, Di~Nuovo, Dione, Dirix, Do, Dobrovoljc, D{\"o}hmer, Doyle, Dozat, Droganova, Duran, Dwivedi, Ebert, Eckhoff, Eguchi, Eiche, Eiselen, Eli, Elkahky, Ephrem, Erina, Erjavec, Eslami, Essaidi, Etienne, Evelyn, Facundes, Farkas, Favero, Ferdaousi, Fernanda, Fernandez~Alcalde, Fethi, Foster, Fransen, Freitas, Fujita, Gajdo{\v s}ov{\'a}, Galbraith, Galy, Gamba, Garcia, G{\"a}rdenfors, Gaustad, Gen{\c c}, Gerardi, Gerdes, Gessler, Ginter, Godoy, Goenaga, Gojenola, G{\"o}k{\i}rmak, Goldberg, G{\'o}mez~Guinovart, Gonz{\'a}lez~Saavedra, Grici{\=u}t{\.e}, Grioni, Grobol, Gr{\= u}z{\={\i}}tis, Guillaume, Guiller, Guillot-Barbance, G{\"u}ng{\"o}r,
  Habash, Hafsteinsson, Haji{\v c}, Haji{\v c}~jr., H{\"a}m{\"a}l{\"a}inen, H{\`a}~M{\~y}, Han, Hanifmuti, Harada, Hardwick, Harris, Hassert, Haug, Heinecke, Hellwig, Hennig, Hladk{\'a}, Hlav{\'a}{\v c}ov{\'a}, Hociung, Hoefels, Hohle, Huang, Huerta~Mendez, Hwang, Ikeda, Iliadou, Ingason, Ion, Irimia, Ishola, Islamaj, Ito, Iurescia, Jagodzi{\'n}ska, Jannat, Jel{\'{\i}}nek, Jha, Jiang, Jobanputra, Johannsen, J{\'o}nsd{\'o}ttir, J{\o}rgensen, Juutinen, Ka{\c s}{\i}kara, Kabaeva, Kahane, Kanayama, Kanerva, Kara, Karah{\'o}ǧa, K{\aa}sen, Kayadelen, Kengatharaiyer, Kettnerov{\'a}, Kharatyan, Kirchner, Klementieva, Klyachko, Kocharov, K{\"o}hn, K{\"o}ksal, Kopacewicz, Korkiakangas, K{\"o}se, Koshevoy, Kotsyba, Kova{\v c}i{\'c}, Kovalevskait{\.e}, Krek, Krishnamurthy, K{\"u}bler, Kuqi, Kuyruk{\c c}u, Kuzgun, Kwak, Kyle, Laan, Laippala, Lambertino, Lando, Larasati, Lavrentiev, Lee, L{\^e}~H{\`{\^o}}ng, Lenci, Lertpradit, Leung, Levina, Levine, Li, Li, Li, Li, Li, Lim, Lima~Padovani, Lin, Lind{\'e}n, Liu, Ljube{\v
  s}i{\'c}, Lobzhanidze, Loginova, Lopes, Lusito, Lutgen, Luthfi, Luukko, Lyashevskaya, Lynn, Macketanz, Mahamdi, Maillard, Makarchuk, Makazhanov, Mambrini, Mandl, Manning, Manurung, Mar{\c s}an, M{\u a}r{\u a}nduc, Mare{\v c}ek, Marheinecke, Markantonatou, Mart{\'{\i}}nez~Alonso, Mart{\'{\i}}n~Rodr{\'{\i}}guez, Martins, Martins, Ma{\v s}ek, Matsuda, Matsumoto, Mazzei, {McDonald}, {McGuinness}, Mehta, M{\'e}nard, Mendon{\c c}a, Merzhevich, Meurer, Miekka, Milano, Miller, Mischenkova, Missil{\"a}, Mititelu, Mitrofan, Miyao, Mojiri~Foroushani, Moln{\'a}r, Moloodi, Montemagni, More, Moreno~Romero, Moretti, Mori, Morioka, Moro, Mortensen, Moskalevskyi, Muischnek, Munro, Murawaki, M{\"u}{\"u}risep, Nainwani, Nakhl{\'e}, Navarro~Hor{\~n}iacek, Nedoluzhko, Ne{\v s}pore-B{\=e}rzkalne, Nevaci, Nguy{\~{\^e}}n~Th{\d i}, Nguy{\~{\^e}}n Th{\d i}~Minh, Nikaido, Nikolaev, Nitisaroj, Norrman, Nourian, Nunes, Nurmi, Ojala, Ojha, {\'O}lad{\'o}ttir, Ol{\'u}{\`o}kun, Omura, Onwuegbuzia, Ordan, Osenova, {\"O}stling, Ott,
  {\O}vrelid, {\"O}zate{\c s}, {\"O}z{\c c}elik, {\"O}zg{\"u}r, {\"O}zt{\"u}rk~Ba{\c s}aran, Paccosi, Palmero~Aprosio, Panova, Pardo, Park, Partanen, Pascual, Passarotti, Patejuk, Paulino-Passos, Pedonese, Peljak-{\L}api{\'n}ska, Peng, Peng, Pereira, Pereira, Perez, Perkova, Perrier, Petrov, Petrova, Peverelli, Phelan, Pierre-Louis, Piitulainen, Pinter, Pinto, Pintucci, Pirinen, Pitler, Plamada, Plank, Plum, Poibeau, Ponomareva, Popel, Pretkalni{\c n}a, Pretorius, Pr{\'e}vost, Prokopidis, Przepi{\'o}rkowski, Pugh, Puolakainen, Purschke, Pyysalo, Qi, Querido, R{\"a}{\"a}bis, Rademaker, Rahoman, Rama, Ramasamy, Ramisch, Ramos, Rashel, Rasooli, Ravishankar, Real, Rebeja, Reddy, Regnault, Rehm, Riabi, Riabov, Rie{\ss}ler, Rimkut{\.e}, Rinaldi, Rituma, Rizqiyah, Rocha, R{\"o}gnvaldsson, Roksandic, Romanenko, Rosa, Roșca, Rovati, Rozonoyer, Rudina, Rueter, Ruffolo, R{\'u}narsson, Sadde, Safari, Sahala, Saleh, Salomoni, Samard{\v z}i{\'c}, Samson, S{\'a}nchez-Rodr{\'{\i}}guez, Sanguinetti, San{\i}yar, S{\"a}rg,
  Sartor, Sarymsakova, Sasaki, Saul{\={\i}}te, Savary, Sawanakunanon, Saxena, Scannell, Scarlata, Schang, Schneider, Schuster, Schwartz, Seddah, Seeker, Sellmer, Seraji, Shahzadi, Shen, Shimada, Shirasu, Shishkina, Shohibussirri, Shvedova, Siewert, Sigurðsson, Silva, Silveira, Silveira, Silveira, Simi, Simionescu, Simk{\'o}, {\v S}imkov{\'a}, S{\'{\i}}monarson, Simov, Sitchinava, Sither, Smith, Soares-Bastos, Solberg, Sonnenhauser, Sourov, Sprugnoli, Stamou, Steingr{\'{\i}}msson, Stella, Stephen, Straka, Strickland, Strnadov{\'a}, Suhr, Sulestio, Sulubacak, Suzuki, Swanson, Sz{\'a}nt{\'o}, Taguchi, Taji, Tamburini, Tan, Tanaka, Tanaya, Tavoni, Tella, Tellier, Testori, Thomas, T{\i}ra{\c s}, Tonelli, Torga, Toska, Trosterud, Trukhina, Tsarfaty, T{\"u}rk, Tyers, {\TH}{\'o}rðarson, {\TH}orsteinsson, Uematsu, Untilov, Ure{\v s}ov{\'a}, Uria, Uszkoreit, Utka, Vagnoni, Vajjala, Vak, van~der Goot, Vanhove, van Niekerk, van Noord, Varga, Vedenina, Venturi, Villemonte de~la Clergerie, Vincze, Vissamsetty, Vlasova,
  Vligouridou, Wakasa, Wallenberg, Wallin, Walsh, Wang, Washington, Wendt, Widmer, Wigderson, Wijono, Wille, Williams, Wir{\'e}n, Wittern, Woldemariam, Wong, Wr{\'o}blewska, Wu, Yako, Yamashita, Yamazaki, Yan, Yasuoka, Yavrumyan, Yenice, Y{\i}landilo{\u g}lu, Y{\i}ld{\i}z, Yu, Yuliawati, {\v Z}abokrtsk{\'y}, Zahra, Zeldes, Zhou, Zhu, Zhu, Zhuravleva, and Ziane}]{ud214}
Daniel Zeman, Joakim Nivre, Mitchell Abrams, Elia Ackermann, No{\"e}mi Aepli, Hamid Aghaei, {\v Z}eljko Agi{\'c}, Amir Ahmadi, Lars Ahrenberg, Chika~Kennedy Ajede, Salih~Furkan Akkurt, Gabriel{\.e} Aleksandravi{\v c}i{\=u}t{\.e}, Ika Alfina, Avner Algom, Khalid Alnajjar, Chiara Alzetta, Erik Andersen, Lene Antonsen, Tatsuya Aoyama, and 597 others. 2024.
\newblock \href {http://hdl.handle.net/11234/1-5502} {{Universal Dependencies 2.14}}.
\newblock {LINDAT}/{CLARIAH}-{CZ} digital library at the Institute of Formal and Applied Linguistics ({{\'U}FAL}), Faculty of Mathematics and Physics, Charles University.

\end{thebibliography}
\appendix
\section{Languages and Resources}
\label{sec:appendix-languages}

\begin{table*}
\adjustbox{max width=\textwidth}{%
\begin{tabular}{@{}lll@{}}
\toprule
\textbf{Resource} & \textbf{Resource URL} \\ \midrule
Universal Dependencies 2.14 \cite{ud214} & \https{hdl.handle.net/11234/1-5502} \\
SIB-200 \cite{adelani-etal-2024-sib} & \https{huggingface.co/datasets/Davlan/sib200}\\
Grambank v1.0.3 \cite{grambank_release} &  \https{zenodo.org/records/7844558}\\
Glottolog v5.0 \cite{glottolog2024} & \https{zenodo.org/records/10804582} \\
ASJP v17 \cite{asjp} & \https{asjp.clld.org} \\
\citet{jaeger2018extracting} & \https{osf.io/cufv7} \\
lang2vec v1.1.6 \citep{lang2vec_package} & \https{github.com/antonisa/lang2vec} \\
UDPipe~2 (2.12) \citep{udpipe212}  & \https{https://ufal.mff.cuni.cz/udpipe/2/models} \\
mBERT base uncased \cite{devlin-etal-2019-bert} & \https{huggingface.co/google-bert/bert-base-multilingual-uncased} \\
lme4 \cite{bates2015fitting} & \https{github.com/lme4/lme4} \\
scikit-learn 1.5 \cite{scikit-learn} & \https{scikit-learn.org} \\
uroman 1.3.1.1 \cite{hermjakob-etal-2018-box} & \https{github.com/isi-nlp/uroman} \\
GlotScript \cite{kargaran-etal-2024-glotscript} & \https{github.com/cisnlp/GlotScript} \\
SciPy \cite{2020SciPy-NMeth} & \https{scipy.org} \\
\bottomrule
\end{tabular}
}
\caption{\textbf{Details about the data, model, and software resources used in this paper.}}
\label{tab:resources}
\end{table*}

Table~\ref{tab:resources} lists the versions of the datasets, models, and software we use.
Lang2vec \cite{littell-etal-2017-uriel} contains information on syntax, phonology, and phoneme inventories from WALS \cite{wals}, SSWL \cite{sswl}, Ethnologue \cite{ethnologue2015}, and PHOIBLE \cite{phoible}.
SIB-200 \cite{adelani-etal-2024-sib}
is an annotated subset of FLORES-200 \citep{nllb-22}, which in turn builds on previous versions and extensions of FLORES \citep{goyal-etal-2022-flores, flores1-19, mt4nko-23, indictrans2-23}.
UDPipe~2 \citep{udpipe212} was trained on Universal Dependencies 2.12.\footnote{\url{http://hdl.handle.net/11234/1-5150}}
Its input representations include the last four layers of base-size uncased mBERT \cite{devlin-etal-2019-bert}.
This project uses the universal romanizer software `uroman' written by Ulf Hermjakob, USC Information Sciences Institute (2015-2020).

\begin{table*}
\centering
\begin{minipage}{.49\textwidth}%
\adjustbox{max width=\textwidth}{\begin{tabular}{@{}lllll@{}}
\toprule
\textbf{ISO} & \textbf{Name} & \textbf{Family} & \textbf{UD} & \textbf{SIB-200} \\ \midrule
abk & Abkhazian & Abkhaz-Adyge & ab\_abnc & -- \\
abq & Abaza & Abkhaz-Adyge & abq\_atb & -- \\
ace & Acehnese & Austronesian & -- & ace\_\{Arab, Latn\} \\
aeb (ara) & Tunisian Arabic & Afro-Asiatic & -- & aeb\_Arab \\
afr & Afrikaans & Indo-European & af\_afribooms & afr\_Latn \\
aii & Assyrian Neo-Aramaic & Afro-Asiatic & aii\_as & -- \\
ajp (apc, ara) & South Levantine Arabic & Afro-Asiatic & ajp\_madar & ajp\_Arab \\
aka & Akan & Atlantic-Congo & -- & aka\_Latn \\
akk & Akkadian & Afro-Asiatic & akk\_\{pisandub, riao\} & -- \\
aln & Gheg Albanian & Indo-European & aln\_gps & -- \\
als & Tosk Albanian & Indo-European & sq\_tsa & als\_Latn \\
amh & Amharic & Afro-Asiatic & am\_att & amh\_Ethi \\
apc (ara) & North Levantine Arabic & Afro-Asiatic & -- & apc\_Arab \\
apu & Apurinã & Arawakan & apu\_ufpa & -- \\
aqz & Akuntsu & Tupian & aqz\_tudet & -- \\
arb (ara) & Standard Arabic & Afro-Asiatic & ar\_\{padt, pud\} & arb\_\{Arab, Latn\} \\
arr & Karo & Tupian & arr\_tudet & -- \\
ary (ara) & Moroccan Arabic & Afro-Asiatic & -- & ary\_Arab \\
arz (ara) & Egyptian Arabic & Afro-Asiatic & -- & arz\_Arab \\
asm & Assamese & Indo-European & -- & asm\_Beng \\
ast & Asturian & Indo-European & -- & ast\_Latn \\
awa & Awadhi & Indo-European & -- & awa\_Deva \\
ayr & Central Aymara & Aymaran & -- & ayr\_Latn \\
azb (aze) & South Azerbaijani & Turkic & -- & azb\_Arab \\
aze & Azerbaijani & Turkic & az\_tuecl & -- \\
azj (aze) & North Azerbaijani & Turkic & -- & azj\_Latn \\
azz & Highland Puebla Nahuatl & Uto-Aztecan & azz\_itml & -- \\
bak & Bashkir & Turkic & -- & bak\_Cyrl \\
bam & Bambara & Mande & bm\_crb & bam\_Latn \\
ban & Balinese & Austronesian & -- & ban\_Latn \\
bar & Bavarian & Indo-European & bar\_maibaam & -- \\
bej & Bedawiyet & Afro-Asiatic & bej\_nsc & -- \\
bel & Belarusian & Indo-European & be\_hse & bel\_Cyrl \\
bem & Bemba & Atlantic-Congo & -- & bem\_Latn \\
ben & Bengali & Indo-European & bn\_bru & ben\_Beng \\
bho & Bhojpuri & Indo-European & bho\_bhtb & bho\_Deva \\
bjn & Banjar & Austronesian & -- & bjn\_\{Arab, Latn\} \\
bod & Tibetan & Sino-Tibetan & -- & bod\_Tibt \\
bor & Borôro & Bororoan & bor\_bdt & -- \\
bos & Bosnian & Indo-European & -- & bos\_Latn \\
bre & Breton & Indo-European & br\_keb & -- \\
bug & Buginese & Austronesian & -- & bug\_Latn \\
bul & Bulgarian & Indo-European & bg\_btb & bul\_Cyrl \\
bxr (bua) & Russia Buriat & Mongolic-Khitan & bxr\_bdt & -- \\
cat & Catalan & Indo-European & ca\_ancora & cat\_Latn \\
ceb & Cebuano & Austronesian & ceb\_gja & ceb\_Latn \\
ces & Czech & Indo-European & \begin{tabular}[t]{@{}l@{}}cs\_\{cac, cltt, fictree,\\ pdt, poetry, pud\}\end{tabular} & ces\_Latn \\
chu & Old Church Slavonic & Indo-European & cu\_proiel & -- \\
cjk & Chokwe & Atlantic-Congo & -- & cjk\_Latn \\
ckb & Central Kurdish & Indo-European & -- & ckb\_Arab \\
ckt & Chukot & Ch.-Kamchatkan & ckt\_hse & -- \\
cmn (zho) & Mandarin Chinese & Sino-Tibetan & \begin{tabular}[t]{@{}l@{}}zh\_\{beginner, cfl,\\ gsd, gsdsimp, hk,\\ patentchar, pud\}\end{tabular} & zho\_\{Hans, Hant\} \\
cop & Coptic & Afro-Asiatic & cop\_scriptorium & -- \\
cpg & Cappadocian Greek & Indo-European & cpg\_tuecl & -- \\
crh & Crimean Tatar & Turkic & -- & crh\_Latn \\
cym & Welsh & Indo-European & cy\_ccg & cym\_Latn \\
dan & Danish & Indo-European & da\_ddt & dan\_Latn \\
deu & German & Indo-European & de\_\{gsd, hdt, lit, pud\} & deu\_Latn \\
dik & Southwestern Dinka & Nilotic & -- & dik\_Latn \\
dyu & Dyula & Mande & -- & dyu\_Latn \\
dzo & Dzongkha & Sino-Tibetan & -- & dzo\_Tibt \\
egy & Ancient Egyptian & Afro-Asiatic & egy\_ujaen & -- \\
ekk (est) & Standard Estonian & Uralic & et\_\{edt, ewt\} & est\_Latn \\
ell & Modern Greek & Indo-European & el\_\{gdt, gud\} & ell\_Grek \\
eme & Emerillon & Tupian & eme\_tudet & -- \\
eng & English & Indo-European & \begin{tabular}[t]{@{}l@{}}en\_\{atis, ctetex, eslspok,\\ ewt, gentle, gum, lines,\\ partut, pronouns, pud\}\end{tabular} & eng\_Latn \\
epo & Esperanto & (Constructed) & -- & epo\_Latn \\
ess & Central Siberian Yupik & Eskimo-Aleut & ess\_sli & -- \\
eus & Basque & (Isolate) & eu\_bdt & eus\_Latn \\
ewe & Ewe & Atlantic-Congo & -- & ewe\_Latn \\
fao & Faroese & Indo-European & fo\_\{farpahc, oft\} & fao\_Latn \\
fij & Fijian & Austronesian & -- & fij\_Latn \\
fin & Finnish & Uralic & fi\_\{ftb, ood, pud, tdt\} & fin\_Latn \\
fon & Fon & Atlantic-Congo & -- & fon\_Latn \\
fra & French & Indo-European & \begin{tabular}[t]{@{}l@{}}fr\_\{fqb, gsd,\\ parisstories, partut,\\ pud, rhapsodie, sequoia\}\end{tabular} & fra\_Latn \\
frm & Middle French & Indo-European & frm\_profiterole & -- \\
fro & Old French & Indo-European & fro\_profiterole & -- \\
fur & Friulian & Indo-European & -- & fur\_Latn \\
fuv & Nigerian Fulfulde & Atlantic-Congo & -- & fuv\_Latn \\
gaz & West Central Oromo & Afro-Asiatic & -- & gaz\_Latn \\
gla & Gaelic & Indo-European & gd\_arcosg & gla\_Latn \\
gle & Irish & Indo-European & ga\_\{cadhan, idt, twittirish\} & gle\_Latn \\
glg & Galician & Indo-European & gl\_\{ctg, pud, treegal\} & glg\_Latn \\
glv & Manx & Indo-European & gv\_cadhan & -- \\
got & Gothic & Indo-European & got\_proiel & -- \\
grc & Ancient Greek & Indo-European & grc\_\{perseus, proiel, ptnk\} & -- \\
grn & Guarani & Tupian & gn\_oldtudet & grn\_Latn \\
gsw & Swiss German & Indo-European & gsw\_uzh & -- \\
gub & Guajajára & Tupian & gub\_tudet & -- \\
guj & Gujarati & Indo-European & gu\_gujtb & guj\_Gujr \\
gun & Mbyá Guaraní & Tupian & gun\_thomas & -- \\
hat & Haitian & Indo-European & ht\_autogramm & hat\_Latn \\
hau & Hausa & Afro-Asiatic & \begin{tabular}[t]{@{}l@{}}ha\_\{northernautogramm,\\ southernautogramm\}\end{tabular} & hau\_Latn \\
\bottomrule
\end{tabular}
}
\end{minipage}
\begin{minipage}{.49\textwidth}%
\adjustbox{max width=\textwidth}{\begin{tabular}{@{}lllll@{}}
\toprule
\textbf{ISO} & \textbf{Name} & \textbf{Family} & \textbf{UD} & \textbf{SIB-200} \\ \midrule

hbo & Ancient Hebrew & Afro-Asiatic & hbo\_ptnk & -- \\
heb & Hebrew & Afro-Asiatic & he\_\{htb, iahltwiki\} & heb\_Hebr \\
hin & Hindi & Indo-European & hi\_\{hdtb, pud\} & hin\_Deva \\
hit & Hittite & Indo-European & hit\_hittb & -- \\
hne & Chhattisgarhi & Indo-European & -- & hne\_Deva \\
hrv & Croatian & Indo-European & hr\_set & hrv\_Latn \\
hsb & Upper Sorbian & Indo-European & hsb\_ufal & -- \\
hun & Hungarian & Uralic & hu\_szeged & hun\_Latn \\
hye & Armenian & Indo-European & hy\_\{armtdp, bsut\} & hye\_Armn \\
hyw & Western Armenian & Indo-European & hyw\_armtdp & -- \\
ibo & Igbo & Atlantic-Congo & -- & ibo\_Latn \\
ilo & Iloko & Austronesian & -- & ilo\_Latn \\
ind & Indonesian & Austronesian & id\_\{csui, gsd, pud\} & ind\_Latn \\
isl & Icelandic & Indo-European & is\_\{gc, icepahc, modern, pud\} & isl\_Latn \\
ita & Italian & Indo-European & \begin{tabular}[t]{@{}l@{}}it\_\{isdt, markit, old,\\ parlamint, partut, postwita,\\ pud, twittiro, valico, vit\}\end{tabular} & ita\_Latn \\
jaa & Madí & Arawan & jaa\_jarawara & -- \\
jav & Javanese & Austronesian & jv\_csui & jav\_Latn \\
jpn & Japanese & Japonic & ja\_\{gsd, gsdluw, pud, pudluw\} & jpn\_Jpan \\
kab & Kabyle & Afro-Asiatic & -- & kab\_Latn \\
kac & Jingpho & Sino-Tibetan & -- & kac\_Latn \\
kam & Kamba & Atlantic-Congo & -- & kam\_Latn \\
kan & Kannada & Dravidian & -- & kan\_Knda \\
kas & Kashmiri & Indo-European & -- & kas\_\{Arab, Deva\} \\
kat & Georgian & Kartvelian & ka\_glc & kat\_Geor \\
kaz & Kazakh & Turkic & kk\_ktb & kaz\_Cyrl \\
kbp & Kabiyè & Atlantic-Congo & -- & kbp\_Latn \\
kea & Kabuverdianu & Indo-European & -- & kea\_Latn \\
khk & Halh Mongolian & Mongolic-Khitan & -- & khk\_Cyrl \\
khm & Central Khmer & Austroasiatic & -- & khm\_Khmr \\
kik & Gikuyu & Atlantic-Congo & -- & kik\_Latn \\
kin & Kinyarwanda & Atlantic-Congo & -- & kin\_Latn \\
kir & Kyrgyz & Turkic & ky\_\{ktmu, tuecl\} & kir\_Cyrl \\
kmb & Kimbundu & Atlantic-Congo & -- & kmb\_Latn \\
kmr & Kurmanji & Indo-European & kmr\_mg & kmr\_Latn \\
knc & Central Kanuri & Saharan & -- & knc\_\{Arab, Latn\} \\
koi & Komi-Permyak & Uralic & koi\_uh & -- \\
kon & Kongo & Atlantic-Congo & -- & kon\_Latn \\
kor & Korean & Koreanic & ko\_\{gsd, kaist, pud\} & kor\_Hang \\
kpv & Komi-Zyrian & Uralic & kpv\_\{ikdp, lattice\} & -- \\
krl & Karelian & Uralic & krl\_kkpp & -- \\
lao & Lao & Tai-Kadai & -- & lao\_Laoo \\
lat & Latin & Indo-European & \begin{tabular}[t]{@{}l@{}}la\_\{circse, ittb, llct, perseus,\\ proiel, udante\}\end{tabular} & -- \\
lij & Ligurian & Indo-European & lij\_glt & lij\_Latn \\
lim & Limburgan & Indo-European & -- & lim\_Latn \\
lin & Lingala & Atlantic-Congo & -- & lin\_Latn \\
lit & Lithuanian & Indo-European & lt\_\{alksnis, hse\} & lit\_Latn \\
lmo & Lombard & Indo-European & -- & lmo\_Latn \\
ltg & Latgalian & Indo-European & ltg\_cairo & ltg\_Latn \\
ltz & Luxembourgish & Indo-European & lb\_luxbank & ltz\_Latn \\
lua & Luba-Lulua & Atlantic-Congo & -- & lua\_Latn \\
lug & Ganda & Atlantic-Congo & -- & lug\_Latn \\
luo & Dholuo & Nilotic & -- & luo\_Latn \\
lus & Lushai & Sino-Tibetan & -- & lus\_Latn \\
lvs (lav) & Standard Latvian & Indo-European & lv\_\{cairo, lvtb\} & lvs\_Latn \\
lzh & Classical Chinese & Sino-Tibetan & lzh\_\{kyoto, tuecl\} & -- \\
mag & Magahi & Indo-European & -- & mag\_Deva \\
mai & Maithili & Indo-European & -- & mai\_Deva \\
mal & Malayalam & Dravidian & ml\_ufal & mal\_Mlym \\
mar & Marathi & Indo-European & mr\_ufal & mar\_Deva \\
mdf & Moksha & Uralic & mdf\_jr & -- \\
min & Minangkabau & Austronesian & -- & min\_\{Arab, Latn\} \\
mkd & Macedonian & Indo-European & mk\_mtb & mkd\_Cyrl \\
mlt & Maltese & Afro-Asiatic & mt\_mudt & mlt\_Latn \\
mni & Manipuri & Sino-Tibetan & -- & mni\_Beng \\
mos & Mossi & Atlantic-Congo & -- & mos\_Latn \\
mpu & Makuráp & Tupian & mpu\_tudet & -- \\
mri & Maori & Austronesian & -- & mri\_Latn \\
mya & Burmese & Sino-Tibetan & -- & mya\_Mymr \\
myu & Mundurukú & Tupian & myu\_tudet & -- \\
myv & Erzya & Uralic & myv\_jr & -- \\
nds & Low Saxon & Indo-European & nds\_lsdc & -- \\
nhi & W.\ S.\ Puebla Nahuatl & Uto-Aztecan & nhi\_itml & -- \\
nld & Dutch & Indo-European & nl\_\{alpino, lassysmall\} & nld\_Latn \\
nno (nor) & Norwegian Nynorsk & Indo-European & no\_nynorsk & nno\_Latn \\
nob (nor) & Norwegian Bokmål & Indo-European & no\_bokmaal & nob\_Latn \\
npi & Nepali & Indo-European & -- & npi\_Deva \\
nqo & N'Ko & (Constructed) & -- & nqo\_Nkoo \\
nso & Northern Sotho & Atlantic-Congo & -- & nso\_Latn \\
nus & Nuer & Nilotic & -- & nus\_Latn \\
nya & Chewa & Atlantic-Congo & -- & nya\_Latn \\
oci & Occitan & Indo-European & -- & oci\_Latn \\
olo & Livvi & Uralic & olo\_kkpp & -- \\
orv & Old East Slavic & Indo-European & \begin{tabular}[t]{@{}l@{}}orv\_\{birchbark, rnc,\\ ruthenian, torot\}\end{tabular} & -- \\
ory & Odia & Indo-European & -- & ory\_Orya \\
ota & Ottoman Turkish & Turkic & ota\_\{boun, dudu\} & -- \\
otk & Old Turkish & Turkic & otk\_clausal & -- \\
pad & Paumarí & Arawan & pad\_tuecl & -- \\
pag & Pangasinan & Austronesian & -- & pag\_Latn \\
pan & Panjabi & Indo-European & -- & pan\_Guru \\
pap & Papiamento & Indo-European & -- & pap\_Latn \\
pbt & Southern Pashto & Indo-European & -- & pbt\_Arab \\
pcm & Nigerian Pidgin & Indo-European & pcm\_nsc & -- \\
pes (fas) & Iranian Persian & Indo-European & fa\_\{perdt, seraji\} & pes\_Arab \\
\bottomrule
\end{tabular}
}
\end{minipage}%
\caption{\textbf{Languages and datasets used in our experiments.} Continued and explained in next table.}
\label{tab:languages-i}
\end{table*}
\begin{table*}
\centering
\begin{minipage}{.49\textwidth}%
\adjustbox{max width=\textwidth}{\begin{tabular}{@{}lllll@{}}
\toprule
\textbf{ISO} & \textbf{Name} & \textbf{Family} & \textbf{UD} & \textbf{SIB-200} \\ \midrule
plt & Plateau Malagasy & Austronesian & -- & plt\_Latn \\
pol & Polish & Indo-European & pl\_\{lfg, pdb, pud\} & pol\_Latn \\
por & Portuguese & Indo-European & \begin{tabular}[t]{@{}l@{}}pt\_\{bosque, cintil, gsd,\\ petrogold, porttinari, pud\}\end{tabular} & por\_Latn \\
prs (fas) & Dari & Indo-European & -- & prs\_Arab \\
qpm (bul) & Pomak & Indo-European & qpm\_philotis & -- \\
quc & K'iche' & Mayan & quc\_iu & -- \\
quy & Ayacucho Quechua & Quechuan & -- & quy\_Latn \\
ron & Romanian & Indo-European & \begin{tabular}[t]{@{}l@{}}ro\_\{art, nonstandard,\\ rrt, simonero, tuecl\}\end{tabular} & ron\_Latn \\
run & Rundi & Atlantic-Congo & -- & run\_Latn \\
rus & Russian & Indo-European & \begin{tabular}[t]{@{}l@{}}ru\_\{gsd, poetry,\\ pud, syntagrus, taiga\}\end{tabular} & rus\_Cyrl \\
sag & Sango & Atlantic-Congo & -- & sag\_Latn \\
sah & Yakut & Turkic & sah\_yktdt & -- \\
san & Sanskrit & Indo-European & sa\_\{ufal, vedic\} & san\_Deva \\
sat & Santali & Austroasiatic & -- & sat\_Olck \\
say & Saya & Afro-Asiatic & say\_autogramm & -- \\
scn & Sicilian & Indo-European & -- & scn\_Latn \\
sga & Old Irish & Indo-European & sga\_\{dipsgg, dipwbg\} & -- \\
shn & Shan & Tai-Kadai & -- & shn\_Mymr \\
sin & Sinhala & Indo-European & si\_stb & sin\_Sinh \\
sjo & Xibe & Tungusic & sjo\_xdt & -- \\
slk & Slovak & Indo-European & sk\_snk & slk\_Latn \\
slv & Slovenian & Indo-European & sl\_\{ssj, sst\} & slv\_Latn \\
sme & Northern Sami & Uralic & sme\_giella & -- \\
smo & Samoan & Austronesian & -- & smo\_Latn \\
sms & Skolt Sami & Uralic & sms\_giellagas & -- \\
sna & Shona & Atlantic-Congo & -- & sna\_Latn \\
snd & Sindhi & Indo-European & -- & snd\_Arab \\
som & Somali & Afro-Asiatic & -- & som\_Latn \\
sot & Southern Sotho & Atlantic-Congo & -- & sot\_Latn \\
spa & Castilian & Indo-European & es\_\{ancora, coser, gsd, pud\} & spa\_Latn \\
srd & Sardinian & Indo-European & -- & srd\_Latn \\
srp & Serbian & Indo-European & sr\_set & srp\_Cyrl \\
ssw & Swati & Atlantic-Congo & -- & ssw\_Latn \\
sun & Sundanese & Austronesian & -- & sun\_Latn \\
swe & Swedish & Indo-European &  \begin{tabular}[t]{@{}l@{}}sv\_\{lines,\\ pud, talbanken\}\end{tabular} & swe\_Latn \\
swh & Swahili & Atlantic-Congo & -- & swh\_Latn \\
szl & Silesian & Indo-European & -- & szl\_Latn \\
tam & Tamil & Dravidian & ta\_\{mwtt, ttb\} & tam\_Taml \\
\bottomrule
\end{tabular}
}
\end{minipage}
\begin{minipage}{.49\textwidth}%
\adjustbox{max width=\textwidth}{\begin{tabular}{@{}lllll@{}}
\toprule
\textbf{ISO} & \textbf{Name} & \textbf{Family} & \textbf{UD} & \textbf{SIB-200} \\ \midrule
taq & Tamasheq & Afro-Asiatic & -- & taq\_\{Latn, Tfng\} \\
tat & Tatar & Turkic & tt\_nmctt & tat\_Cyrl \\
tel & Telugu & Dravidian & te\_mtg & tel\_Telu \\
tgk & Tajik & Indo-European & -- & tgk\_Cyrl \\
tgl & Tagalog & Austronesian & tl\_\{trg, ugnayan\} & tgl\_Latn \\
tha & Thai & Tai-Kadai & th\_pud & tha\_Thai \\
tir & Tigrinya & Afro-Asiatic & -- & tir\_Ethi \\
tpi & Tok Pisin & Indo-European & -- & tpi\_Latn \\
tpn & Tupinambá & Tupian & tpn\_tudet & -- \\
tsn & Tswana & Atlantic-Congo & tn\_popapolelo & tsn\_Latn \\
tso & Tsonga & Atlantic-Congo & -- & tso\_Latn \\
tuk & Turkmen & Turkic & -- & tuk\_Latn \\
tum & Tumbuka & Atlantic-Congo & -- & tum\_Latn \\
tur & Turkish & Turkic & \begin{tabular}[t]{@{}l@{}}tr\_\{atis, boun,\\ framenet, gb, kenet,\\ penn, pud, tourism\}\end{tabular} & tur\_Latn \\
twi & Twi & Atlantic-Congo & -- & twi\_Latn \\
tzm & C.\ Atlas Tamazight & Afro-Asiatic & -- & tzm\_Tfng \\
uig & Uyghur & Turkic & ug\_udt & uig\_Arab \\
ukr & Ukrainian & Indo-European & uk\_iu & ukr\_Cyrl \\
umb & Umbundu & Atlantic-Congo & -- & umb\_Latn \\
urb & Kaapor & Tupian & urb\_tudet & -- \\
urd & Urdu & Indo-European & ur\_udtb & urd\_Arab \\
uzn & Northern Uzbek & Turkic & -- & uzn\_Latn \\
vec & Venetian & Indo-European & -- & vec\_Latn \\
vep & Veps & Uralic & vep\_vwt & -- \\
vie & Vietnamese & Austroasiatic & vi\_\{tuecl, vtb\} & vie\_Latn \\
war & Waray & Austronesian & -- & war\_Latn \\
wbp & Warlpiri & Pama-Nyungan & wbp\_ufal & -- \\
wol & Wolof & Atlantic-Congo & wo\_wtb & wol\_Latn \\
xav & Xavánte & Nuclear-Macro-Je & xav\_xdt & -- \\
xcl & Classical Armenian & Indo-European & xcl\_caval & -- \\
xho & Xhosa & Atlantic-Congo & -- & xho\_Latn \\
xnr & Kangri & Indo-European & xnr\_kdtb & -- \\
xum & Umbrian & Indo-European & xum\_ikuvina & -- \\
ydd (yid) & Eastern Yiddish & Indo-European & -- & ydd\_Hebr \\
yor & Yoruba & Atlantic-Congo & yo\_ytb & yor\_Latn \\
yrl & Nhengatu & Tupian & yrl\_complin & -- \\
yue & Yue Chinese & Sino-Tibetan & yue\_hk & yue\_Hant \\
\multicolumn{2}{@{}l}{zsm (zlm, msa) Std. Malay} & Austronesian & -- & zsm\_Latn \\
zul & Zulu & Atlantic-Congo & -- & zul\_Latn \\ \bottomrule
\end{tabular}
}
\end{minipage}%
\caption{\textbf{Languages and datasets used in our experiments, continued.} ISO 639-3 codes in parentheses denote macrolanguage codes. Language family information is sourced from Glottolog \cite{glottolog2024}.}
\label{tab:languages-ii}
\end{table*}

Tables~\ref{tab:languages-i} and~\ref{tab:languages-ii} show the languages and datasets included in our experiments.
Their geographic distribution is pictured in Figure~\ref{fig:map-languages-datasets}.

\subsection{Excluded UD treebanks}
\label{sec:appendix-excluded-treebanks}

We exclude treebanks with code-switched language data (\dataset{qaf}{arabizi}, \dataset{qfn}{fame}, \dataset{qtd}{sagt}, \dataset{qte}{tect}), without text (\dataset{ar}{nyuad}, \dataset{en}{gumreddit}, \dataset{gun}{dooley}, \dataset{ja}{bccwj}, \dataset{ja}{bccwjluw}), with glossed language (\dataset{swl}{sslc}), and where the division into training and test split changed between releases 2.12 and 2.14 (\dataset{tr}{imst}).
We also exclude test sets with fewer than 20 sentences (\dataset{kfm}{aha}, \dataset{nap}{rb}, \dataset{nyq}{aha}, \dataset{soj}{aha}) and training sets with less than 100 sentences (\dataset{bxr}{bdt}, \dataset{hsb}{ufal}, \dataset{kk}{ktb}, \dataset{kmr}{mg}, \dataset{lij}{glt}, \dataset{olo}{kkpp}).
We additionally exclude other training datasets that were not included in UDPipe~2 (\dataset{ky}{ktmu}, \dataset{de}{lit}, \dataset{de}{pud}).
Finally, we exclude the Czech UDPipe-2 models, as they use embeddings from a pretrained Czech model rather than mBERT.

\subsection{Excluded SIB-200 languages}
\label{sec:appendix-excluded-sib200}

We exclude three Arabic dialects whose sentences in SIB-200 are almost identical to the Modern Standard Arabic (\dataset{arb}{Arab}) sentences: \dataset{ars}{Arab}, \dataset{acm}{Arab}, \dataset{acq}{Arab}. This is a known issue for FLORES-200, from which SIB-200 is derived.%
\footnote{\https{github.com/openlanguagedata/flores/issues/8}}

For the transliteration experiment, we exclude Japanese (\dataset{jpn}{Jpan}) since uroman is not able to properly transliterate kanji as of version 1.3.1.1.
We also exclude Mandarin with traditional characters (\dataset{zho}{Hant}) and Cantonese (\dataset{yue}{Hant}) since uroman crashes when trying to transliterate the corresponding datasets.

\begin{table}
\adjustbox{max width=\linewidth}{%
\begin{tabular}{lrrrrrrr}
\toprule
& \grambank & \syntax & \phono & \inventory & \asjp & \genetic & \geo \\
\midrule
\syntax & \cellcolor[HTML]{6464ff} \color{black!10} 0.61  \\
\phono & \cellcolor[HTML]{b8b8ff} 0.28 & \cellcolor[HTML]{b4b4ff} 0.30  \\
\inventory & \cellcolor[HTML]{c6c6ff} 0.22 & \cellcolor[HTML]{d6d6ff} 0.16 & \cellcolor[HTML]{b4b4ff} 0.30 \\
\asjp & \cellcolor[HTML]{6e6eff} \color{black!10} 0.57 &\cellcolor[HTML]{8a8aff} 0.45 & \cellcolor[HTML]{b2b2ff} 0.30 & \cellcolor[HTML]{b6b6ff} 0.29 \\
\genetic & \cellcolor[HTML]{7070ff} \color{black!10} 0.56 &\cellcolor[HTML]{8686ff} 0.47 & \cellcolor[HTML]{b0b0ff} 0.31 & \cellcolor[HTML]{bebeff} 0.25 & \cellcolor[HTML]{2222ff} \color{black!10} 0.87 \\
\geo & \cellcolor[HTML]{9292ff} 0.42 & \cellcolor[HTML]{a6a6ff} 0.35 & \cellcolor[HTML]{9c9cff} 0.39 & \cellcolor[HTML]{e6e6ff} 0.10 & \cellcolor[HTML]{a2a2ff} 0.36 & \cellcolor[HTML]{9898ff} 0.40 \\
\midrule
\overlapword{} (UD) & \cellcolor[HTML]{8e8eff} 0.44 & \cellcolor[HTML]{a6a6ff} 0.35 & \cellcolor[HTML]{c0c0ff} 0.24 & \cellcolor[HTML]{9a9aff} 0.39 & \cellcolor[HTML]{6868ff} \color{black!10} 0.59 &\cellcolor[HTML]{8888ff} 0.47 & \cellcolor[HTML]{d4d4ff} 0.17 \\
\overlapsubword{} (topics) & \cellcolor[HTML]{b2b2ff} 0.30 & \cellcolor[HTML]{b8b8ff} 0.28 & \cellcolor[HTML]{d8d8ff} 0.15 & \cellcolor[HTML]{9494ff} 0.42 & \cellcolor[HTML]{9c9cff} 0.39 & \cellcolor[HTML]{9e9eff} 0.38 & \cellcolor[HTML]{e6e6ff} 0.10 \\
\overlaptri{} (topics) & \cellcolor[HTML]{d6d6ff} 0.16 & \cellcolor[HTML]{c2c2ff} 0.24 & \cellcolor[HTML]{e2e2ff} 0.12 & \cellcolor[HTML]{a0a0ff} 0.37 & \cellcolor[HTML]{c2c2ff} 0.24 & \cellcolor[HTML]{c2c2ff} 0.24 & \cellcolor[HTML]{fffcfc} -0.01 \\
\overlaptri{} (top.\ tr.) & \cellcolor[HTML]{b6b6ff} 0.29 & \cellcolor[HTML]{c0c0ff} 0.25 & \cellcolor[HTML]{ccccff} 0.20 & \cellcolor[HTML]{9e9eff} 0.38 & \cellcolor[HTML]{acacff} 0.33 & \cellcolor[HTML]{a8a8ff} 0.34 & \cellcolor[HTML]{f2f2ff} 0.05 \\
\overlapchar{} (UD) & \cellcolor[HTML]{bcbcff} 0.26 & \cellcolor[HTML]{9e9eff} 0.38 & \cellcolor[HTML]{a6a6ff} 0.35 & \cellcolor[HTML]{b8b8ff} 0.28 & \cellcolor[HTML]{b4b4ff} 0.29 & \cellcolor[HTML]{b8b8ff} 0.28 & \cellcolor[HTML]{c4c4ff} 0.23 \\
\overlapchar{} (topics) & \cellcolor[HTML]{e4e4ff} 0.10 & \cellcolor[HTML]{d4d4ff} 0.17 & \cellcolor[HTML]{e0e0ff} 0.12 & \cellcolor[HTML]{babaff} 0.27 & \cellcolor[HTML]{dadaff} 0.14 & \cellcolor[HTML]{d6d6ff} 0.16 & --- \\
\overlapchar{} (top.\ tr.) & \cellcolor[HTML]{f0f0ff} 0.06 & \cellcolor[HTML]{e6e6ff} 0.10 & \cellcolor[HTML]{fff6f6} -0.03 & \cellcolor[HTML]{d6d6ff} 0.16 & \cellcolor[HTML]{f4f4ff} 0.04 & \cellcolor[HTML]{eeeeff} 0.06 & --- \\
\bottomrule
\end{tabular}
}
\caption{\textbf{Correlations (Pearson's \textit{r}) between linguistic similarity measures (top) and between dataset similarity and linguistic similarity measures (bottom).} Correlations with \textit{p}-values >= 0.05 are replaced with ---. 
Correlations between linguistic similarity measures are for the full set of languages included in our study.
Key: %
\grambank=Grambank similarity, \syntax=syntactic similarity (lang2vec), \phono=phonological similarity, \inventory=similarity of phoneme inventories, \asjp=lexical similarity, \genetic=phylogenetic relatedness, \geo=geographic proximity, \overlapword=word overlap, \overlaptri=character trigram overlap, \overlapchar=character overlap, 
top.\ tr.=transliterated topic classification data.
}
\label{tab:correlations-between-similarity-measures}
\end{table}

\section{Correlations between similarity measures}
\label{sec:appendix-correlations-between-measures}

Table~\ref{tab:correlations-between-similarity-measures} shows how the different dataset-independent similarity measures are correlated with each other (top) and with the dataset-dependent similarity measures (bottom).

\section{Hyperparameters and Standard Deviations of Topic Classification Models}
\label{sec:appendix-topic-models}

\begin{table}
\centering
\adjustbox{max width=\linewidth}{%
\begin{tabular}{@{}lll@{}}
\toprule
\textbf{Parameter} & \textbf{MLP ($n$-grams)} & \textbf{MLP (mBERT)} \\
\midrule
\textit{N-grams}\\
   Min. length  & \textbf{1} & --- \\
   Max. length  & 2, 3, \textbf{4} & --- \\
   Type & char, \textbf{char\_wb} & --- \\
   Max. features & 5000, \textbf{unlimited} & --- \\
   \midrule
   Max. epochs & 5, 10, \textbf{20}, 200, 300, 400 & 5, 10, \textbf{20}, 30, 50, 200 \\
   Learning rate & 0.0005, \textbf{0.001}, 0.002 & 0.0005, \textbf{0.001}, 0.002 \\
   Optimizer & \textbf{adam} & \textbf{adam} \\
   \bottomrule
\end{tabular}
}
\caption{\textbf{Hyperparameters used in the grid search for the topic classification models.} Values in \textbf{bold} are the ones used in the final models.}
\label{tab:hyperparams}
\end{table}

Table~\ref{tab:hyperparams} shows which hyperparameter values we included in the grid search for the topic classification models (the $n$-gram model trained and evaluated on the non-transliterated data, and the model using mBERT-based representations).
We used the average development set scores for models (monolingually) evaluated on the following languages for hyperparameter tuning: 
\dataset{arb}{Arab}, 
\dataset{ayr}{Latn},
\dataset{eng}{Latn}, 
\dataset{eus}{Latn}, 
\dataset{grn}{Latn}, 
\dataset{kan}{Knda},
\dataset{kat}{Geor}, 
\dataset{kor}{Hang}, 
\dataset{quy}{Latn}, 
\dataset{vie}{Latn},  
\dataset{zho}{Hans}.

We calculate standard deviations for the final hyperparameter selection and the above-mentioned languages (evaluated monolingually on their development sets) over five random seeds.
For the $n$-gram model, the standard deviations are between $0.0088$ (\dataset{ayr}{Latn}) and $0.0286$ (\dataset{kat}{Geor}), with a mean of $0.0181$.
For the mBERT-based model, the standard deviations are between $0.0081$ (\dataset{eng}{Latn}) and $0.0244$ (\dataset{grn}{Latn}), with a mean of $0.0160$.

\section{NLP Task Results}
\label{sec:appendix-results-and-heatmaps}

\subsection{Random baselines}
\label{sec:appendix-random-baselines}

For POS tagging and topic classification, we consider random baselines that randomly predict one of the 17~POS tags (or one of the seven topics) and are thus correct 5.9\,\% (or 14.3\,\%) of the time.

For each setup, all within-language performances (the diagonals in Figure~\ref{fig:heatmaps-intext}) are above this threshold.
Thus, all models learned something about the task, regardless of the training language and could be expected to be able to transfer some of that knowledge to a well-suited test language.

For some of the train–test language combinations, we see performances that are worse than random chance.
This is the case for 3.1\,\% of the POS tagging results, 26.6\,\% of \topicsbase{}'s results, 14.5\,\% of \topicstranslit{}'s results, and 6.8\,\% of the results by \topicsmbert. 
These worse-than-random transfer results are also meaningful: a POS tagger (or parser) trained on one language might learn grammatical patterns that run counter to how another language works, or a topic classification model might be mislead by superficial string overlaps between datasets that are false friends.

\subsection{Heatmaps of transfer results}
\label{sec:appendix-heatmaps}

We include large, labelled versions of the heatmaps in Figure~\ref{fig:heatmaps-intext}.
We order the languages in each heatmap by clustering its rows (target languages) produced using \citeposs{ward1963hierarchical} method, and applying the same order to the source languages.
Because the heatmaps take up a lot of space, they are placed at the end of the appendix.

\begin{figure*}
    \centering
    \includegraphics[height=0.9\textheight]{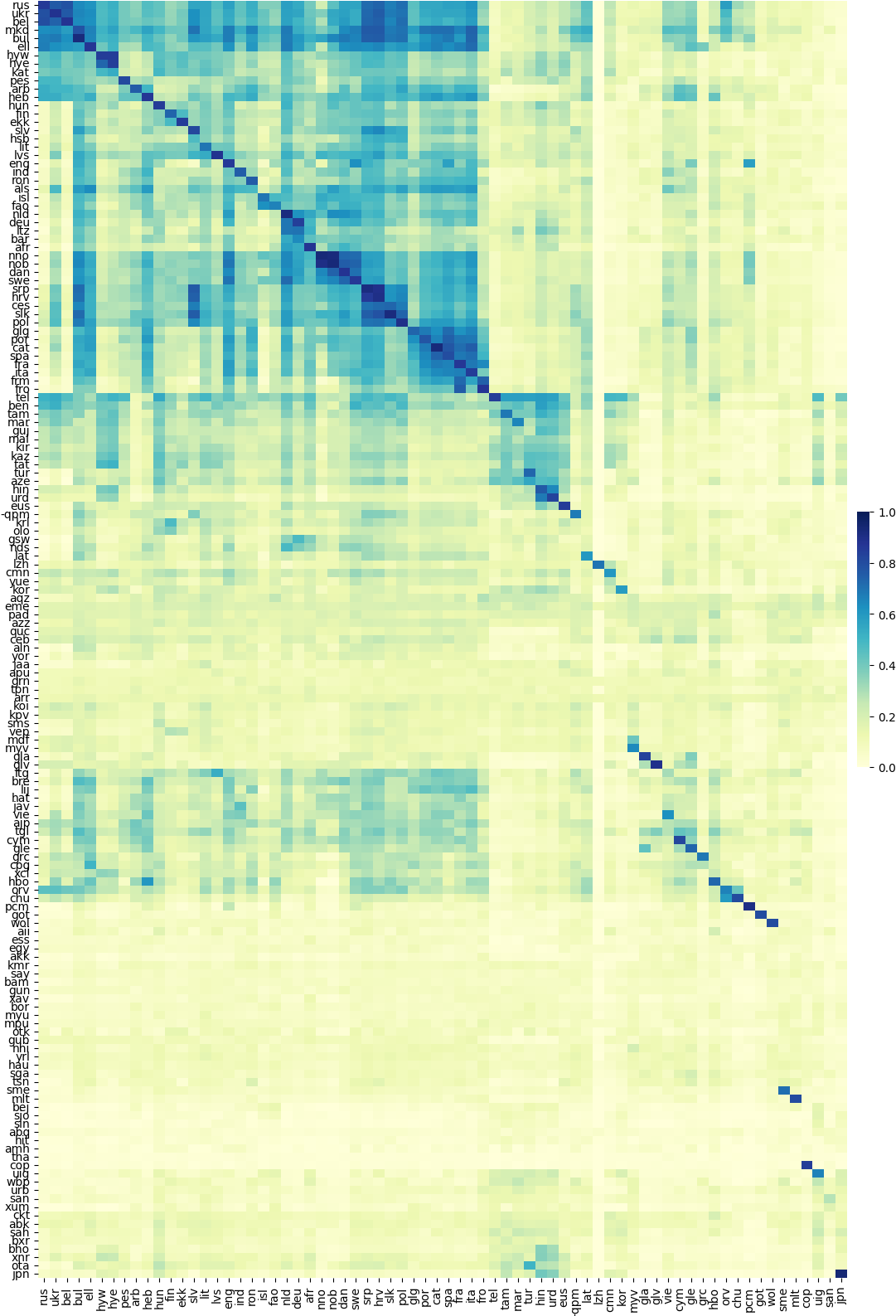}
    \caption{\textbf{Labelled attachment scores} for all combinations of source (columns) and target languages (rows), ordered by target language clusters (Ward's method).
    The darker a cell, the better the score.
    }
    \label{fig:heatmap-las}
\end{figure*}

\begin{figure*}
    \centering
    \includegraphics[height=0.9\textheight]{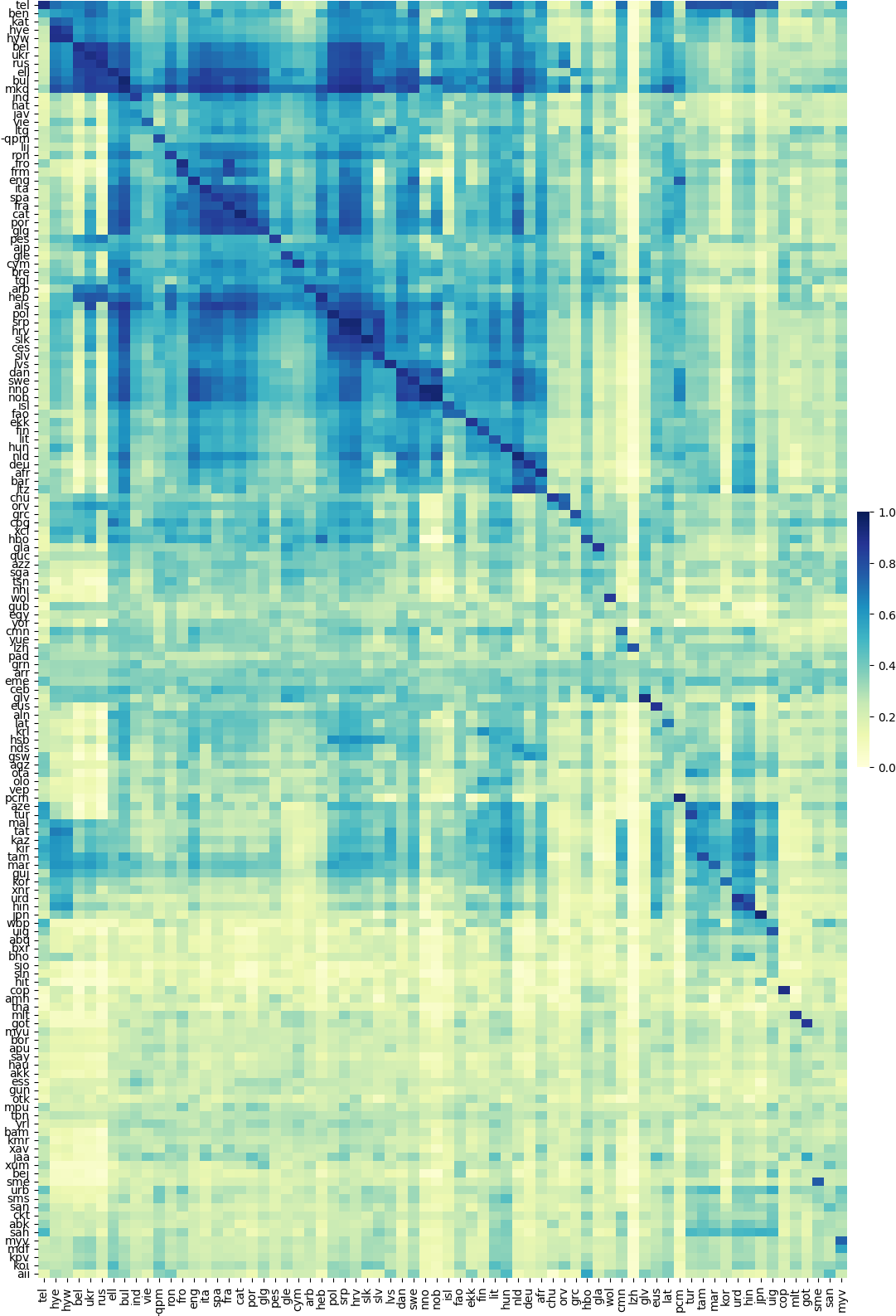}
    \caption{\textbf{Unlabelled attachment scores} for all combinations of source (columns) and target languages (rows), ordered by target language clusters (Ward's method).
    The darker a cell, the better the score.
    }
    \label{fig:heatmap-uas}
\end{figure*}

\begin{figure*}
    \centering
    \includegraphics[width=\textwidth]{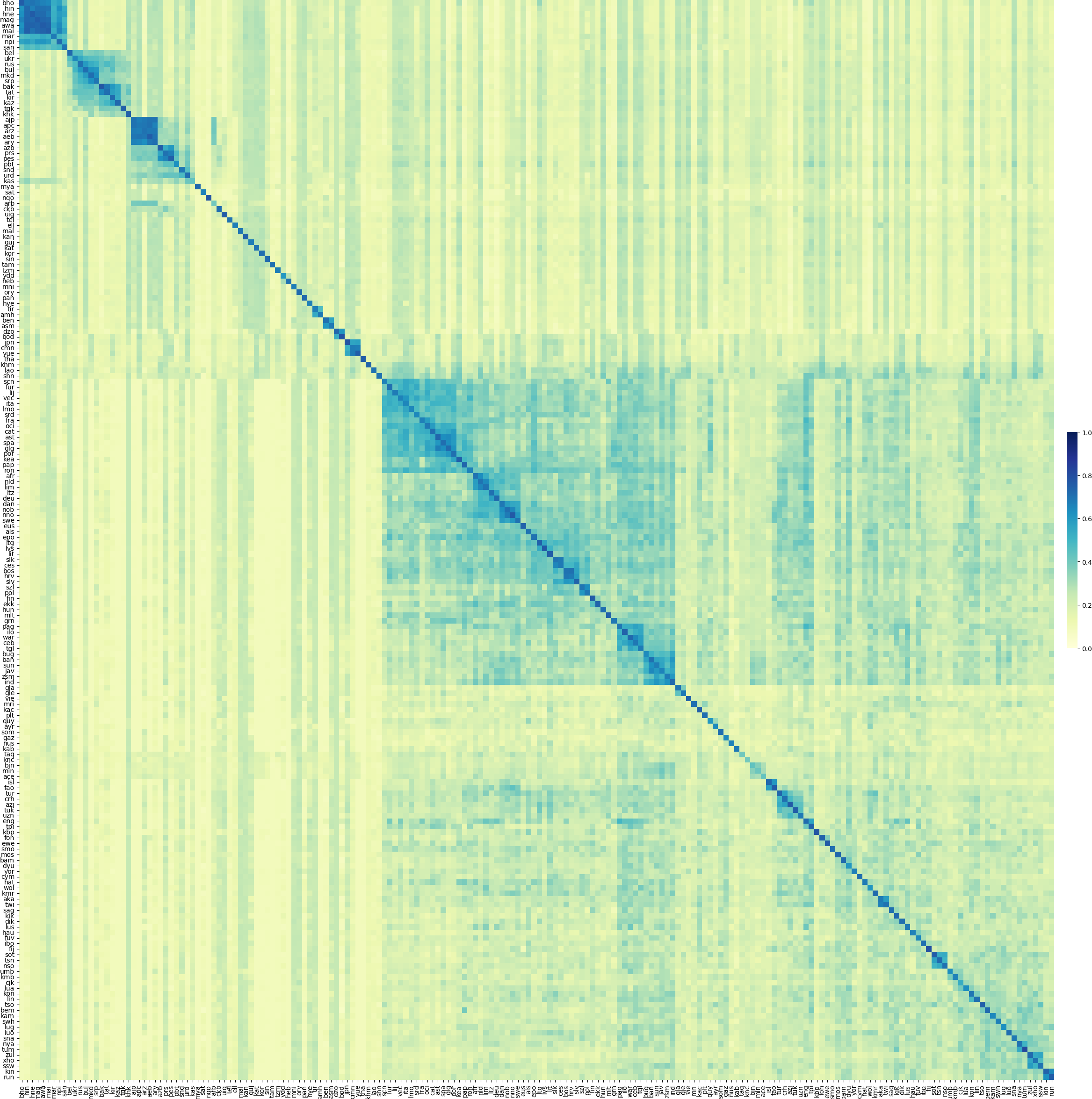}
    \caption{\textbf{Topic classification accuracy scores (MLP with n-grams, original writing systems)} for all combinations of source (columns) and target languages (rows), ordered by target language clusters (Ward's method).
    The darker a cell, the better the score.
    }
    \label{fig:heatmap-topics-original}
\end{figure*}

\begin{figure*}
    \centering
    \includegraphics[width=\textwidth]{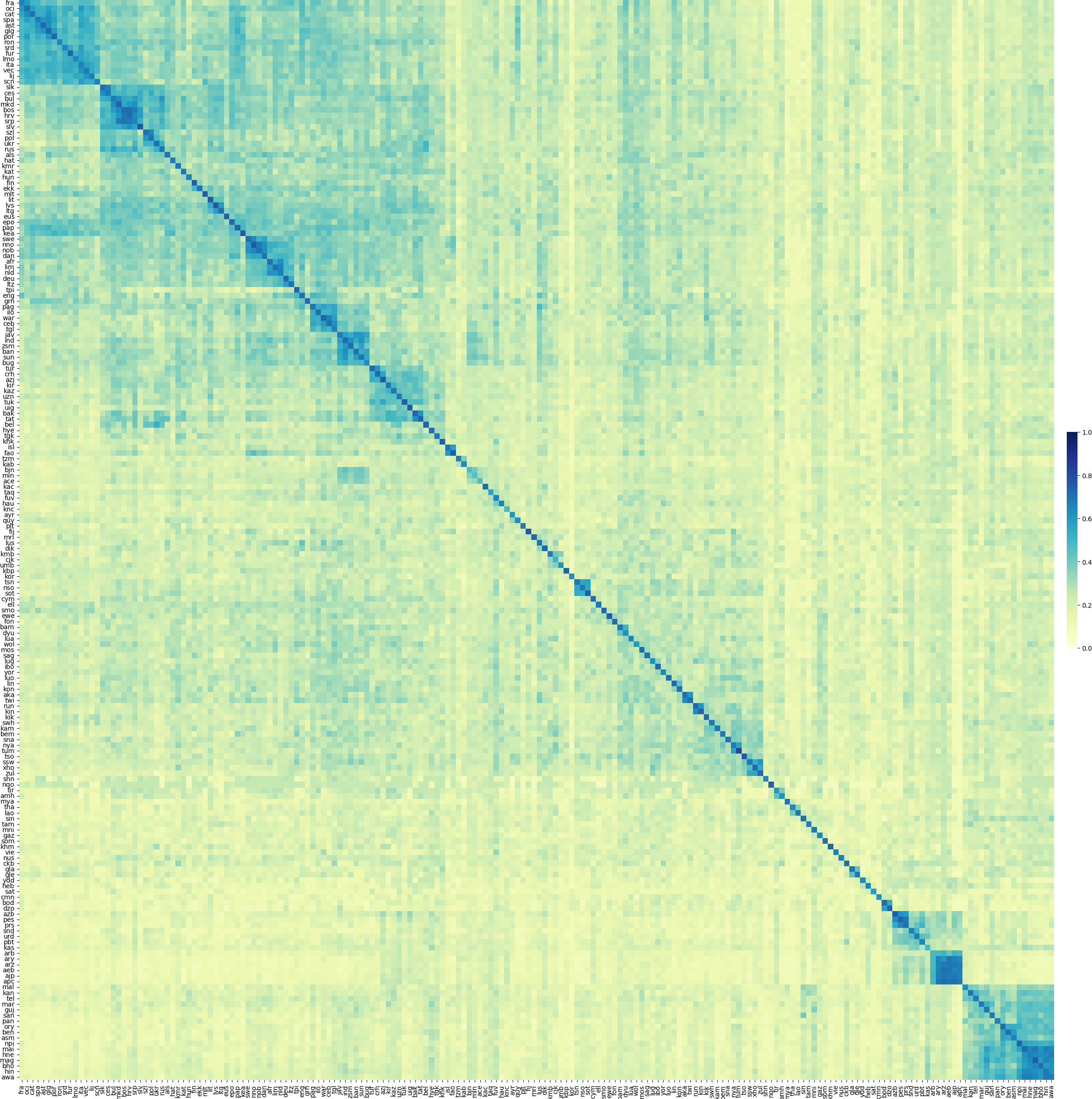}
    \caption{\textbf{Topic classification accuracy scores (MLP with n-grams, transliterated data)} for all combinations of source (columns) and target languages (rows), ordered by target language clusters (Ward's method).
    The darker a cell, the better the score.
    }
    \label{fig:heatmap-topics-translit}
\end{figure*}

\begin{figure*}
    \centering
    \includegraphics[width=\textwidth]{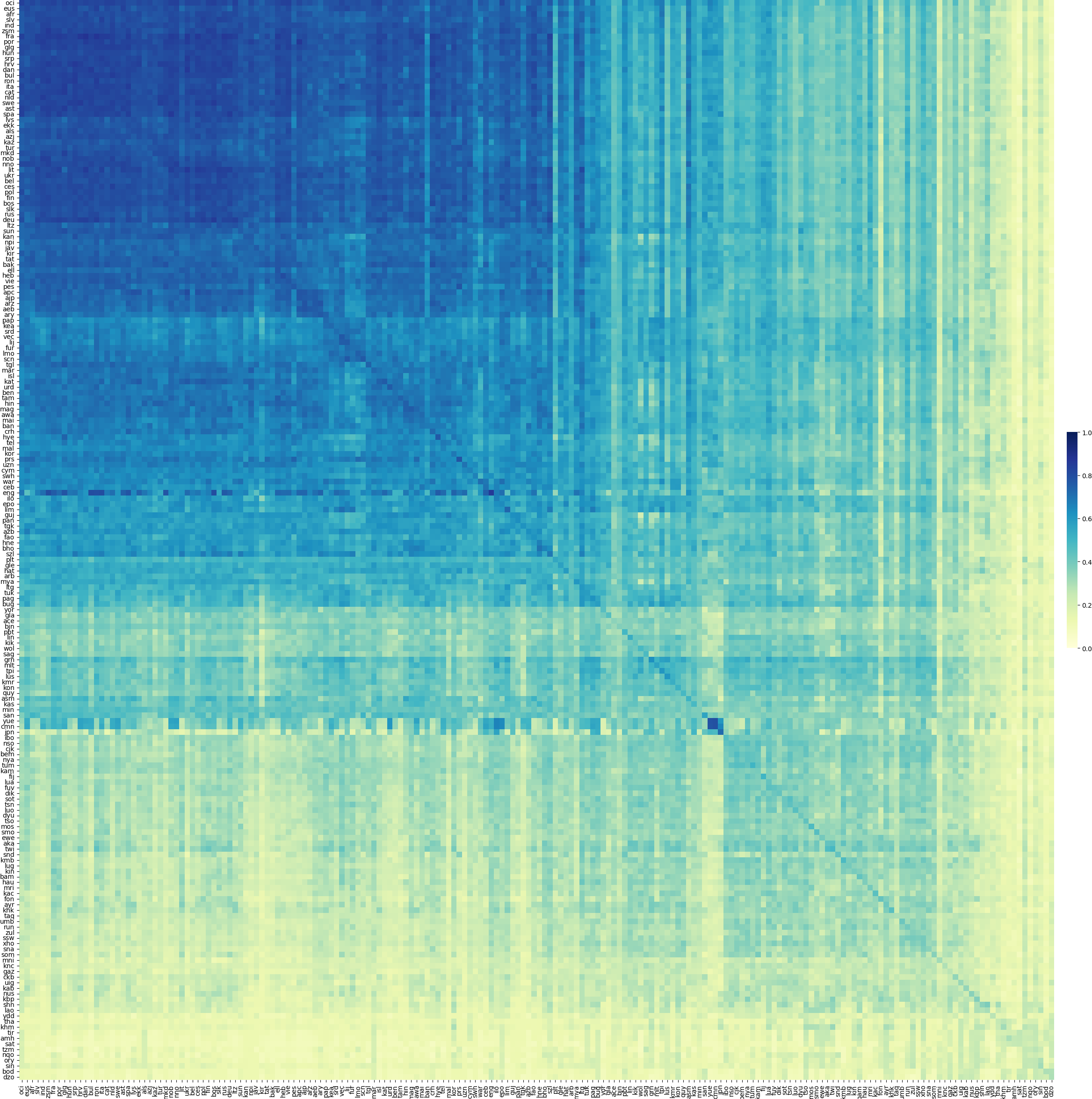}
    \caption{\textbf{Topic classification accuracy scores (MLP with mBERT representations)} for all combinations of source (columns) and target languages (rows), ordered by target language clusters (Ward's method).
    The darker a cell, the better the score.
    }
    \label{fig:heatmap-topics-mbert}
\end{figure*}

Figure~\ref{fig:heatmap-pos} shows the POS tagging accuracies for all language pairs.
Figures \ref{fig:heatmap-las} (LAS) and~\ref{fig:heatmap-uas} (UAS) present the parsing scores.
Figure~\ref{fig:heatmap-topics-original} shows the topic classification accuracies for \topicsbase, and Figure~\ref{fig:heatmap-topics-translit} for \topicstranslit.
Finally, Figure~\ref{fig:heatmap-topics-mbert} shows the results for the mBERT-based topic classification model.

\subsection{Robustness across datasets of the same language}
\label{sec:appendix-robustness}

In UD, many languages have multiple training and/or test datasets: there are 124 training datasets in 70 languages and 268 test datasets in 153 languages. These datasets differ in the sentences that are annotated (oftentimes, they are from different sources and text genres) and often they are annotated by different people.
Treebanks in the same language use the same writing system, with the exception of Sanskrit (and Mandarin Chinese if distinguishing between traditional and simplified characters).
In SIB-200, there are eight languages that have two datasets each, which differ in their writing system.

We compare how robust the transfer patterns are across datasets of the same language.
For test datasets, we calculate the correlation between the different models' scores on a pair of datasets.
For training datasets, we calculate the correlation between the evaluation scores produced by each pair of models trained on datasets of the same language.

\paragraph{Parsing}
For parsing (LAS), most datasets of the same language produce very similar transfer patterns.
Pearson's $r$ and Spearman's $\rho$ tend to produce very similar correlation numbers.
Most correlations (both from the training and testing side) are above $0.9$, and most correlations below that are still above $0.8$.\footnote{%
For the training datasets, we see exceptions ($r<0.8$) for two of the five Latin training sets (llct \& proiel: $r=0.79$) and the two German training sets (hdt \& gsd: $r=0.78$).
For the test datasets, the following treebank pairs show correlations with $r<0.8$:
two of the four Old East Slavic treebanks (birchbark \& rnc: $r=0.79$), 
several of the seven Mandarin Chinese treebanks (hk \& patentchar: $r=0.63$, beginner \& patentchar: $r=0.63$, beginner \& gsd: $r=0.69$. cfl \& patentchar: $r=0.71$, gsdsimp \& hk: $r=0.72$, beginner \& gsdsimp: $r=0.72$, gsd \& hk: $r=0.72$, cfl \& gsd: $r=0.76$, beginner \& pud: $r=0.79$, cfl \& gsdsimp: $r=0.79$), 
the two Akkadian treebanks (pisandub \& riao: $r=0.79$),
the two Tamil treebanks (mwtt \& ttb: $r=0.76$),
and the two Sanskrit treebanks (ufal \& vedic: $r=0.24$).
}

\paragraph{POS tagging}
Most treebank pairs of the same language also show very similar transfer patterns.
Again, most correlations are close to 1.\footnote{%
For the training treebanks, we see correlations ($r$) below $0.8$ only for the following pairs:
the two German training sets (hdt \& gsd: $r=0.75$), the two Persian ones (perft \& seraji: $r=0.73$), and the two Korean ones (gsd \& kaist: $r=0.72$).
For the test datasets this applies to
some of the six Latin treebanks (circse \& udante: $r=0.79$, circse \& ittb: $r=0.79$),
several of the eight Turkish datasets (atis \& tourism: $r=0.55$, pud \& tourism: $r=0.73$, penn \& tourism: $r=0.74$, boun \& tourism: $r=0.74$, gb \& tourism: $r=0.78$, kenet \& tourism: $r=0.79$),
several of the seven Mandarin Chinese treebanks (hk \& patentchar: $r=0.77$, beginner \& patentchar: $r=0.78$, patentchar \& pud: $r=0.79$, cfl \& patentchar: $r=0.79$, gsd \& patentchar: $r=0.79$),
two of the four Old East Slavic treebanks (birchbark \& rnc: $r=0.75$),
the two Akkadian treebanks (pisandub \& riao: $r=0.75$),
and both Sanskrit treebanks (ufal \& vedic: $r=0.57$).}

\paragraph{\topicsbase}
The writing system matters, and datasets in different writing systems (but the same language) show different transfer patterns.
The two Mandarin Chinese datasets (traditional vs.\ simplified characters) show similar patterns (correlation as training sets: $r=0.96$, as test sets: $r= 0.96$).
For the test datasets, nearly all other correlations are insignificant, except for the Arabic and Devanagari Kashmiri datasets ($r=0.43$).
For the training datasets, all correlations other than for the Mandarin datasets are either close to zero or negative.
The strongest negative correlation is for Arabic- vs.\ Latin-script Modern Standard Arabic ($r=-0.58$).

\paragraph{\topicstranslit}
For the transliterated version, all correlations are positive and/or close to zero, but not very high (the highest correlations are for the transliterated versions of the Tifinagh and Latin-script Tamasheq datasets: test $r=0.63$, train $r=0.34$). 
Note that we only have one transliterated version of the Mandarin Chinese datasets, as the transliteration tool did not work for the traditional script version.
We hypothesize that the correlations for the transliterated data are fairly low since they do not necessarily have many $n$-grams in common.
For instance, many of the languages with two datasets have one Arabic-script version and one Latin-script version. The latter contains vowels, while the automatically produced transliteration of the former only includes consonants.

\paragraph{\topicsmbert}
For \topicsmbert, the correlations tend to be higher than for the $n$-gram-based models.
For the training data, all correlations are above $0.6$ except for Achinese (which shows no significant correlation).
For the test data, the results are more split, with mostly positive correlations, but also some negative ones (Minangkabau, $-0.24$; Banjar, $r=-0.32$).
We hypothesize that the test data shows less coherent correlation patterns due to the inclusion of a language (in a given script) in mBERT's test data having a stronger effect on the classification results than the inclusion of the training dataset's language.

\subsection{Effect of writing system}
\label{sec:results-writing-systems}

We compare transfer between languages using the same writing system to transfer across writing systems.
For UD, we use GlotScript \cite{kargaran-etal-2024-glotscript} to determine the scripts; for SIB-200, writing system information is included as metadata.

\textbf{Transfer between datasets with the same writing systems generally works better than between different scripts, however this is in part due to the language selection rather than the scripts themselves.}
Two-sample Kolmogorov–Smirnov tests indicate that the results within vs.\ across scripts come from different distributions (\textit{p}-values all <0.0001; statistics: 0.51 for \topicsbase, 0.26 for \topicsmbert, 0.16 for POS accuracy, 0.18 for LAS).
However, the results of \topicstranslit{} for datasets that were in the same vs.\ different scripts before transliteration also come from different distributions (\textit{p} <0.0001, statistic: 0.37), indicating that at least for SIB-200, the combinations of languages usually associated with the scripts alone already make an important difference.

Nonetheless, writing systems still play a role, especially for the $n$-gram-based models.
Although \topicsbase{} and \topicstranslit{} achieve the same within-language accuracy (Table~\ref{tab:intra-vs-crosslingual-results}), \topicstranslit{} performs slightly better cross-lingually.
Its within-dataset performance is slightly lower than for \topicsbase{} (69.4\% vs.\ 70.3\%), likely due to some language-specific information being lost when diacritics are removed. This would be compensated in the within-language performance by improved transfer between datasets of languages that originally had different scripts.

\subsubsection{Effects of transliteration}
\label{sec:appendix-transliteration}

\begin{figure*}%
    \centering
    \includegraphics[width=0.9\textwidth]{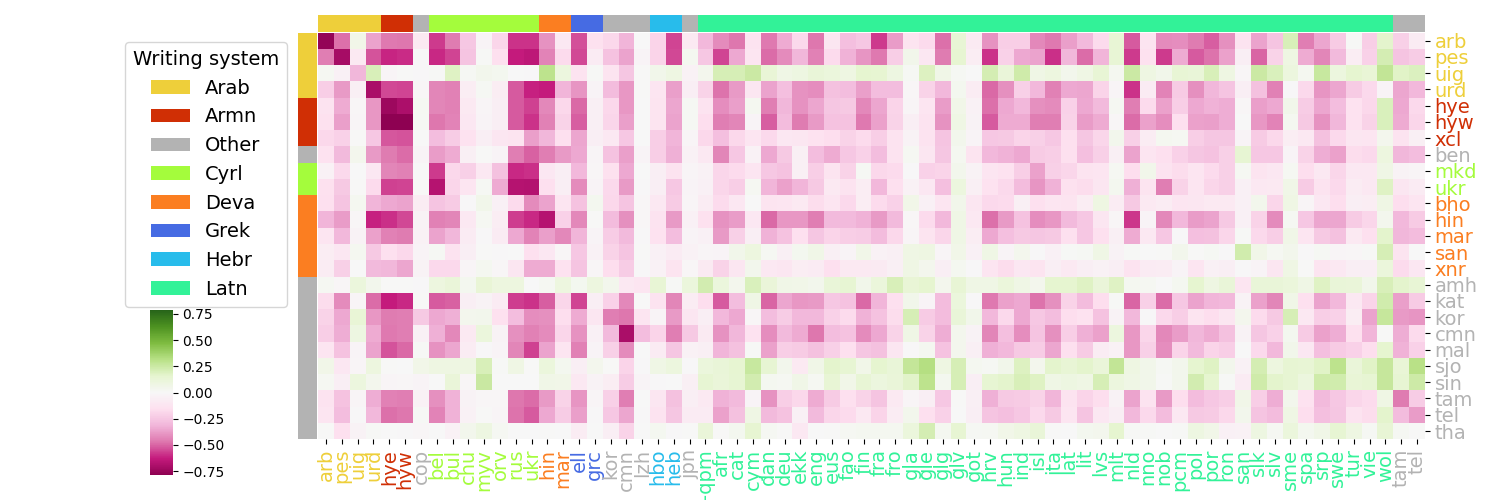}
    \includegraphics[width=0.9\textwidth]{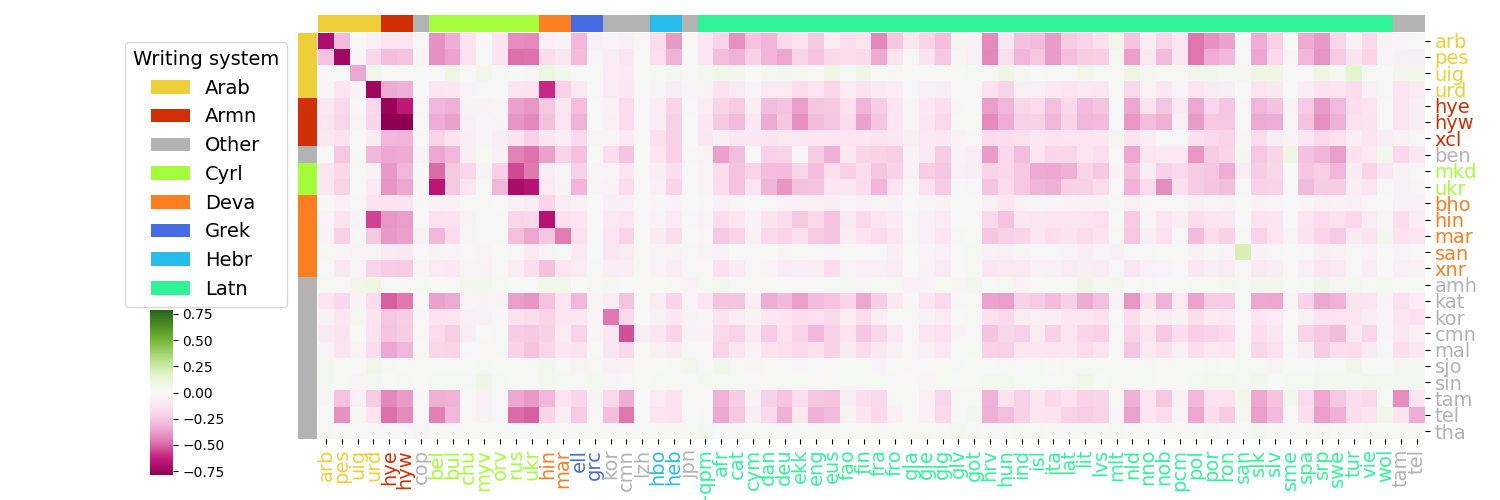}
    \caption{\textbf{Differences between the performance on the original test treebanks and their transliterated counterparts for POS tagging (top) and parsing (LAS, bottom).}
    Rows are for test sets, columns for training sets.
    Pink cells mark configurations where scores are better on the original data; green where scores are better on the transliterated treebank.
    Original writing systems are colour-coded. Writing systems in grey appear only for one language.
    }
    \label{fig:translit-differences-pos-las}
\end{figure*}

\paragraph{Transliterated UD treebanks}
Thirty-five UD test treebanks (in 25 languages) come with token-level Latin transliterations. 
We compare performance on the original data with performance on transliterated data.
Figure~\ref{fig:translit-differences-pos-las} shows the performance differences of the POS taggers and parsers when evaluated on transliterated instead of original-script data.
Performance is worse on most transliterated test treebanks, even when the models were trained on Latin-script treebanks.\footnote{%
The exceptions are three languages whose scripts appear neither in any training treebanks nor in mBERT's pretraining data (Amharic, Xibe, Sinhala), and one language that is usually written in Arabic, but unrelated to other languages using this writing system (Uyghur).}
This is in line with results from \citet{pires-etal-2019-multilingual}, who observe that mBERT performs worse on transliterated than original-script data.

\begin{figure*}[tb]
    \centering
    \includegraphics[width=\textwidth, trim={12.8cm 0 0 0}, clip]{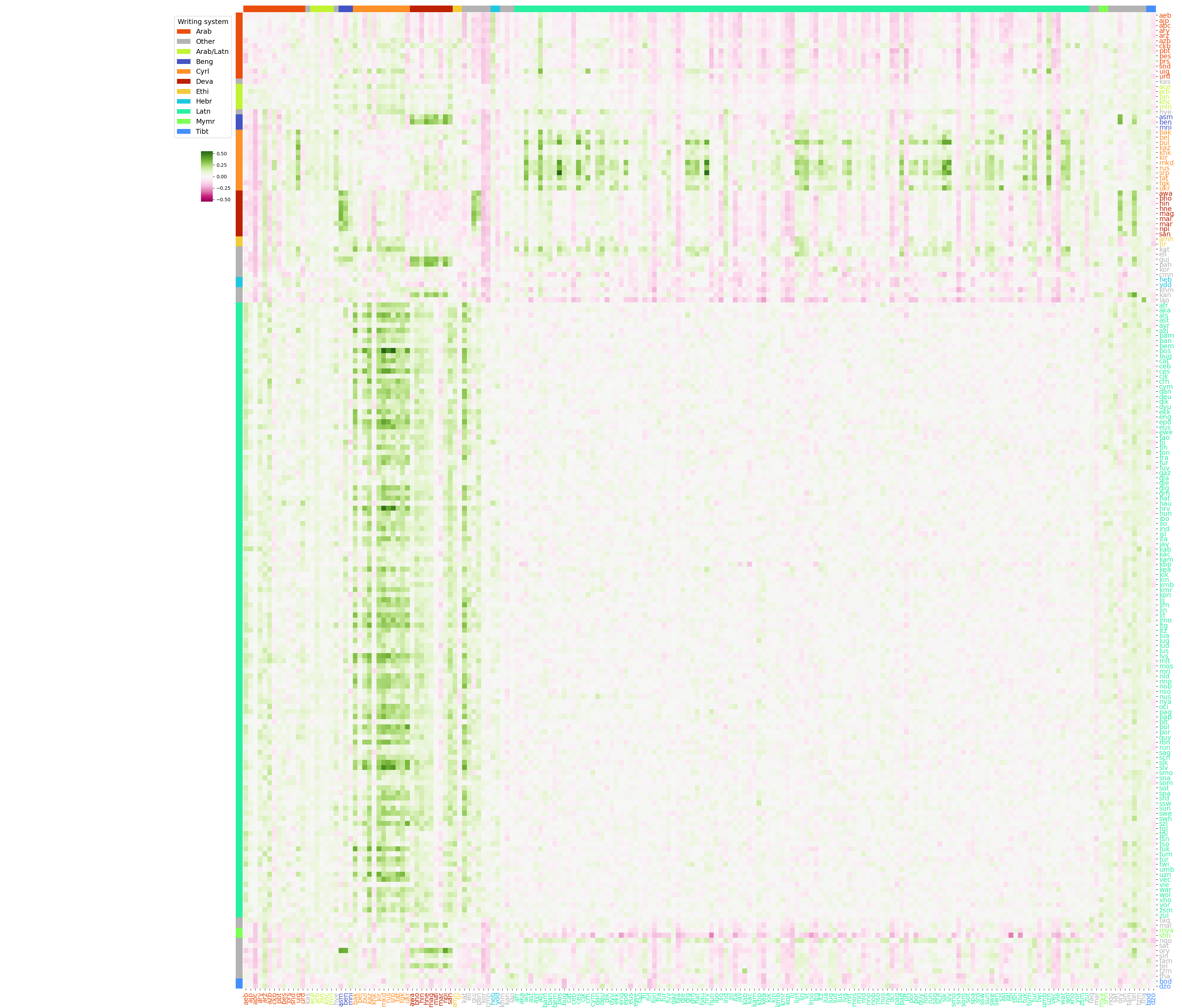}
    \caption{\textbf{Differences between the n-gram-based topic classification performance on the original test languages and their transliterated counterparts.}
    Rows are for test sets, columns for training sets.
    Pink cells mark configurations where scores are better on the original data; green where scores are better on the transliterated data.
    Original writing systems are colour-coded. Writing systems in grey appear only in one language.
    }
    \label{fig:translit-differences-topics}
\end{figure*}

\paragraph{Topic classification}
We compare the results of \topicsbase{} and \topicstranslit.
Figure~\ref{fig:translit-differences-topics} shows the topic classification performance difference between the $n$-gram model trained and evaluated on the original data and the one trained and evaluated on transliterated data.
For \topicstranslit, transfer between many original Cyrillic- and Latin-script language pairs is improved.

Eight of the languages in SIB-200 come in two script versions each.
The writing systems are completely distinct, except for Mandarin Chinese, which has versions in traditional and simplified characters.
For \topicsbase{}, transfer between the two scripts of a language is unsurprisingly always much lower than the performance on the same script (with accuracies between 8.3\% and 25.9\% for in-language cross-script transfer -- excluding the Mandarin Chinese entries, for which cross-script accuracy is up to 64.7\% -- vs.\ within-language, within-script accuracies between 60.8\% and 75.9\% for the same languages).
Although transfer between the transliterated versions of the datasets works better (between 14.2\% and 50.5\%), the accuracies are still much lower than the within-language, within-original-script accuracies (between 58.3\% and 74.0\%). This is likely due to different transliteration conventions for different writing systems (and due to missing vowels in the transliterated versions of abjads).

The patterns are also similar for \topicsmbert, with within-language, cross-script scores between 8.3\% and 45.6\%, compared to within-language, within-script accuracies between 42.2\% and 79.9\%.

\section{Correlations}
\label{sec:appendix-correlation-tables}

\subsection{Correlations between task results}
\label{sec:appendix-correlation-tables-tasks}

\begin{table}
\centering
\adjustbox{max width=\linewidth}{%
\begin{tabular}{@{}lccccc}
\toprule
& \multirow{2}{*}{POS} & \multicolumn{1}{c}{\multirow{2}{*}{LAS}} & \multicolumn{1}{c}{\multirow{2}{*}{UAS}} & \multicolumn{2}{c}{\texttt{topics}} \\
&  & \multicolumn{1}{c}{} & \multicolumn{1}{c}{} & \texttt{base} & \texttt{trans} \\
\midrule
LAS & 0.86 \\
UAS & 0.83 & 0.95 \\
\topicsbase & 0.39 & 0.43 & 0.40 \\
\topicstranslit & 0.40 & 0.53 & 0.48 & 0.68 \\
\topicsmbert & 0.64 & 0.58 & 0.56 & 0.28 & 0.36 \\
\bottomrule
\end{tabular}
}
\caption{\textbf{Correlations between task results} (Pearson's~\textit{r}, all \textit{p-}values are below 0.0001). Where possible, correlations are on a dataset level, otherwise on a language level.}
\label{tab:correlations-between-tasks}
\vspace{-10pt}
\end{table}

Table~\ref{tab:correlations-between-tasks} shows the correlations between task results. It is described in \S\ref{sec:results-across-task}.
The correlations across different task types only involve the 55 training and 84 test languages that appear both in UD and in SIB-200 (54 and 82 in the comparisons with the transliterated SIB-200 data).

\subsection{Mixed effects models}
\label{sec:appendix-correlation-tables-mixed-effects}
\begin{table*}[tb]
    \centering
    \adjustbox{max width=\textwidth}{%
    \newcommand{\insignificant}[1]{\color{black!30}#1}
\newlength{\minispace}
\setlength{\minispace}{3pt}
\newlength{\mediumspace}
\setlength{\mediumspace}{6pt}

\begin{tabular}{@{}r@{\hspace{\minispace}}l@{}r@{\hspace{\mediumspace}}r@{\hspace{\minispace}}lr@{\hspace{\mediumspace}}r@{\hspace{\minispace}}lr@{\hspace{\mediumspace}}r@{\hspace{\minispace}}lr@{\hspace{\mediumspace}}r@{\hspace{\minispace}}lr@{\hspace{\mediumspace}}r@{\hspace{\minispace}}lr@{\hspace{\mediumspace}}r@{\hspace{\minispace}}l@{}}
\toprule
&  & \multicolumn{3}{c}{\textbf{POS}} & \multicolumn{3}{c}{\textbf{LAS}} & \multicolumn{3}{c}{\textbf{UAS}} & \multicolumn{3}{c}{\textbf{\topicsbase}} & \multicolumn{3}{c}{\textbf{\topicstranslit}} & \multicolumn{3}{c}{\textbf{\topicsmbert}} \\
& \textbf{Fixed effect} & \textbf{Est.} & \multicolumn{1}{l}{\boldmath$\chi^2$} & \textbf{p} & \textbf{Est.} & \multicolumn{1}{l}{\boldmath$\chi^2$} & \textbf{p} & \textbf{Est.} & \multicolumn{1}{l}{\boldmath$\chi^2$} & \textbf{p} & \textbf{Est.} & \multicolumn{1}{l}{\boldmath$\chi^2$} & \textbf{p} & \textbf{Est.} & \multicolumn{1}{l}{\boldmath$\chi^2$} & \textbf{p} & \textbf{Est.} & \multicolumn{1}{l}{\boldmath$\chi^2$} & \textbf{p} \\
 \midrule
& (Intercept) & -0.30 & -4.09 &  & -0.60 & -9.75 &  & -0.41 & -5.58 &  & 0.05 & 0.91 &  & -0.12 & -2.28 &  & -0.13 & -1.45 &  \\
\multirow{2}{*}{\textrm{corr}\hspace{-1em}$\left.\begin{array}{l}
                \\
                \\
                \end{array}\right\lbrace$} & \grambank & 0.15 & 10.92 & *** & 0.27 & 43.08 & *** & 0.27 & 30.25 & *** & \insignificant{-0.06} & \insignificant{2.84} & \insignificant{.} & \insignificant{0.04} & \insignificant{1.42} & \insignificant{} & \insignificant{0.07} & \insignificant{1.51} & \insignificant{} \\
& \syntax & 0.21 & 67.5 & *** & 0.53 & 503.28 & *** & 0.56 & 415.29 & *** & \insignificant{0.00} & \insignificant{0.00} & \insignificant{} & \insignificant{-0.01} & \insignificant{0.26} & \insignificant{} & 0.07 & 6.38 & * \\
& \phono & -0.07 & 4.19 & * & \insignificant{-0.03} & \insignificant{0.77} & \insignificant{} & \insignificant{0.02} & \insignificant{2.28} & \insignificant{} & \insignificant{0.02} & \insignificant{0.87} & \insignificant{} & \insignificant{-0.02} & \insignificant{1.07} & \insignificant{} & 0.14 & 12.12 & *** \\
& \inventory & 0.30 & 20.17 & *** & \insignificant{0.07} & \insignificant{1.14} & \insignificant{} & \insignificant{0.00} & \insignificant{0} & \insignificant{} & \insignificant{-0.01} & \insignificant{0.01} & \insignificant{} & 0.12 & 6.04 & * & \insignificant{0.10} & \insignificant{1.04} & \insignificant{} \\
\multirow{2}{*}{\textrm{corr}\hspace{-1em}$\left.\begin{array}{l}
                \\
                \\
                \end{array}\right\lbrace$} & \asjp & \insignificant{-0.07} & \insignificant{3.66} & \insignificant{.} & -0.13 & 16.71 & *** & -0.19 & 23.58 & *** & \insignificant{0.07} & \insignificant{3.65} & \insignificant{.} & 0.24 & 64.27 & *** & -0.16 & 7.64 & **  \\
& \genetic & 0.28 & 95.21 & *** & 0.36 & 194.29 & *** & 0.32 & 110.28 & *** & 0.19 & 54.13 & *** & 0.06 & 5.65 & * & 0.21 & 21.86 & *** \\
& \geo & 0.15 & 22.92 & *** & 0.18 & 39.61 & *** & 0.19 & 32.08 & *** & 0.04 & 54.13 & *** & 0.07 & 16.15 & *** & \insignificant{0.06} & \insignificant{3.31} & \insignificant{.} \\
& \overlapword/\overlaptri/\overlapsubword & \insignificant{-0.03} & \insignificant{0.14} & \insignificant{} & 0.36 & 21.00 & *** & \insignificant{0.05} & \insignificant{0.26} & \insignificant{} & 0.52 & 125.83 & *** & 0.75 & 477.22 & *** & \insignificant{-0.12} & \insignificant{3.00} & \insignificant{.} \\
\multirow{2}{*}{\textrm{corr}\hspace{-1em}$\left.\begin{array}{l}
                \\
                \\
                \end{array}\right\lbrace$} & \overlapchar & \insignificant{0.01} & \insignificant{0.273} & \insignificant{} & -0.07 & 24.14 & *** & -0.04 & 4.90 & * & 0.25 & 136.93 & *** & \insignificant{-0.08} & \insignificant{2.83} & \insignificant{.} & 0.26 & 7.37 & *** \\
& same\_scriptTrue & 0.13 & 311.95 & *** & 0.06 & 67.62 & *** & 0.10 & 142.68 & *** & -0.03 & 9.16 & ** &  &  &  & -0.04 & -3.17 & *** \\
& mbert\_testTrue & 0.22 & 20.81 & *** & 0.12 & 16.75 & *** & 0.14 & 13.98 & *** &  &  &  &  &  &  & 0.26 & 8.76 & *** \\
& \trainingsize & 0.00 & 26.34 & *** & 0.00 & 4.90 & * & 0.00 & 6.52 & * &  &  &  &  &  &  &  &  &  \\
\midrule
& \# Train langs & 19 &  &  & 19 &  &  & 19 &  &  & 42 &  &  & 40 &  &  & 42 &  &  \\
& \# Test langs & 31 &  &  & 31 &  &  & 31 &  &  & 42 &  &  & 40 &  &  & 42 &  & \\
\bottomrule
\end{tabular}}
    \caption{\textbf{Linear mixed effects model results for each experiment.}
    \textit{P}-values are based on model comparison: *** = < 0.001, ** = < 0.01, * = < 0.05, . = < 0.1. Entries with p-values of >=0.05 are in grey.
    The last two rows show the number of training and test languages included in the analysis (i.e., the language pairs for which no fixed effects had missing information).
    ``Corr'' indicates pairs of strongly correlated fixed effects (see text, \S\ref{sec:appendix-correlation-tables-mixed-effects}).
    ``Same\_script'' is True iff the training and test datasets use the same writing system;
    ``mbert\_test'' is True iff the test language is one of mBERT's pretraining languages.
    }
    \label{tab:mixed-effects}
\end{table*}

As described at the end of \S\ref{sec:results-similarity-measures},
we fit one linear mixed effects model per experiment.
We model the NLP results (POS accuracy, parsing scores, topic classification accuracy) as the dependent variable, and the training and test languages as random effects.
The fixed effects are the similarity measures, as well as binary variables (dummy-coded) indicating whether the training and test datasets have the same writing system and whether the test language is among mBERT's pretraining languages.
Because the models can only be fit for entries where no data points are missing (i.e., \geo{}, \asjp{}, \syntax{}, \phono{}, \inventory{}, and \grambank{} are all defined for the language pair at hand), the number of language pairs included in each mixed effects analysis is much smaller than for the correlations calculated independently per effect in \S\ref{sec:results-similarity-measures}.

Collinearity was observed between several fixed effects in all models (with correlation coefficients between $-0.769$ and $-0.819$ for phylogenetic relatedness and lexical similarity, between $-0.442$ and $-0.496$ for character overlap and sharing the same writing system, and between $-0.447$ and $-0.509$ for \grambank{} and \syntax{}).
We mark them as such in Table~\ref{tab:mixed-effects}, which shows the estimates and their significance values. We report the significance values based on model comparison (i.e., by comparing the full model and a model with one predictor taken out) and thus significance values are robust to collinearity. 

The values for these related effects should be interpreted with these correlations in mind, as this can impact the estimates for these variables. 

\subsection{Correlations between NLP results and similarity measures}
\label{sec:appendix-correlation-tables-similarity-and-tasks}
Tables~\ref{tab:correlations-pos-las-i} and~\ref{tab:correlations-pos-las-ii} show the correlations between POS/parsing scores and the similarity measures for each test language.
Tables~\ref{tab:correlations-topics-i}, \ref{tab:correlations-topics-ii}, and~\ref{tab:correlations-topics-iii} show the same for the topic classification experiments.
Because these tables take up a lot of space, they are placed at the very end of the appendix.

\FloatBarrier

\begin{figure*}
    \centering
    \includegraphics[height=0.9\textheight]{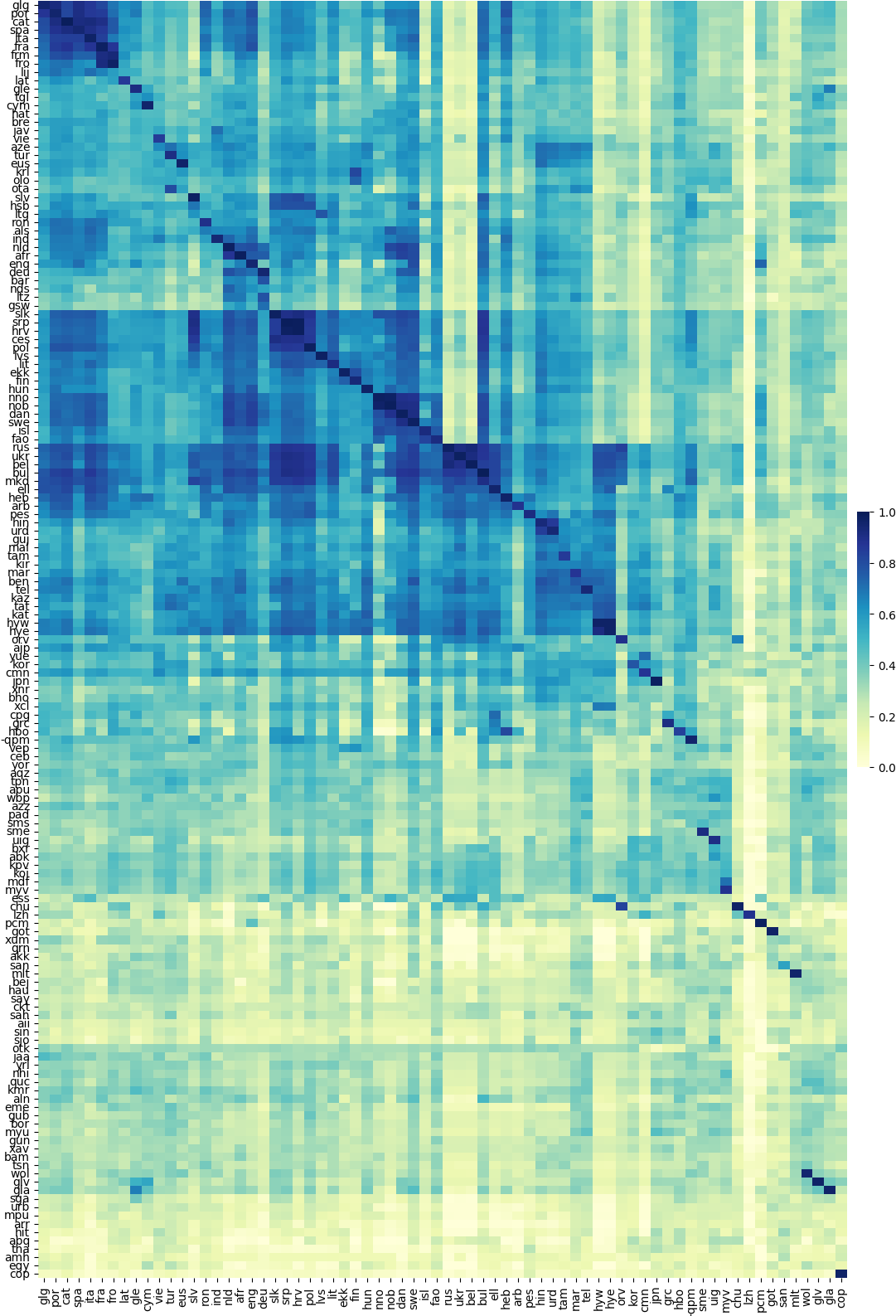}
    \caption{\textbf{POS tagging accuracy scores} for all combinations of source (columns) and target languages (rows), ordered by target language clusters (Ward's method).
    The darker a cell, the better the score.
    }
    \label{fig:heatmap-pos}
\end{figure*}

\begin{table*}
\centering
\adjustbox{max width=0.8\textwidth}{%
\begin{tabular}{@{}lll|rrrrrrrrrr|rrrrrrrrrr@{}}
\toprule
\multicolumn{3}{c|}{Test language} & \multicolumn{10}{c|}{POS} & \multicolumn{10}{c}{LAS} \\
lang & \multicolumn{2}{l|}{script mBERT} & \trainingsize & \phono & \inventory & \geo & \syntax & \grambank & \genetic & \asjp & \overlapchar & \overlapword & \trainingsize & \phono & \inventory & \geo & \syntax & \grambank & \genetic & \asjp & \overlapchar & \overlapword \\
\midrule
abk & Cyrl &  & \cellcolor[HTML]{f7b396} -36\rlap{\textsuperscript{*}} & \cellcolor[HTML]{dcdddd} --- & N/A & \cellcolor[HTML]{dcdddd} --- & \cellcolor[HTML]{b1cbfc} 29 & \cellcolor[HTML]{dcdddd} --- & \cellcolor[HTML]{dcdddd} --- & \cellcolor[HTML]{dcdddd} --- & \cellcolor[HTML]{dcdddd} --- & \cellcolor[HTML]{afcafc} 30 & \cellcolor[HTML]{f6bda2} -30 & \cellcolor[HTML]{dcdddd} --- & N/A & \cellcolor[HTML]{dcdddd} --- & \cellcolor[HTML]{85a8fc} 54\rlap{\textsuperscript{*}} & \cellcolor[HTML]{dcdddd} --- & \cellcolor[HTML]{dcdddd} --- & \cellcolor[HTML]{dcdddd} --- & \cellcolor[HTML]{dcdddd} --- & \cellcolor[HTML]{afcafc} 30\rlap{\textsuperscript{*}} \\
abq & Cyrl &  & \cellcolor[HTML]{dcdddd} --- & N/A & N/A & \cellcolor[HTML]{dcdddd} --- & N/A & N/A & \cellcolor[HTML]{dcdddd} --- & \cellcolor[HTML]{dcdddd} --- & \cellcolor[HTML]{bad0f8} 24 & \cellcolor[HTML]{dcdddd} --- & \cellcolor[HTML]{dcdddd} --- & N/A & N/A & \cellcolor[HTML]{dcdddd} --- & N/A & N/A & \cellcolor[HTML]{dcdddd} --- & \cellcolor[HTML]{dcdddd} --- & \cellcolor[HTML]{dcdddd} --- & \cellcolor[HTML]{dcdddd} --- \\
\bottomrule
\end{tabular}
}
\caption{\textbf{Correlations between POS/LAS results and similarity measures.} Continued/explained in next table.}
\label{tab:correlations-pos-las-i}
\end{table*}

\begin{table*}
\centering
\adjustbox{max width=0.8\textwidth}{%
\begin{tabular}{@{}lll|rrrrrrrrrr|rrrrrrrrrr@{}}
\toprule
\multicolumn{3}{c|}{Test language} & \multicolumn{10}{c|}{POS} & \multicolumn{10}{c}{LAS} \\
lang & \multicolumn{2}{l|}{script mBERT} & \trainingsize & \phono & \inventory & \geo & \syntax & \grambank & \genetic & \asjp & \overlapchar & \overlapword & \trainingsize & \phono & \inventory & \geo & \syntax & \grambank & \genetic & \asjp & \overlapchar & \overlapword \\
\midrule
kor & Hang & x & \cellcolor[HTML]{dcdddd} --- & \cellcolor[HTML]{dcdddd} --- & \cellcolor[HTML]{dcdddd} --- & \cellcolor[HTML]{adc9fd} 32\rlap{\textsuperscript{*}} & \cellcolor[HTML]{84a7fc} 55\rlap{\textsuperscript{*}} & \cellcolor[HTML]{89acfd} 52\rlap{\textsuperscript{*}} & \cellcolor[HTML]{a7c5fe} 35\rlap{\textsuperscript{*}} & \cellcolor[HTML]{9ebeff} 40\rlap{\textsuperscript{*}} & \cellcolor[HTML]{a9c6fd} 34\rlap{\textsuperscript{*}} & \cellcolor[HTML]{a3c2fe} 37\rlap{\textsuperscript{*}} & \cellcolor[HTML]{dcdddd} --- & \cellcolor[HTML]{94b6ff} 46 & \cellcolor[HTML]{aec9fc} 31 & \cellcolor[HTML]{a5c3fe} 36\rlap{\textsuperscript{*}} & \cellcolor[HTML]{5a78e4} \color{black!10} 79\rlap{\textsuperscript{*}} &\cellcolor[HTML]{6485ec} \color{black!10} 73\rlap{\textsuperscript{*}} &\cellcolor[HTML]{81a4fb} 57\rlap{\textsuperscript{*}} & \cellcolor[HTML]{7da0f9} 59\rlap{\textsuperscript{*}} & \cellcolor[HTML]{7ea1fa} 58\rlap{\textsuperscript{*}} & \cellcolor[HTML]{7ea1fa} 58\rlap{\textsuperscript{*}} \\
\bottomrule
\end{tabular}
}
\caption{\textbf{Correlations (Pearson's \textit{r} $\times$ 100, to save space) between POS/LAS results and similarity measures for each test language, continued.} 
Where a training or test language has multiple datasets, we use language-wise score averages for calculating the correlations.
The asterisk* denotes correlations with a \textit{p}-value below 0.01.
Grey cells with a bar (---) denote correlations with a \textit{p}-value of 0.05 or above.
Square brackets [ ] mean that no entry for this ISO code was found in the linguistic databases, so the entry for its macrolanguage code was used instead (\texttt{bul} for \texttt{qpm}, \texttt{apc} for \texttt{ajp}, \texttt{lav} for \texttt{lvs}).
`N/A' means that no correlation score could be calculated due to missing entries in the linguistic databases.
The bottom rows show the mean scores across test languages (`---' entries are treated as zeros, `N/A' entries are ignored).
Below each row of mean scores are the corresponding 95\% confidence intervals (numbers separated by semicolons).
`avg mBERT' (`avg $\neg$mBERT') is for the test languages that are (are not) in mBERT's pretraining data.
}
\label{tab:correlations-pos-las-ii}
\end{table*}

\begin{table*}
\centering
\adjustbox{max width=\textwidth}{%
\begin{tabular}{@{}l@{}l@{}c@{}|rrrrrrrrr|rrrrrrrrr|rrrrrrrrr@{}}
\toprule
\multicolumn{3}{c|}{Test language} & \multicolumn{9}{c|}{Original} & \multicolumn{9}{c|}{Transliterated} & \multicolumn{9}{c}{mBERT} \\
lang & \multicolumn{2}{l@{}|}{script \hspace{1mm} mB.~} & \phono & \inventory & \geo & \syntax & \grambank & \genetic & \asjp & \overlapchar & \overlaptri  & \phono & \inventory & \geo & \syntax & \grambank & \genetic & \asjp & \overlapchar & \overlaptri & \phono & \inventory & \geo & \syntax & \grambank & \genetic & \asjp & \overlapchar & \overlapsubword \\
\midrule
ace & \multicolumn{2}{@{}l@{}|}{Arab, Latn} & N/A & \cellcolor[HTML]{aac7fd} 33\rlap{\textsuperscript{*}} & \cellcolor[HTML]{dcdddd} --- & \cellcolor[HTML]{dcdddd} --- & \cellcolor[HTML]{a2c1ff} 38\rlap{\textsuperscript{*}} & \cellcolor[HTML]{89acfd} 52\rlap{\textsuperscript{*}} & \cellcolor[HTML]{8fb1fe} 49\rlap{\textsuperscript{*}} & \cellcolor[HTML]{9dbdff} 41\rlap{\textsuperscript{*}} & \cellcolor[HTML]{7b9ff9} \color{black!10} 60\rlap{\textsuperscript{*}} & N/A &\cellcolor[HTML]{b6cefa} 26\rlap{\textsuperscript{*}} & \cellcolor[HTML]{b3cdfb} 28\rlap{\textsuperscript{*}} & \cellcolor[HTML]{dcdddd} --- & \cellcolor[HTML]{97b8ff} 44\rlap{\textsuperscript{*}} & \cellcolor[HTML]{82a6fb} 56\rlap{\textsuperscript{*}} & \cellcolor[HTML]{88abfd} 53\rlap{\textsuperscript{*}} & \cellcolor[HTML]{dcdddd} --- & \cellcolor[HTML]{9ebeff} 40\rlap{\textsuperscript{*}} & N/A & \cellcolor[HTML]{dcdddd} --- & \cellcolor[HTML]{f4c5ad} -25\rlap{\textsuperscript{*}} & \cellcolor[HTML]{aac7fd} 33\rlap{\textsuperscript{*}} & \cellcolor[HTML]{dcdddd} --- & \cellcolor[HTML]{dcdddd} --- & \cellcolor[HTML]{dcdddd} --- & \cellcolor[HTML]{88abfd} 53\rlap{\textsuperscript{*}} & \cellcolor[HTML]{b3cdfb} 28\rlap{\textsuperscript{*}} \\
\bottomrule
\end{tabular}
}
\caption{\textbf{Correlations between topic classification results and similarity measures.} Continued in next table; full caption in Table~\ref{tab:correlations-topics-iii}.}
\label{tab:correlations-topics-i}
\end{table*}

\begin{table*}
\centering
\adjustbox{max width=\textwidth}{%
\begin{tabular}{@{}l@{}l@{}c@{}|rrrrrrrrr|rrrrrrrrr|rrrrrrrrr@{}}
\toprule
\multicolumn{3}{c|}{Test language} & \multicolumn{9}{c|}{Original} & \multicolumn{9}{c|}{Transliterated} & \multicolumn{9}{c}{mBERT} \\
lang & \multicolumn{2}{l@{}|}{script \hspace{1mm} mB.~} & \phono & \inventory & \geo & \syntax & \grambank & \genetic & \asjp & \overlapchar & \overlaptri  & \phono & \inventory & \geo & \syntax & \grambank & \genetic & \asjp & \overlapchar & \overlaptri & \phono & \inventory & \geo & \syntax & \grambank & \genetic & \asjp & \overlapchar & \overlapsubword \\
\midrule
kmb & Latn &  & N/A & N/A & \cellcolor[HTML]{adc9fd} 32\rlap{\textsuperscript{*}} & N/A & N/A & \cellcolor[HTML]{89acfd} 52\rlap{\textsuperscript{*}} & \cellcolor[HTML]{7b9ff9} \color{black!10} 60\rlap{\textsuperscript{*}} &\cellcolor[HTML]{8badfd} 51\rlap{\textsuperscript{*}} & \cellcolor[HTML]{7295f4} \color{black!10} 65\rlap{\textsuperscript{*}} & N/A & N/A &\cellcolor[HTML]{b7cff9} 25\rlap{\textsuperscript{*}} & N/A & N/A & \cellcolor[HTML]{92b4fe} 47\rlap{\textsuperscript{*}} & \cellcolor[HTML]{7093f3} \color{black!10} 66\rlap{\textsuperscript{*}} &\cellcolor[HTML]{dcdddd} --- & \cellcolor[HTML]{90b2fe} 48\rlap{\textsuperscript{*}} & N/A & N/A & \cellcolor[HTML]{9bbcff} 42\rlap{\textsuperscript{*}} & N/A & N/A & \cellcolor[HTML]{7da0f9} 59\rlap{\textsuperscript{*}} & \cellcolor[HTML]{9dbdff} 41\rlap{\textsuperscript{*}} & \cellcolor[HTML]{88abfd} 53\rlap{\textsuperscript{*}} & \cellcolor[HTML]{6f92f3} \color{black!10} 67\rlap{\textsuperscript{*}} \\
\bottomrule
\end{tabular}
}
\caption{\textbf{Correlations between topic classification results and similarity measures.} Continued and explained in next table.}
\label{tab:correlations-topics-ii}
\end{table*}

\begin{table*}
\centering
\adjustbox{max width=\textwidth}{%
\begin{tabular}{@{}llc|rrrrrrrrr|rrrrrrrrr|rrrrrrrrr@{}}
\toprule
\multicolumn{3}{c|}{Test language} & \multicolumn{9}{c|}{Original} & \multicolumn{9}{c|}{Transliterated} & \multicolumn{9}{c}{mBERT} \\
lang & script & mB. & \phono & \inventory & \geo & \syntax & \grambank & \genetic & \asjp & \overlapchar & \overlaptri  & \phono & \inventory & \geo & \syntax & \grambank & \genetic & \asjp & \overlapchar & \overlaptri & \phono & \inventory & \geo & \syntax & \grambank & \genetic & \asjp & \overlapchar & \overlapsubword \\
\midrule
tum & Latn &  & N/A & N/A & \cellcolor[HTML]{9bbcff} 42\rlap{\textsuperscript{*}} & N/A & N/A & \cellcolor[HTML]{6f92f3} \color{black!10} 67\rlap{\textsuperscript{*}} &\cellcolor[HTML]{6384eb} \color{black!10} 74\rlap{\textsuperscript{*}} &\cellcolor[HTML]{90b2fe} 48\rlap{\textsuperscript{*}} & \cellcolor[HTML]{7597f6} \color{black!10} 64\rlap{\textsuperscript{*}} & N/A & N/A &\cellcolor[HTML]{9bbcff} 42\rlap{\textsuperscript{*}} & N/A & N/A & \cellcolor[HTML]{7699f6} \color{black!10} 63\rlap{\textsuperscript{*}} &\cellcolor[HTML]{5f7fe8} \color{black!10} 76\rlap{\textsuperscript{*}} &\cellcolor[HTML]{dcdddd} --- & \cellcolor[HTML]{88abfd} 53\rlap{\textsuperscript{*}} & N/A & N/A & \cellcolor[HTML]{b7cff9} 25\rlap{\textsuperscript{*}} & N/A & N/A & \cellcolor[HTML]{92b4fe} 47\rlap{\textsuperscript{*}} & \cellcolor[HTML]{96b7ff} 45\rlap{\textsuperscript{*}} & \cellcolor[HTML]{82a6fb} 56\rlap{\textsuperscript{*}} & \cellcolor[HTML]{7ea1fa} 58\rlap{\textsuperscript{*}} \\
\bottomrule
\end{tabular}
}
\caption{\textbf{Correlations (Pearson's \textit{r} $\times$ 100, to save space) between topic classification results and similarity measures for each test language, continued.} 
Where a training or test language has multiple datasets (one per writing system), we use language-wise score averages for calculating the correlations.
The asterisk* denotes correlations with a \textit{p}-value below 0.01.
Grey cells with a bar (---) denote correlations with a \textit{p}-value of 0.05 or above.
Square brackets [ ] mean that no entry for this ISO code was found in the linguistic databases, so the entry for its macrolanguage code was used instead (\texttt{aze} for \texttt{azb} and \texttt{azj}, \texttt{apc} for \texttt{ajp}, \texttt{lav} for \texttt{lvs}, \texttt{zlm} for \texttt{zsm}).
`N/A' means that no correlation score could be calculated due to missing entries in the linguistic databases (or due to missing transliterations in the case of the transliteration experiment).
The bottom rows show the mean scores across test languages (`---' entries are treated as zeros, `N/A' entries are ignored).
Below each row of mean scores are the corresponding 95\% confidence intervals (numbers separated by semicolons).
`avg mBERT' (`avg $\neg$mBERT') is for the test languages that are (are not) in mBERT's pretraining data.
}
\label{tab:correlations-topics-iii}
\end{table*}

\end{document}